\documentclass[journal=jctcce,manuscript=article]{achemso}

\usepackage[T1]{fontenc} 
\usepackage{hyperref}       
\usepackage{graphicx}       
\usepackage{url}            
\usepackage{booktabs}       
\usepackage{amsfonts}       
\usepackage{natbib}
\usepackage{subcaption}
\usepackage{standalone}
\usepackage{amsmath}
\usepackage{siunitx}
\usepackage{import}
\usepackage[draft]{minted}
\usepackage{multirow}
\usepackage{wrapfig}
\usepackage{siunitx}

\newcommand{\codeblock}[1]{\texttt{#1}}

\author{Zijie Li}
\affiliation[MechE]{Department of Mechanical Engineering,
Carnegie Mellon University,
Pittsburgh PA, USA}
\author{Kazem Meidani}
\affiliation[MechE]{Department of Mechanical Engineering,
Carnegie Mellon University,
Pittsburgh PA, USA}
\author{Prakarsh Yadav}
\affiliation[MechE]{Department of Mechanical Engineering,
Carnegie Mellon University,
Pittsburgh PA, USA}

\author{Amir Barati Farimani}
\affiliation[MechE]{Department of Mechanical Engineering,
Carnegie Mellon University,
Pittsburgh PA, USA}
\alsoaffiliation[ML]{Machine Learning Department,
Carnegie Mellon University,
Pittsburgh PA, USA}
\alsoaffiliation[Chem]{Department of Chemical Engineering,
Carnegie Mellon University,
Pittsburgh PA, USA}
\email{barati@cmu.edu}

\title
  {Graph Neural Networks Accelerated Molecular Dynamics}


\begin{document}

\begin{abstract}

Molecular Dynamics (MD) simulation is a powerful tool for understanding the dynamics and structure of matter. Since the resolution of MD is atomic-scale, achieving long time-scale simulations with femtosecond integration is very expensive. In each MD step, numerous iterative computations are performed to calculate energy based on different types of interaction and their corresponding spatial gradients. These repetitive computations can be learned and surrogated by a deep learning model like a Graph Neural Network (GNN). In this work, we developed a GNN Accelerated Molecular Dynamics (GAMD) model that directly predicts forces given the state of the system (atom positions, atom types), bypassing the evaluation of potential energy. By training the GNN on a variety of data sources (simulation data derived from classical MD and density functional theory), we show that GAMD can predict the dynamics of two typical molecular systems, Lennard-Jones system and Water system, in the NVT ensemble with velocities regulated by a thermostat. We further show that GAMD's learning and inference are agnostic to the scale, where it can scale to much larger systems at test time. We also perform a comprehensive benchmark test comparing our implementation of GAMD to production-level MD softwares, showing GAMD's competitive performance on the large-scale simulation.
\end{abstract}

\section{Introduction}

Molecular Dynamics (MD) has been extensively used in a wide range of applications in material science, physical chemistry, and biophysics during the last decades \cite{MD-for-all-2018, Karplus2002-MD-biomolecules}. MD simulations can provide the understanding and precise predictions of many intricate systems. MD simulations rely on the knowledge of forces acting on every particle in the system and solving equations of motion to generate the trajectories of atoms in an iterative manner. Ab initio approaches like density functional theory (DFT) \cite{DFT-review-2014} are more accurate methods that consider the electronic structure of atoms. However, they are prohibitively time-expensive, especially for large many-body systems \cite{Unke-2020-PES-MLST}. Alternatively, forces can be calculated from empirical interatomic potentials for different environments bypassing the electronic structures in the system, resulting in faster simulations than ab initio methods\cite{review-FFs-2018}. Nevertheless, MD simulations have limitations including their long computational times \cite{Paquet2015-MD-time}, as well as the accuracy and generalizability of the empirical potentials \cite{Unke-2020-PES-MLST, review-FFs-2018}. We will expand upon these two aspects in the following paragraphs. 

The first major limitation of Molecular Dynamics is its computational intensity. Ensuring the stability of MD simulations with atomic-scale resolution requires integration time steps in the order of femtoseconds. This imposes a significant time constraint for many biochemical and macromolecular systems which have much longer time scales of dynamics (nano- or micro-seconds), such as simulating large scale systems like viruses for realistic time scales \cite{MD-for-all-2018, Paquet2015-MD-time, time-MD-2010}. Achieving these long time-scale simulations is compute-intensive, or even impossible in some cases, despite using state-of-the-art computational hardware, such as GPUs, and acceleration techniques. 

Each iteration of MD simulations requires a huge amount of computations including the calculation of forces acting on each particle summed up from various bonded or non-bonded potentials in the system \cite{review-FFs-2018}. Depending on the type and the scale of the simulations, they contain many repetitive computations especially in calculating forces for each atom as the negative gradient of the empirical potentials \cite{Chmiela-GDML-2017, hu-2021-forcenet}. Avoiding these time-consuming and iterative steps provides an opportunity to conduct more time-efficient MD simulations \cite{fast-covariant-force-NNFF-2019, Park2021-npj-direct-force, hu-2021-forcenet}. Another promising approach to reduce the amount of computation in complex systems has been through the coarse-grained force fields \cite{Coarse-GNN-2020-JCP}. However, while some thermodynamic consistency can be preserved by coarse-grained networks, there is still considerable information loss on structural details and properties of the systems \cite{CGNet-2019-Coarse,coarse-CGSchNet-2020}.

The second major limitation of classical MD simulations stems from the limitation of accuracy in calculations and approximations required to describe complex interatomic potentials between particles. There are a wide variety of interactions depending on the types of particles (bonded or non-bonded interactions), and the system composition. Finding the appropriate functional forms that satisfy the precision requirement is a challenging task \cite{review-FFs-2018, SpookyNet-2021}. Recently, machine learning (ML) has been viewed as a promising candidate for learning force fields \cite{review-Noe-2020, review-Barati-2020, unke2021-ML-Forcefield, ML-Forcefield-2017-Botu, ML-forcefield-Ab-initio-2017, ML-forcefield-nature-2018, ML-interatomic-2019, ml-potential-review, nnp-sodium, nnp-zinc, DeepPMD, ManySpecies-mlp}. 

In the last decade, advances in machine learning have enabled learning the dynamics of many complex physics-based systems from training data. Currently, there are active avenues of research with aim to learn models which simulate complex systems governed by well-known, partially-known, or even unknown underlying physics \cite{Sanchez-graph-GCN-ICML-2020, Bapst2020-glassy, Interaction-IN-NIPS-2016}. The learnt model can accurately approximate the true physical model, and in some cases, accelerate the simulation, prediction, and control of the systems by surrogating the full physics model \cite{hu-2021-forcenet}. In addition to the gain in time efficiency, proposed models have to perform well in terms of both accuracy (within the training data distribution) and generalization (beyond the data what the model is trained on). The capabilities of ML, especially neural networks, in learning arbitrary functional forms have made them a suitable choice for the task of learning interatomic potentials in MD \cite{Bartok-ML-modeling, ML-interatomic-2019,  review-Barati-2020, review-Noe-2020, GN-universal-molecules-2019}.

The initial approaches to use machine learning to learn force fields have leveraged feature engineering methods, by finding tailored descriptors that capture the characteristics of the local environment of atoms in the system \cite{Bartok-ML-modeling, ML-forcefield-Ab-initio-2017, ANI-1-Smith-2017, FCHL-revisited-2020, DPMD-2018, fingerprints-o2}. Atom-centered symmetry functions (ACSFs), used in Behler-Parrinello Neural Networks (BPNN) \cite{Behler-2011-ACSF, Behler-NN-Energy-2007, Behler-BPNN2017}, is one of the first examples of using such descriptors for translation- and rotation-invariant energy conservation. Recent advancements of deep neural networks, with automatic feature extraction capabilities, have made it possible to learn the atomic representations directly from the raw atomic coordinates and low-level atomic features, providing an alternative to the manually tailored atomic fingerprints \cite{Conv_fingerprints_NIPS-2015, Molecular-graph-2016, Schutt-insight-DTNN-nature, PhysNet-2019-Unke, schutt2017schnet, Hierarchical-2018-JCP, atomic-fingerprints-2018-ML}. Due to the unstructured position of particles and their limited interactions in an MD system, graph neural networks (GNNs) can be a suitable choice to approach this task \cite{Molecular-graph-2016, Park2021-npj-direct-force}. 

Graph Neural Networks (GNNs), with convolution layers that are specialized for unstructured data, have recently been used to learn various physics-based models and to accelerate them \cite{Mesh-based-GNN-ICLR-2021,GNN-particle-review-2021,LI2022,ogoke2020graph, brandstetter2022message, belbuteperes2020combining}. In molecular systems they have shown a performance boost as well as an end-to-end learning opportunity that bypasses the need for manual feature representations by learning the properties automatically from the atomic observations \cite{Conv_fingerprints_NIPS-2015, CGCNN-Grossman-2018, OGCNN-Karamad-2020, wang2021molclr}. GNNs encode information about particles (their properties and the neighbor interactions) in the nodes and edges of a defined graph. The neighbor interactions can be described in the general form of message passing neural networks (MPNNs) \cite{Gilmer-MP-ICML-2017} where information is passed between neighbor atoms, i.e. nodes, in the graph. 

A distinction between ML models for learning forces in MD simulations can be classified as energy-centric and force-centric models \cite{hu-2021-forcenet, Park2021-npj-direct-force}. Energy-centric models aim to learn the potential energy surface (PES) and obtain forces by calculating derivatives of the PES \cite{Chmiela-GDML-2017}. The majority of the currently proposed ML models belong to this group and directly conserve the system's energy due to their architecture and loss function design \cite{Molecular-graph-2016, PhysNet-2019-Unke}. Gradient-Domain Machine Learning (GDML) and symmetrized GDML (sGDML) \cite{Chmiela-GDML-2017, ML-forcefield-nature-2018} conserve energy by learning explicit gradient function mapping of energy and interatomic forces. Another energy-centric model, SchNet \cite{schutt2017schnet} employs continuous convolution filters to learn features on graph networks for smooth energy predictions. DimeNet \cite{Directional-MPNN-ICLR-2020, DimeNet++-directional-2020} is another more complex energy-centric model that, though slower than SchNet, shows better generalization to different molecules and configurations by having angular information in the feature updates. Several deep learning frameworks have been built based on differentiable physics to employ classical MD and machine learning potentials learned by these energy-centric models \cite{jaxmd2020, TorchMD-2021-JCTC, SchNetPack-JCTC-2019, wang-2020-differentiable}. 


In the force-centric group of models, forces are predicted directly by using networks where the forces predicted by the model, which are then compared against the ground truth forces. In other words, the network's outputs are per-node, i.e. per-atom, forces \cite{hu-2021-forcenet, fast-covariant-force-NNFF-2019, Park2021-npj-direct-force}. While these models do not possess any strict energy conservation mechanism, they have promising features. One challenge with the prediction of forces from PES has been the amplification of noise and error in taking derivatives of energy. This issue has been addressed to an extent by considering terms of both force and energy in the loss function \cite{PES-force-simultaneous-2009,schutt2017schnet, Chmiela-GDML-2017}. However, using a force-only training model can result in better force prediction accuracy \cite{OC20-Catalyst-2021}. The main benefit of using force-centric models is the computational efficiency brought about by avoiding additional calculations required for computing forces in the energy-centric models. These calculations include taking derivatives from the accumulation of several types of potential energies \cite{hu-2021-forcenet,Park2021-npj-direct-force}.  

Force-centric models have been applied in several applications to infer the forces in fluids \cite{Sanchez-graph-GCN-ICML-2020}, glassy systems, and solid-state MD simulations \cite{Bapst2020-glassy,fast-covariant-force-NNFF-2019,Park2021-npj-direct-force}, as well as large-scale quantum property calculations \cite{hu-2021-forcenet}. Bypassing the calculations required for taking derivatives of PES has provided gains in computational efficiency for neural network force field (NNFF) \cite{fast-covariant-force-NNFF-2019} and Graph NNFF (GNNFF) \cite{Park2021-npj-direct-force} models.
By using the DFT-calculated forces as ground truth, GNNFF can learn atomistic level dynamics of solid-state material systems. A force-centric model, ForceNet \cite{hu-2021-forcenet} uses Graph Network-based simulators (GNS) framework to predict forces by using a massive physics-based augmented dataset rather than applying any architectural constraints. These models show faster and accurate predictions of quantum properties and forces in comparison to energy-centric models like SchNet \cite{schutt2017schnet}. 

The energy- and force-centric models have been two lines of research to approach the aforementioned limitations of MD. On one hand, exploiting the power of deep neural networks and differentiable programming to automatically learn representations from the basic atomic features and coordinates has led to more generalizable potential models \cite{DeepPMD, jaxmd2020, TorchMD-2021-JCTC}. On the other hand, force-centric models avoid the computational bottleneck of taking derivatives and provide faster force predictions \cite{hu-2021-forcenet, Park2021-npj-direct-force}. In this work, we combine these two aspects in a Graph Accelerated Molecular Dynamics (GAMD) framework, where we leverage GNNs to learn atomic representations and directly predict per-atom forces from these learned features. We show that by careful training of a rotation-covariant graph model, we can bypass the physics-based neighbor selection and energy calculation steps of traditional MD simulations. In other words, GAMD avoids the computational bottleneck of calculating spatial derivatives from PES in long time scale molecular dynamics. Since GAMD does not derive the forces from PES, it does not conserve potential energy, but it can be used to run the simulation in an NVT ensemble where velocities are regulated by a thermostat of proper intensity. Our work is closely related to prior works on developing force-centric models - NNFF\cite{fast-covariant-force-NNFF-2019}, GNNFF \cite{Park2021-npj-direct-force} and ForceNet \cite{hu-2021-forcenet}. Similar to GNNFF and ForceNet, we use GNN as the backbone to build a molecular force model, however, with different architectures (message passing function and input edge features). Moreover, in this work we investigate the applicability and scalability of using a force-centric model to accelerate the dynamics on the large-scale molecular system such as water system \cite{Morawietz-DFT-NN-2013, Morawietz8368, ab-inito-water-2019-pnas} and conduct a detailed evaluation of the computational time efficiency.

\newpage
\section{Methods}
\subsection{Molecular Representation}
 A molecular system at a certain state can be described as: $\mathbf{X}:(\mathbf{q}_1, \mathbf{q}_2, \dots, \mathbf{q}_n)$, where $\mathbf{q}_i, \forall i \in \{1, 2, \dots, n\}$ denotes the Cartesian coordinates of each atom in the system. Through the network, we represent atoms as nodes in a GNN and interactions between atoms as edges.
Concretely, the node input feature is a one-hot vector $\mathbf{p}_i$, which specifies the atom type. The edge input feature is a vector derived by concatenating inter-atomic distance vector $\mathbf{q}_{ij}$ and a one-hot vector $\mathbf{b}_{ij}$ which indicates the edge type between two atoms (i.e. whether two atoms are bonded in the same molecule). The inter-atomic distance vector is defined by concatenating the directional vector and the norm of relative position: $\mathbf{q}_{ij}=(\frac{\mathbf{q}_{i}-\mathbf{q}_{j}}{||\mathbf{q}_{i}-\mathbf{q}_{j}||_2},||\mathbf{q}_{i}-\mathbf{q}_{j}||_2)$. With the introduction of directional vector, the edge is directed, therefore $\mathbf{e}_{ij}\neq\mathbf{e}_{ji}$. In addition, this makes the network rotational covariant as directional vector varies when the system rotates.

\begin{figure}[H]
    \centering
    \includegraphics[width=\textwidth]{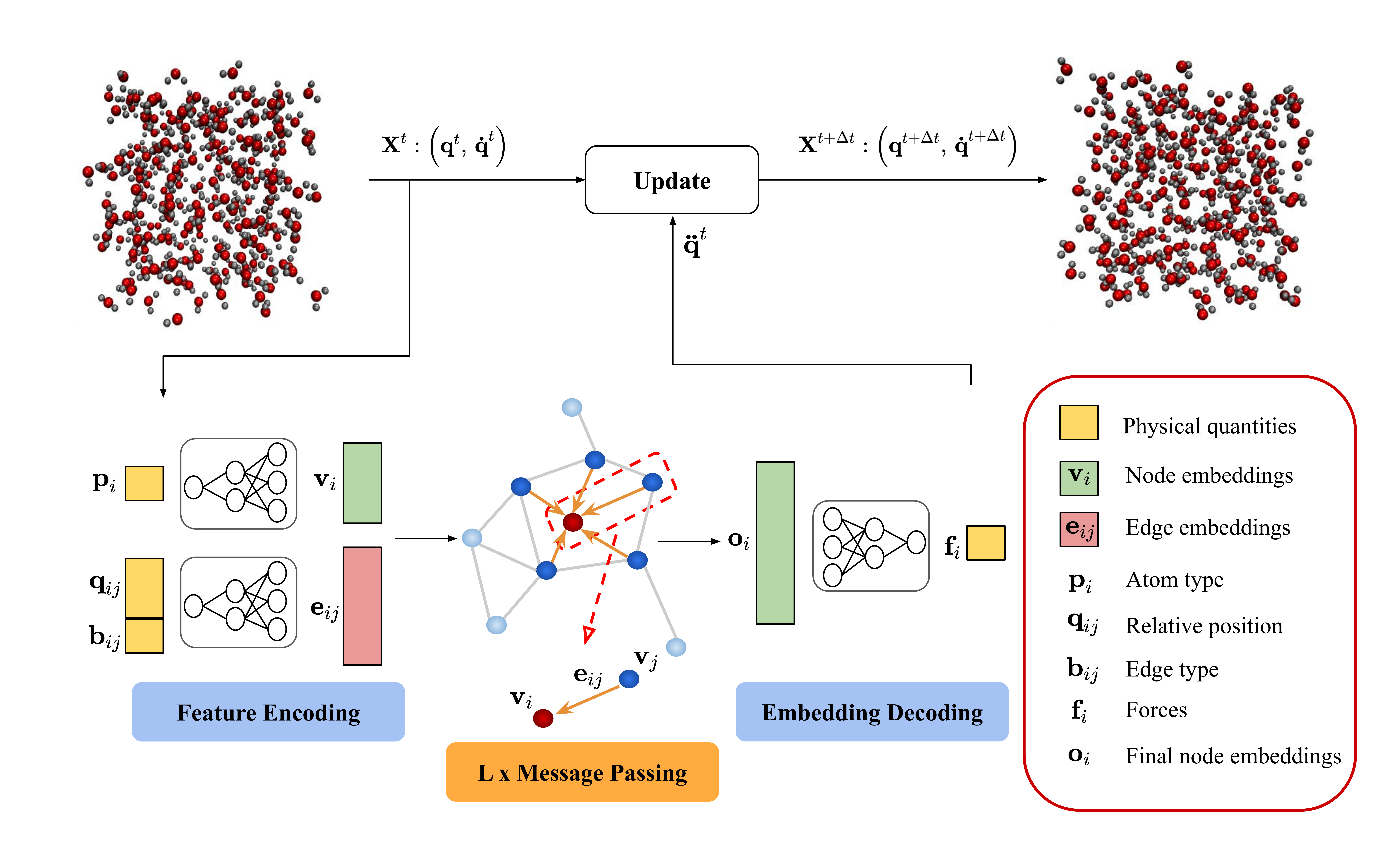}
    \vspace{-6mm}
    \caption{Overview of GAMD schematic. GAMD uses graph neural network (GNN) to predict the forces of atoms $\mathbf{f}_i$ in the molecular system $\forall i \in \{1, 2, \dots, n\}$ given properties of atoms $\mathbf{p}_i$ (i.e. type of atoms), positions of atoms $\mathbf{q}_i$, and edge type between each pair of atoms $\mathbf{b}_{ij}$ (bond information). The features are first encoded using multi-layer perceptron (MLPs) and then inferred using an atom-wise message passing scheme. At last, an MLP decoder predicts forces from the node embeddings. Based on the predicted forces, GAMD forwards the dynamics of the system by imposing acceleration $\mathbf{\ddot{q}}$ on each atom.}
    \label{fig:algorithm overview}
\end{figure}

 To further leverage the expressiveness of neural networks, we lift node input features $\mathbf{p}_i$ and edge input features $(\mathbf{q}_{ij}, \mathbf{b}_{ij})$ into high-dimensional vector embedding $\mathbf{v}^{(0)}_i, \mathbf{e}^{(0)}_{ij} \in \mathbb{R}^{d}$ (with $d$ being the dimension of latent space) via learnable encoders $\epsilon^{V}$ and $\epsilon^{E}$ (Figure~\ref{fig:algorithm overview}, Feature Encoding), which are built upon multi-layer perceptrons (MLPs).
 \begin{equation}
 \begin{split}
      \mathbf{v}^{(0)}_i &= \epsilon^{V}(\mathbf{p}_i) \\
     \mathbf{e}^{(0)}_{ij} &= \epsilon^{E}(\mathbf{q}_{ij}, \mathbf{b}_{ij})
 \end{split}
 \label{eq:embedding}
 \end{equation}
\subsection{Atom-wise Message Passing}
The inference and prediction of molecular information utilizes a recurring atom-wise message passing block (Figure~\ref{fig:algorithm overview}, Message Passing). Inside each block, the center atom $i$ first collects messages from all the neighbor atoms $\forall j \in \mathcal{N}(i)$. The message $\mathbf{m}^{(l)}_{j\rightarrow i}$ is conditioned on the edge embedding $\mathbf{e}^{(l-1)}_{ij}$ which carries the inter-atomic directional information and interaction type between two particles, along with node features from source atom $\mathbf{v}^{(l-1)}_j$ and target atom $\mathbf{v}^{(l-1)}_i$:
\begin{equation}
    \mathbf{m}^{(l)}_{j\rightarrow i} = \Phi^{(l)} \left(\mathbf{v}^{(l-1)}_j, \mathbf{e}^{(l-1)}_{ij}, \mathbf{v}^{(l-1)}_i \right) \odot \mathbf{v}^{(l-1)}_j, \forall j \in \mathcal{N}(i)
\end{equation}
Where $\Phi^{(l)}: \mathbb{R}^{d}\times\mathbb{R}^{d}\times\mathbb{R}^{d} \rightarrow \mathbb{R}^{d}$ is the learnable message function at the $l-th$ block, and $\odot$ denotes element-wise multiplication. 
The message passing mechanism adopted here can also be viewed as an extension of continuous convolution proposed in SchNet \cite{schutt2017schnet}, with a learnable filter conditioned not only on inter-atomic distances but also interaction types between atoms, and atom properties of source and target atoms.

 After all messages from neighbor atoms are collected, they are aggregated together and used to update the node embeddings.
\begin{equation}
        \mathbf{M}^{(l)}_i = \sum_{\forall j \in \mathcal{N}(i)} \mathbf{m}^{(l)}_{j\rightarrow i} 
\end{equation}
\begin{equation}
        \mathbf{\hat{v}}^{(l)}_i = \Theta^{(l)} \left(\mathbf{v}^{(l-1)}_i, \mathbf{M}^{(l)}_i \right)
\end{equation}
Where $\Theta^{(l)}: \mathbb{R}^{d}\times\mathbb{R}^{d} \rightarrow \mathbb{R}^{d}$ is the learnable node update function at the $l-th$ block.
We apply layer-normalization \cite{ba2016layer} to the node embedding before inputting into each message passing layer and use residual connection \cite{he2015deep} at every message passing layer.

We implement the message function $\Phi$,
and the node update function $\Theta$ as MLPs. Notice that, throughout the network, only the node embeddings are updated recursively, the edge embeddings at per-layer are derived via a learnable non-linear transformation $\mathbf{A}(\cdot)$ (which is also implemented as MLP) from the initial encoded edge embeddings: $e^{(l)}_{ij} = \mathbf{A}^{(l)}( e^{(0)}_{ij}$). In general, we find that disabling edge embedding's recursive update will not result in performance degradation, yet it can reduce the overall computational cost (ablation study on the influence of recursive edge embedding update can be found in Section 2 in the Supplementary Information).

\subsection{Graph Neural Force Predictor}
The learnable decoder $\gamma^V: \mathbb{R}^d \rightarrow \mathbb{R}^3$ decodes high-dimensional node embeddings $\mathbf{o}$ ($\mathbf{o}_i=\mathbf{v}_i^{(l)}, \forall i \in {1, 2, \dots, N}$) at the final layer into the Cartesian forces $\mathbf{f}$ (Figure~\ref{fig:algorithm overview}, Embedding Decoding). The predicted forces are then used to update the acceleration $\mathbf{\ddot{q}}$ of each atom. The GNN-based force predictor can be flexibly integrated into any numerical integrator that has a force-based scheme. For instance, a single step in the Velocity-Verlet with GNN-based force predictor $\mathbf{\mathcal{F}(\cdot)}$ can be written as:
\begin{align}
    \mathbf{\dot{q}}_{t+\Delta t/2} &= \mathbf{\dot{q}}_{t} +  \mathbf{\ddot{q}}_t  \Delta t /2 \\
    \mathbf{q}_{t+\Delta t} &= \mathbf{q}_{t} + \mathbf{\dot{q}}_{t+\Delta t/2}\:  \Delta t  \\
    \mathbf{\ddot{q}}_{t+\Delta t} &= \mathbf{\mathcal{F}}\left( \mathbf{p}, \mathbf{b}, \mathbf{q}_{t+\Delta t} \right)/\mathbf{m} \\
    \mathbf{\dot{q}}_{t+\Delta t} &= \mathbf{\dot{q}}_{t+\Delta t/2} +  \mathbf{\ddot{q}}_{t+\Delta t} \: \Delta t / 2 
\end{align}
where $\mathbf{p}: (\mathbf{p}_1, \dots, \mathbf{p}_N)$ denotes the atom type of every atom in the system, $\mathbf{q}: (\mathbf{q}_1, \dots, \mathbf{q}_N)$ denotes the Cartesian coordinates of every atom, and $\mathbf{b}: (\mathbf{b}_1, \dots, \mathbf{b}_{k})$ denotes the edge type between every atom and their neighbor atoms.
\section{Implementation Details}

\subsection{Software}
We implement our model using PyTorch \cite{PyTorch2019} ($1.7.1$), Deep Graph Library ($0.7.0$) \cite{wang2019dgl} and trainer using PyTorch-Lightning ($1.3.0$). The training data based on empirical force fields are generated using off-the-shelf simulators from OpenMM 7\cite{OpenMM}. We use the spatial partitioning module from JAX-MD \cite{jaxmd2020} to maintain the neighbor list of particles in the system.
\subsection{Fixed radius graph}
In GAMD, the graph is constructed via fixed radius neighbor search, where edges are established between every pair of atoms, $(i, j)$, such that $||\mathbf{q}_i-\mathbf{q}_j||_2<r_0$. The naive way to perform this search is by calculating the distance for every pair of particles in the system, which results in $\mathcal{O}(N^2)$ complexity. This strongly limits the scalability and efficiency of the framework. To alleviate this computational overhead, we use a cell list to search for neighbors. We first partition the space into different cells with size $r_0$, and then for each particle, we only search for neighbors among particles within the same cell and adjacent cells. Here we use \codeblock{jax\_md.partition.cell\_list} to partition the space and \codeblock{jax\_md.partition.neighbor\_list} to gather the neighbor list, which has an overall complexity of $\mathcal{O}(N \log N)$.

For all the systems, the cutoff radius $r_0$ was chosen, such that an atom has roughly 20 neighbors on average. This encompasses the range of many types of local interactions between particles, while other long-range interactions can be captured by recursive message passing in the deeper layers. In general, the choice of cutoff radius in GAMD is agnostic to the underlying interaction range of the system, which eliminates the need for selecting neighbors based on different physics rules.

\subsection{Training and dataset}
\paragraph{Dataset generation}
The datasets used in this work are generated from classical molecular dynamics (MD) and density functional theory (DFT).

The classical MD (Lennard Jones\citep{Lennard_Jones_1931}, TIP3P\citep{TIP3P}, TIP4P-Ew\citep{TIP4PEw}) simulations are performed using OpenMM \cite{OpenMM}. The systems considered are uniform single atom or single molecule systems with different simulation box sizes. Periodic Boundary Conditions (PBC) are set in all directions and a cut-off distance of $10 \si{\angstrom}$ was used. Simulations are static in the initial step and pass through a transient state to reach equilibrium under constant volume and temperature (NVT), i.e. canonical ensemble. Velocity verlet integrator along with a Nos\'{e}--Hoover chain thermostat \cite{Nose,Hoover} with a collision frequency of $1.0 / \text{ps}$ and chain length of 10 are used to maintain the constant temperature. We simulate each configuration of molecular system to 50, 000 steps with a time step size of $2.0$ femtoseconds, and store the state of the system every 50 steps. Each configuration is generated by initializing particles in the system with random positions/velocities.

The DFT simulation data of water molecules are obtained from \citet{Morawietz8368}, which were calculated based on revised Perdew–Burke–Ernzerhof
(RPBE)\citep{RPBE} functional with Van der Waals correction using D3 method\citep{D3correction}.

\paragraph{Loss function}

The network is trained in a supervised way by minimizing the L1 distance between the predictions of per-atom forces $\hat{\mathbf{f}}_i$ and ground truth $\mathbf{f}_i$, along with a regularization term that penalizes the total sum of the forces in the system:
\begin{equation}
    \mathcal{L} = \frac{1}{N}\sum_{i=1}^N ||\mathbf{f}_i-\hat{\mathbf{f}}_i|| + \lambda  ||\frac{1}{N}\sum_{i=1}^N \hat{\mathbf{f}}_i||
\end{equation}

In GAMD, there is no hard-coded mechanism to restrict the range of messages a node can receive, and a node in the graph can receive messages from very distant nodes via recursive message passing. The sum of messages of a node in the final layer is essentially the high-dimensional embedding of the per-particle forces. Therefore, when the embedding contains redundant far-away messages, the final force prediction will also be influenced by unnecessary long-range interaction. To encourage the network to learn the minimal message-passing range required for predicting force accurately, we impose L1 regularization on the sum of the per-atom forces.
\begin{figure}[h]
    \centering
    \includegraphics[width=0.5\linewidth]{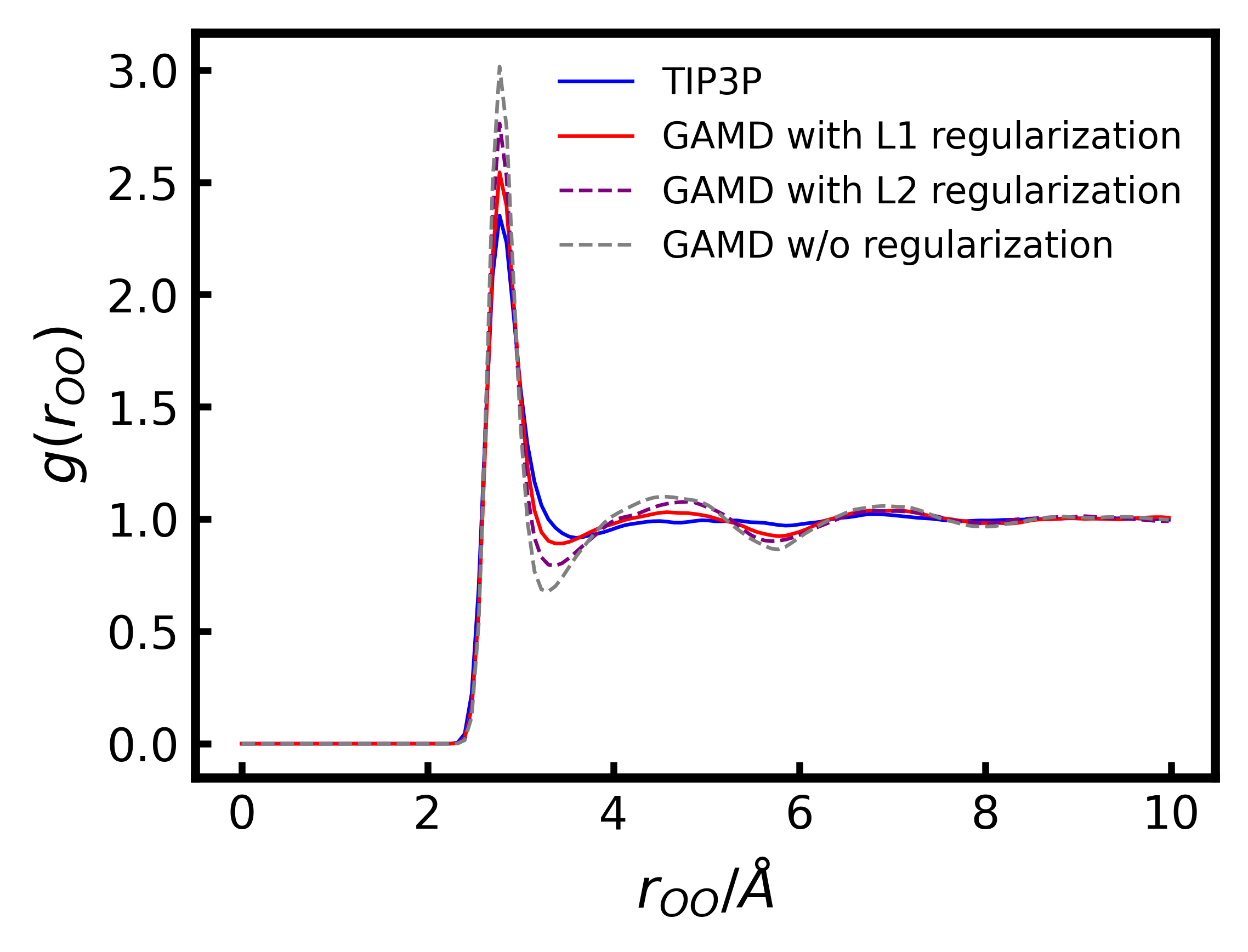}
    \caption{Comparison of models trained under different conditions. We measure the Oxygen-Oxygen RDF of water molecules' trajectories generated from OpenMM's TIP3P model and GAMD models trained using: 1.L1 regularization; 2.L2 regularization; 3.Without regularization. }
    \label{fig:reg vs no-reg}
\end{figure}
As shown in Figure \ref{fig:reg vs no-reg}, the radial distribution functions (RDF) of the models trained with L2 regularization and without regularization have much sharper peaks and larger gaps compared to the ground truth, while there is a smaller gap between the RDF in the ground truth and GAMD trained under L1 regularization. This indicates that with L1 regularization the predicted forces have greater agreement with the ground truth and molecules are more uniformly distributed. Also, it demonstrates that imposing L1 regularization can effectively suppress unnecessary long-range message passing especially when there are multiple types of interaction with different influence ranges in the system.
\paragraph{Training strategy}
We randomly split all datasets into train/test set with $90/10$ ratio. For the data generated from classical MD methods, the whole dataset contains $10, 000$ snapshots of input (i.e. positions) and target (i.e. forces), where $9, 000$ are used for training. For RPBE-D3 data based on DFT calculation, the whole dataset contains $6, 518$ train snapshots and $723$ test snapshots. 

Before calculating the loss, we normalize the ground truth forces, such that it has a zero mean and unit variance.  We optimize the model using the Adam optimizer \cite{kingma2017adam}, with an exponential learning rate scheduler that diminishes learning rate from $\alpha=3\times10^{-4}$ to $\alpha=1\times10^{-7}$. For data derived from classical MD, we train the model for 300k gradient updates. For DFT-based data, as its configurations are more diverse and complex than classical MD generated data, we train the model for 650k gradient updates. 

\paragraph{Model implementation}
We implement all the learnable functions described in the previous section as MLPs with three layers, using Gaussian Error Linear Units (GELUs)\cite{hendrycks2020gaussian} as non-linear activation function. For GAMD trained on DFT data, the embedding size of the network is 256, and the number of message passing layers is 5.  For GAMD trained on MD data, the embedding size is 128, and the number of message passing layers is 4. On the DFT dataset, we do not use bond information as an edge feature as it is not available in the dataset. Following \citet{schutt2017schnet}, we use Gaussian radial basis function to expand the interatomic distance before inputting into the edge encoder (Equation~\eqref{eq:embedding}). We found this resulted in a minor performance change on the investigated molecular systems. More details on the model's architecture and ablation study on architectural choices can be found in the Supplementary Information.

\subsection{Benchmark setting}
To compare GAMD's performance with other classical force calculation methods, we simulate the water system with different sizes using GAMD, OpenMM, and LAMMPS. As GAMD only modifies the force evaluation part of every step in the MD simulation, we exclude the running time of other calculations (e.g. chain propagation, position and velocity update) in the benchmark. All the benchmarks are run on a platform equipped with a single GTX-1080 Ti GPU and i7-8700k CPU. The benchmark results and discussion are presented in Section \ref{sec: benchmark}. Below we provide detailed configurations of benchmark for each package.
\paragraph{GAMD} Each force calculation step of GAMD comprises two parts, building the fixed radius graph and inference. Building the graph comprises updating the neighbor list, transforming the neighbor list into the adjacency matrix, and calculating the edge features between every pair of connected nodes (atoms). Note that in GAMD the rebuilding of the neighbor list and graph structure is performed when particles have traveled a distance larger than a threshold (heuristically we select this threshold as $d_r = r_{\text{cutoff}} / 10$). We conduct a benchmark using GAMD with 4 message passing layers and an embedding size of 128, which contains $\approx$ 650k parameters.
\paragraph{OpenMM} To exclude the influence of other calculations involved in a single step of update, we run two simulations for each benchmark in the OpenMM. The first simulation runs in normal mode and in the second simulation, we remove all the forces defined in the system and run a dummy simulation with no forces being calculated. Then we estimate the time used to calculate forces in every step by: $t = t_0 - t_{\text{dummy}}$, where $t_0$ denotes the time a normal simulation will take and $t_{\text{dummy}}$ denotes the time of dummy simulation.
\paragraph{LAMMPS}
In LAMMPS, the force calculation in an update step mainly consists of four parts according to the description in the official document: \textit{Pair}, \textit{Bond}, \textit{Kspace} and \textit{Neigh}, where \textit{Pair} denotes the evaluation of non-bonded forces, \textit{Bond} denotes the evaluation of bonded interactions, \textit{Kspace} denotes the evaluation of long-range interactions and \textit{Neigh} denotes the neighbor list construction. We report the total time of \textit{Pair}, \textit{Bond} and \textit{Kspace} at each step as the time used for force evaluation, and report \textit{Neigh} as time used for searching neighbors. Visual Molecular Dynamics (VMD) \cite{VMD} is used to create water box system for LAMMPS simulation.

\section{Results and discussion}
In this section, we present the results of GAMD on two common MD simulation systems - Lennard-Jones particles, and water molecules. We calculate the radial distribution function (RDF) to measure the spatial distribution of particles in each of the systems. The RDF $g(r)$ between two types of particles $A$ and $B$ is defined as:
\begin{equation}
    g(r) = \frac{1}{N_{A} N_{B}} \sum_{i=1}^{N_A} \sum_{j=1}^{N_B}
            \langle \delta(||\mathbf{r}_i - \mathbf{r}_j|| - r) \rangle
\end{equation}
where $\langle \cdot \rangle$ denotes the ensemble average, $||\mathbf{r}_i - \mathbf{r}_j||$ denotes the Euclidean distance between two particles $i, j$, $r$ is the radius of the corresponding spherical shell, $N_A$ and $N_B$ denote the number of corresponding types of particles and $\delta$ filters particles' distance not falling into this shell (which we implement as a Gaussian function). Furthermore, we validate the correctness of force prediction by comparing its angle and magnitude against ground truth.
\subsection{Lennard-Jones system}
\label{sec: LJ exp}
\begin{figure}[H]
    \begin{subfigure}{0.40\linewidth}
    \includegraphics[width=\linewidth]{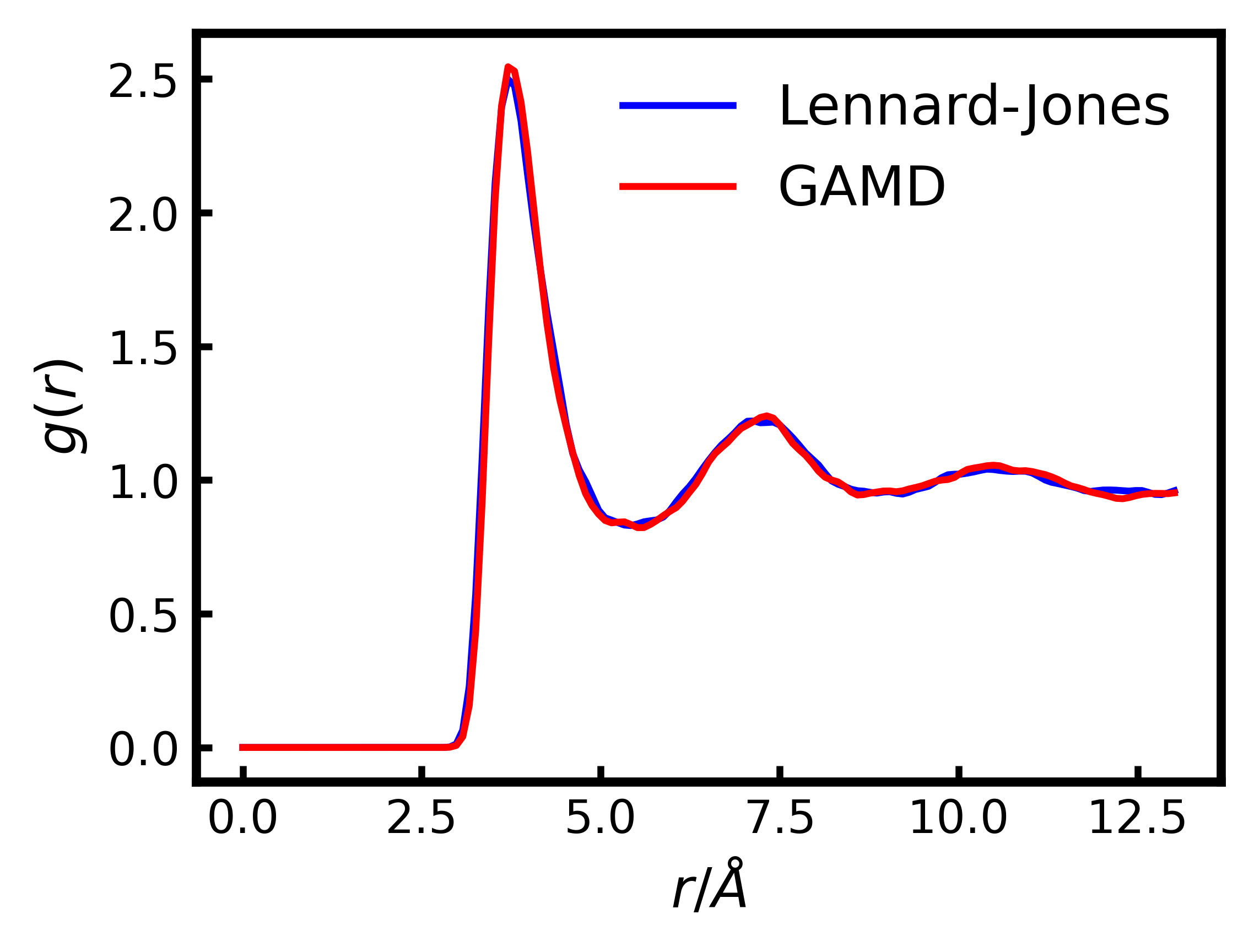}
    \captionsetup{justification=centering}
    \caption{RDF for particles in the Lennard-Jones system.}
       \label{fig:Argon LJ RDF}
    \end{subfigure}
    \begin{subfigure}{0.40\linewidth}
    \includegraphics[width=\linewidth]{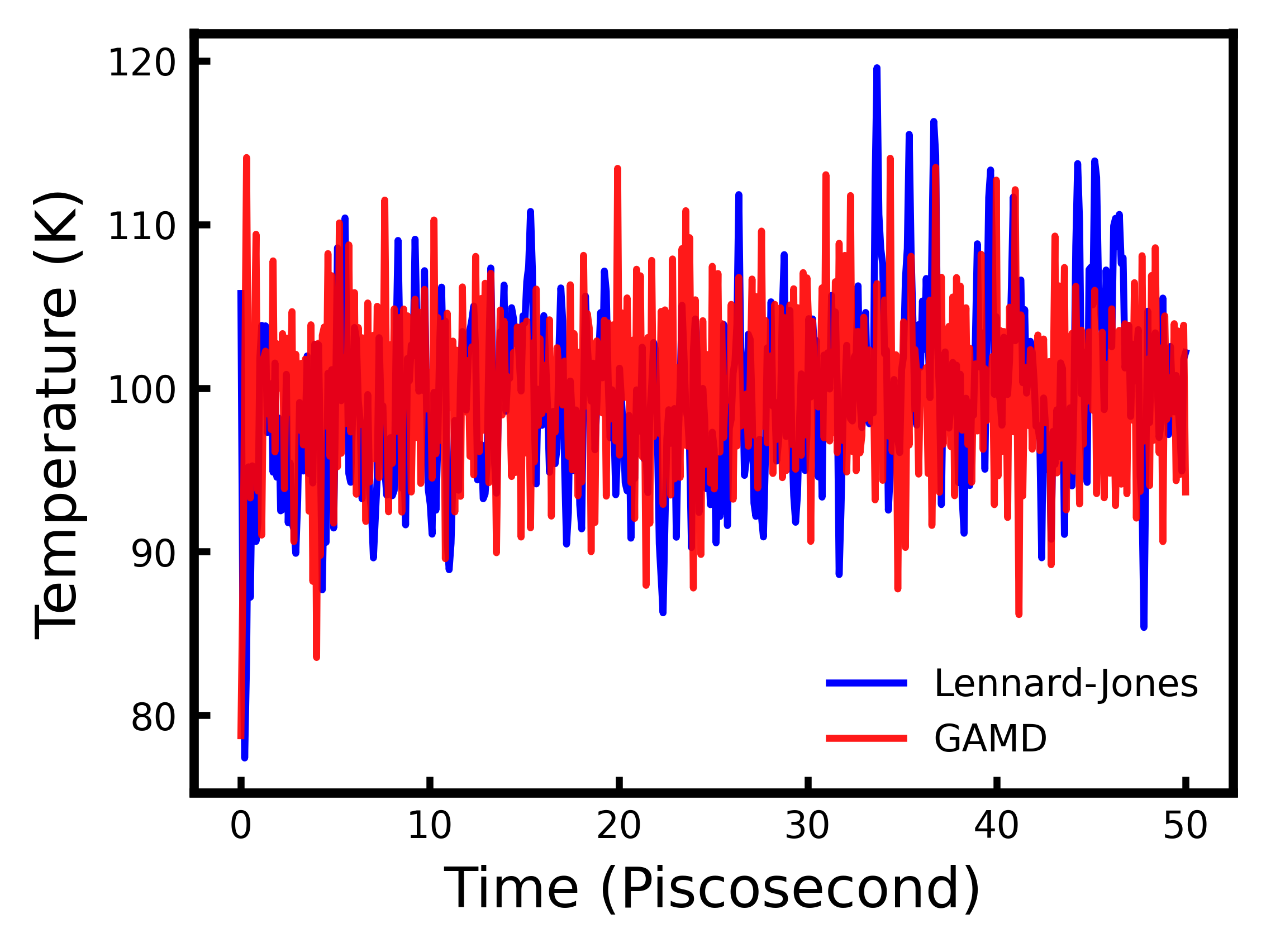}
    \captionsetup{justification=centering}
    \caption{Temperature trend of NVT ensemble in the Lennard-Jones system.}
    \label{fig:Argon LJ KE}
    \end{subfigure}
    \caption{Comparison of GAMD generated trajectory with the ground truth Lennard-Jones system.}
\end{figure}

We first investigate GAMD on a toy system consisting of liquid argon with non-bonded interatomic forces governed by Lennard-Jones (LJ) potential. This system comprises 258 argon atoms at 100 K and the non-bonded Van der Waals (VdW) potentials are approximated by LJ potentials. To evaluate the performance of GAMD in the prediction of interatomic LJ forces, we first compare the RDF of trajectories simulated using LJ potentials and the trained GAMD model (Figure~\ref{fig:Argon LJ RDF}). This shows how the GAMD simulated trajectories preserve the spatial behavior of the classical MD's trajectories. 



Since the temperature is held constant in the NVT ensemble simulation, we can evaluate the temporal behavior of the results by comparing the temperature of the GAMD simulated system with the ground truth. Both of the simulations are initialized by sampling velocities from Boltzmann distribution with the temperature set to 100K, and they reach equilibrium after a few transient steps under the regulation of Nos\'{e}--Hoover chain thermostat with collision frequency: $25.0/ps$. Under the regulation by a thermostat, the trained model can successfully equilibrate to the target temperature (Figure~\ref{fig:Argon LJ KE}). 

\begin{figure}[H]
    \centering
    \includegraphics[width=0.35\linewidth]{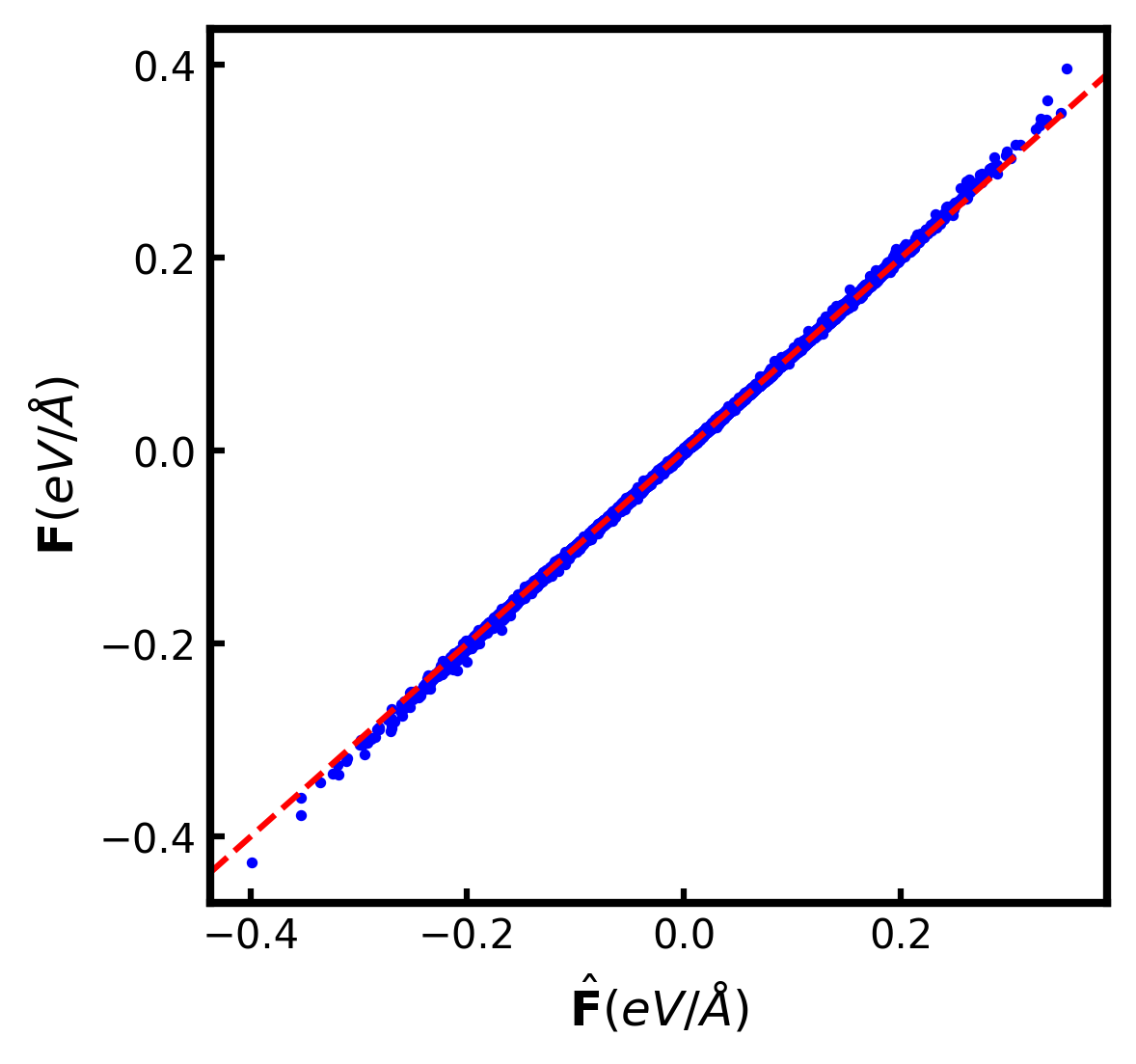}
    \captionsetup{justification=centering}
    \caption{Forces derived from Lennard-Jones potential versus GAMD predicted forces. $\mathbf{F}$ denotes ground truth forces, $\hat{\mathbf{F}}$ denotes GAMD's prediction.}
    \label{fig:force angle LJ}
\end{figure}

The rotation-covariant predicted forces are also compared directly in terms of both direction and magnitude with the ground truth LJ forces calculated in the MD simulations. By comparing $\cos \left<\hat{\mathbf{F}},\mathbf{F}\right>$ between the predicted and ground truth forces, we observe that more than $95 \%$ of the predicted forces have $\cos \left<\hat{\mathbf{F}},\mathbf{F}\right> > 0.995$. The mean absolute error (MAE), root mean squared error (RMSE), relative error, and force direction's agreement on the test set are reported in Table~\ref{Tab: lennard jones force}. The relative error is evaluated as the ratio of mean absolute error to the mean L2 norm of the ground truth forces $\mathbf{F}$: $<\left|\mathbf{F} - \hat{\mathbf{F}} \right|>/<\left||\mathbf{F}|\right|_2>$. Figure~\ref{fig:force angle LJ} describes the agreement between the predicted forces and the ground truth LJ forces.
\begin{table}
\centering
\scalebox{0.87}{
\begin{tabular}{cccccc} 
\toprule
Ground Truth & Test snapshots & $\text{MAE}$ ($meV/ \si{\angstrom}$) & $\text{RMSE}$ ($meV/ \si{\angstrom}$) &$\cos<\hat{\mathbf{F}},\mathbf{F}>$ & Relative error \\ 
\midrule
Lennard Jones~\citep{Lennard_Jones_1931} & 1000 & 0.266 $\pm$ 0.030  & 0.427 $\pm$ 0.119 & 0.997 & 0.61\%  $\pm$ 0.07\% \\
\bottomrule
\end{tabular}}
\caption{Quantitative analysis of force accuracy. The standard deviation is measured by comparing the statistics within a single snapshot to the whole test dataset.}
\label{Tab: lennard jones force}
\end{table}

\subsection{Water system}
\label{sec: water exp}
In the second experiment, we apply GAMD on a system with water molecules, where interactions between particles are more complex and more types of particles are involved in the dynamics. Water molecules are ubiquitous solvents in a variety of molecular systems and their dynamic properties are of great importance. There is a wide array of works employing machine learning methods, especially neural networks to parametrize the potential energy of water molecules and study their interaction\citep{Morawietz-DFT-NN-2013, Morawietz-NN-waterPES, Morawietz8368, ab-inito-water-2019-pnas, Morawietz-NN-waterdimer, bingqing-quantum-water}. In this work, we train and investigate GAMD using data derived from three different force models for water molecules - TIP3P \cite{TIP3P}, TIP4P-Ew \cite{TIP4PEw} and DFT calculation based on revised Perdew–Burke–Ernzerhof functional (RPBE)\cite{RPBE, Morawietz8368}. Note that when training on the four-site model's data, GAMD still adopts a three-site setting, where the forces of fictitious sites are not used. In these experiments, the model should be able to learn not only the non-bonded Van der Waals forces but also electrostatic forces due to the charges assigned to different types of atoms in a water molecule. Moreover, the model should handle two types of atoms, defined as node features, with different potentials and charges along with bonded and non-bonded interactions in and between molecules. 

\begin{figure}[H]
\begin{subfigure}{0.32\linewidth}
\includegraphics[width=\linewidth]{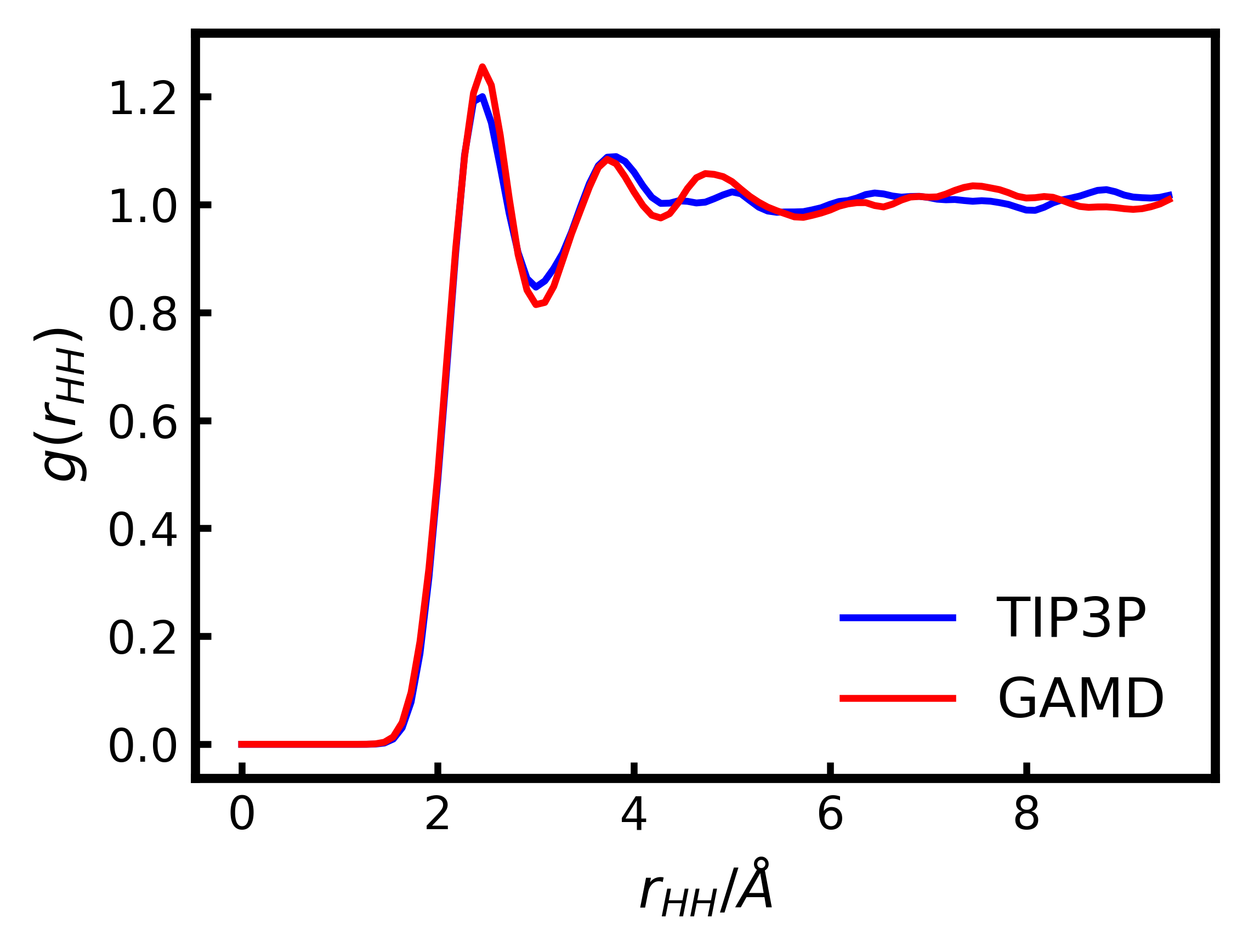}
\caption{\small{Hydrogen-Hydrogen}}
\label{fig:TIP3P RDF rhh}
\end{subfigure}
\begin{subfigure}{0.32\linewidth}
\includegraphics[width=\linewidth]{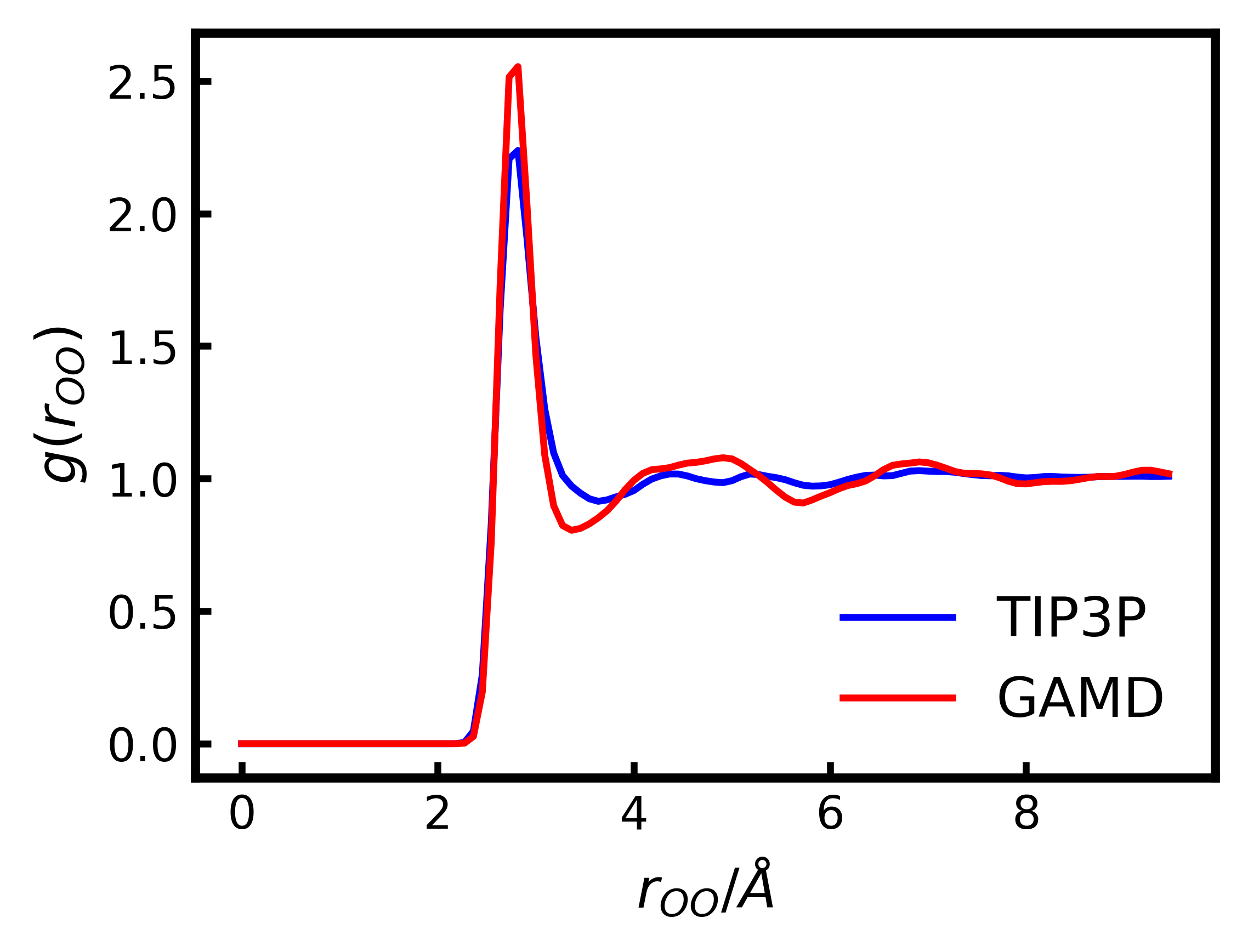}
\caption{\small{Oxygen-Oxygen}}
\label{fig:TIP3P RDF roo}
\end{subfigure}
\begin{subfigure}{0.32\linewidth}
\includegraphics[width=\linewidth]{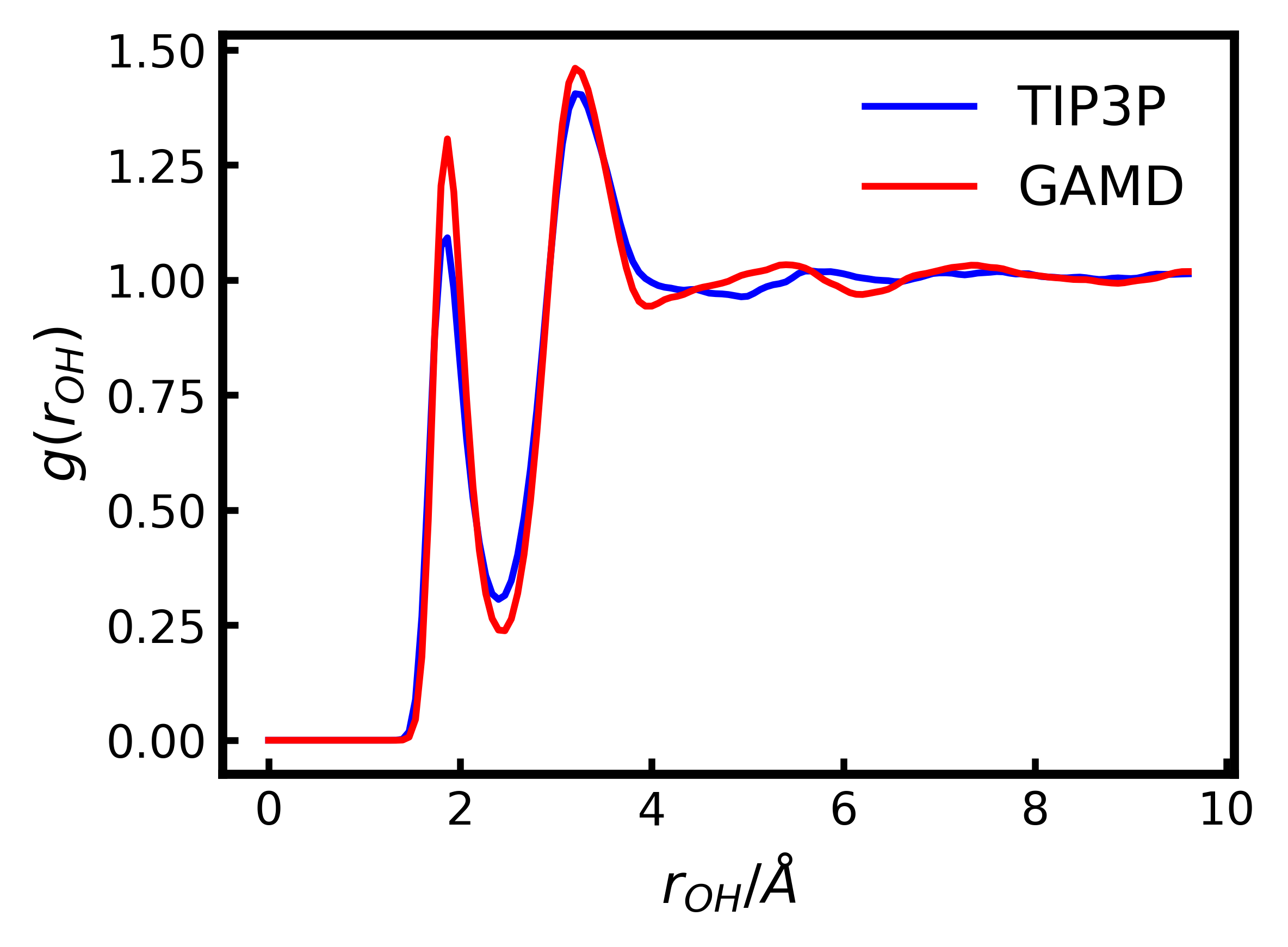}
\caption{\small{Oxygen-Hydrogen}}
\label{fig:TIP3P RDF roh}
\end{subfigure}
\captionsetup{justification=centering}
\caption{Radial distribution function of GAMD (trained on TIP3P\citep{TIP3P} data).}
\label{fig:tip3p rdf}
\end{figure}

\begin{figure}[H]
\begin{subfigure}{0.32\linewidth}
\includegraphics[width=\linewidth]{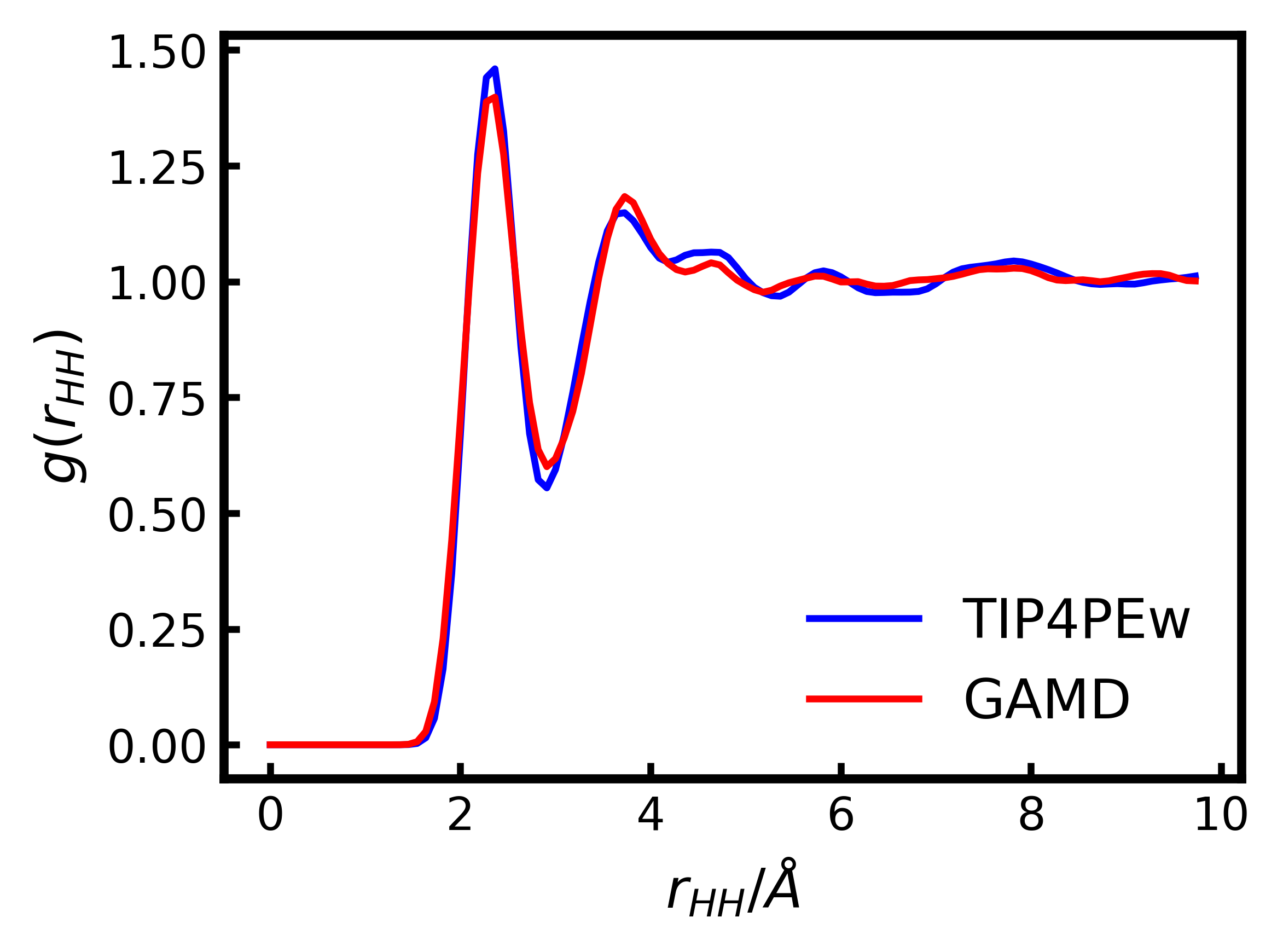}
\caption{Hydrogen-Hydrogen}
\label{fig:TIP4P RDF rhh}
\end{subfigure}
\begin{subfigure}{0.32\linewidth}
\includegraphics[width=\linewidth]{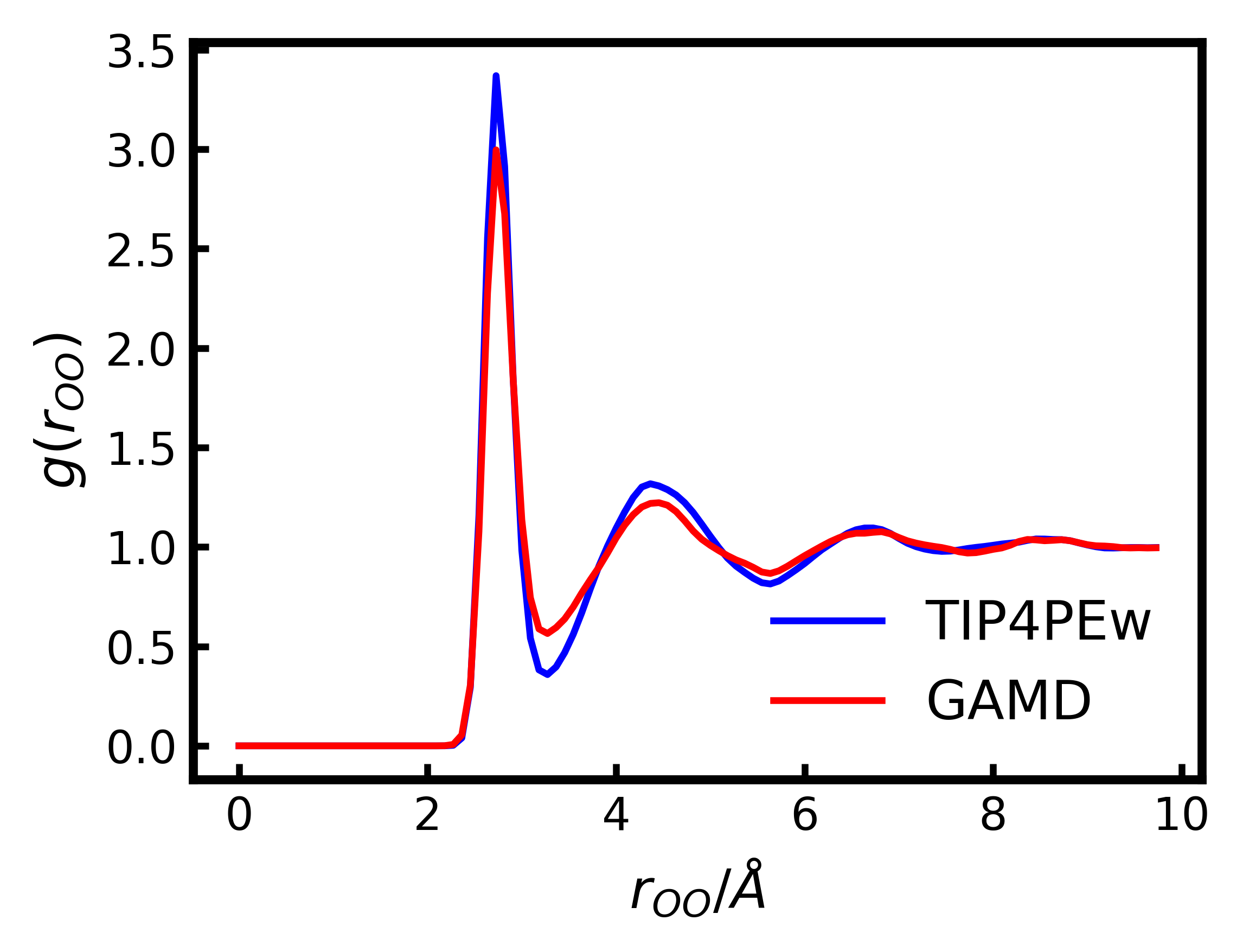}
\caption{Oxygen-Oxygen}
\label{fig:TIP4P RDF roo}
\end{subfigure}
\begin{subfigure}{0.32\linewidth}
\includegraphics[width=\linewidth]{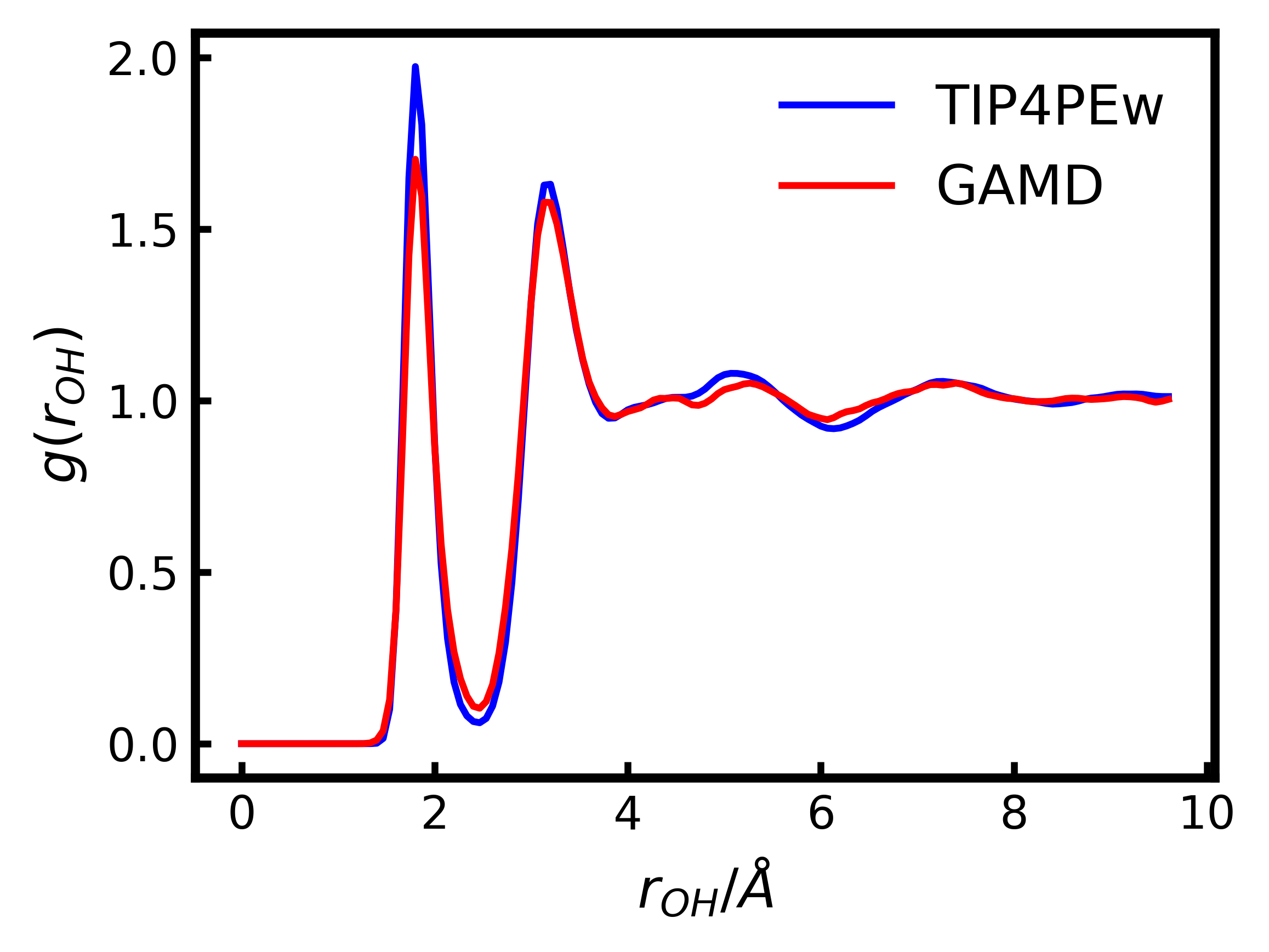}
\caption{Oxygen-Hydrogen}
\label{fig:TIP4P RDF roh}
\end{subfigure}
\captionsetup{justification=centering}
\caption{Radial distribution function of GAMD (trained on TIP4P-Ew\citep{TIP4PEw} data). }
\label{fig:tip4p rdf}
\end{figure}

\begin{figure}[H]
\begin{subfigure}{0.32\linewidth}
\includegraphics[width=\linewidth]{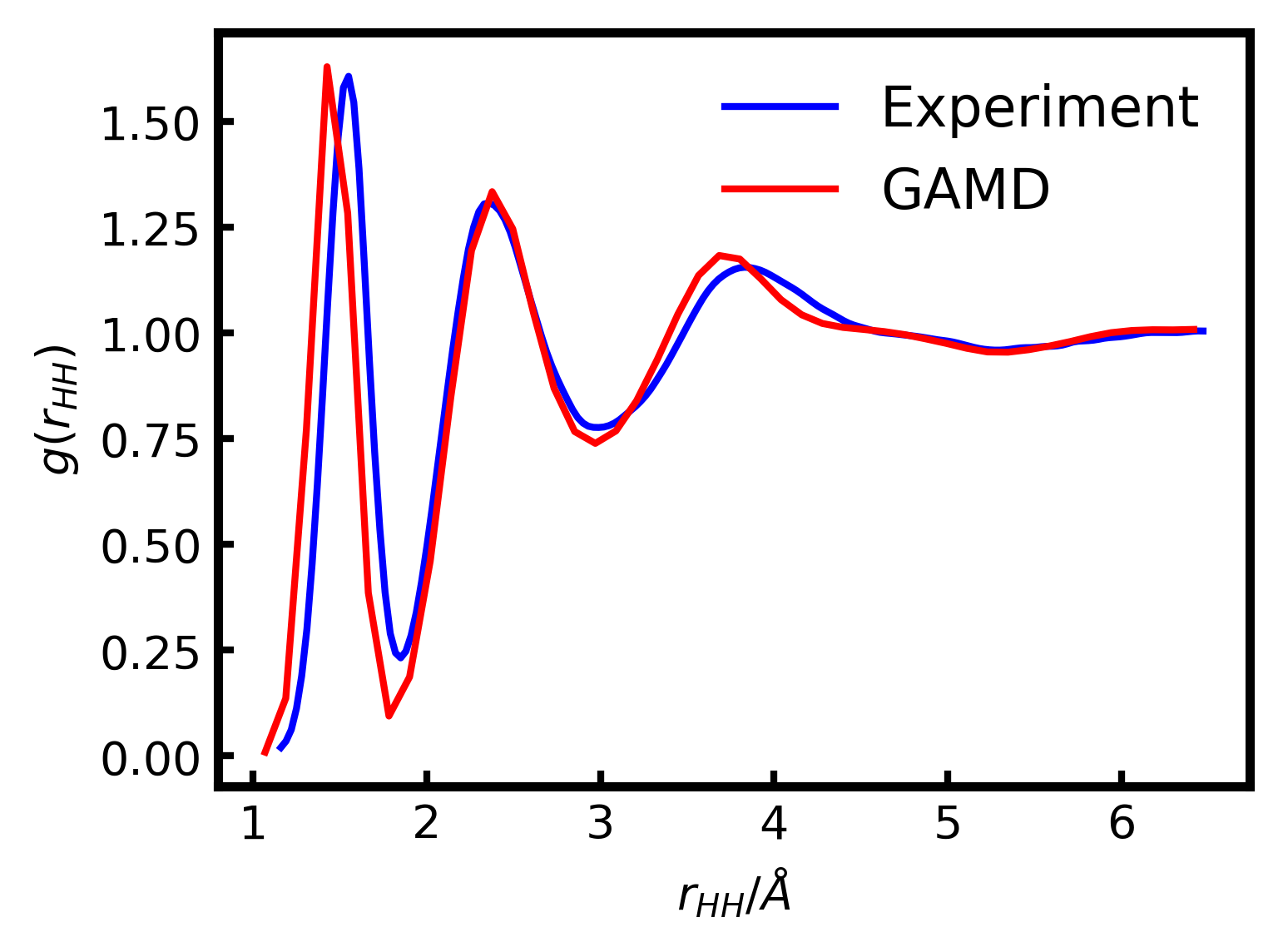}
\caption{Hydrogen-Hydrogen}
\label{fig:DFT RDF rhh}
\end{subfigure}
\begin{subfigure}{0.32\linewidth}
\includegraphics[width=\linewidth]{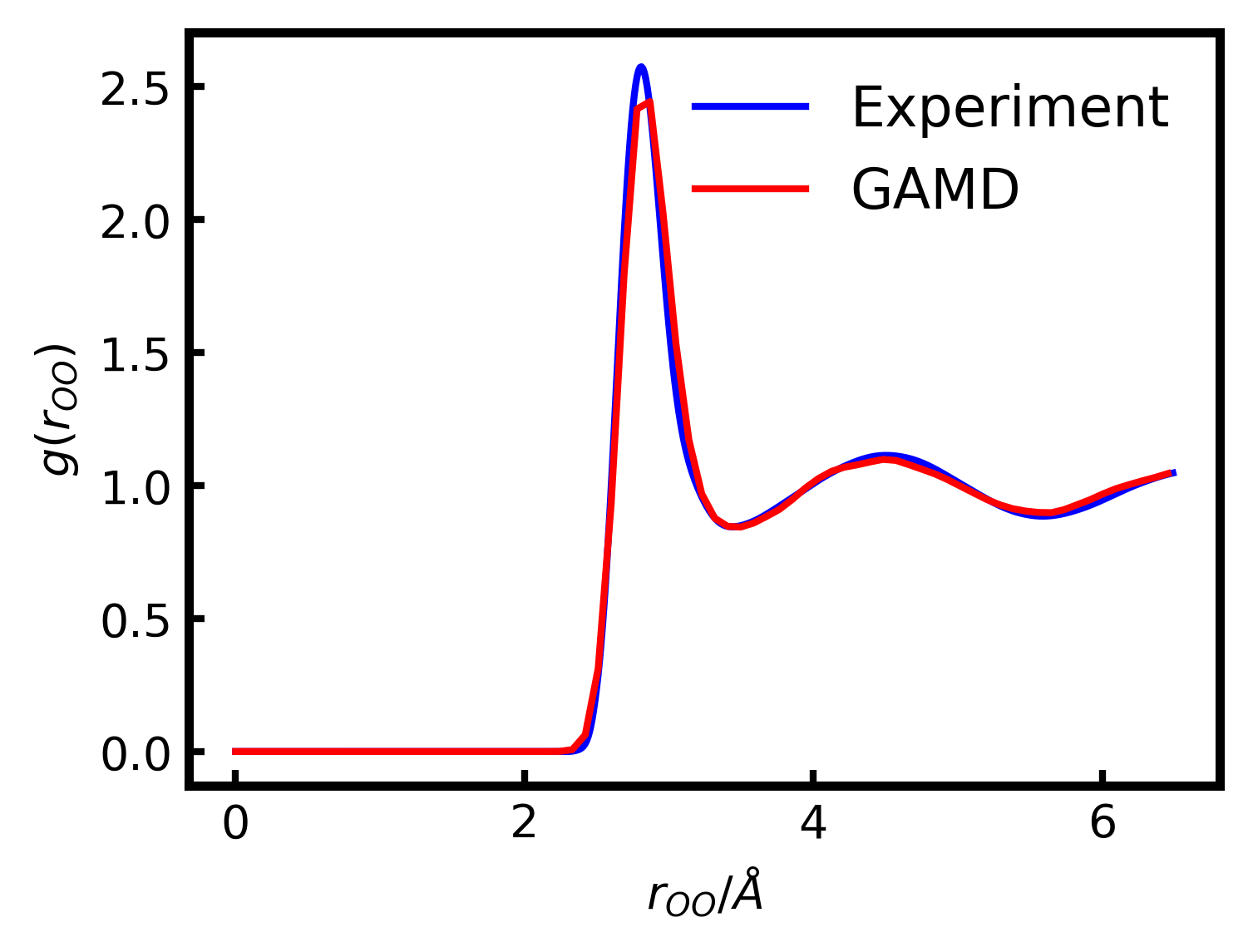}
\caption{Oxygen-Oxygen}
\label{fig:DFT RDF roo}
\end{subfigure}
\begin{subfigure}{0.32\linewidth}
\includegraphics[width=\linewidth]{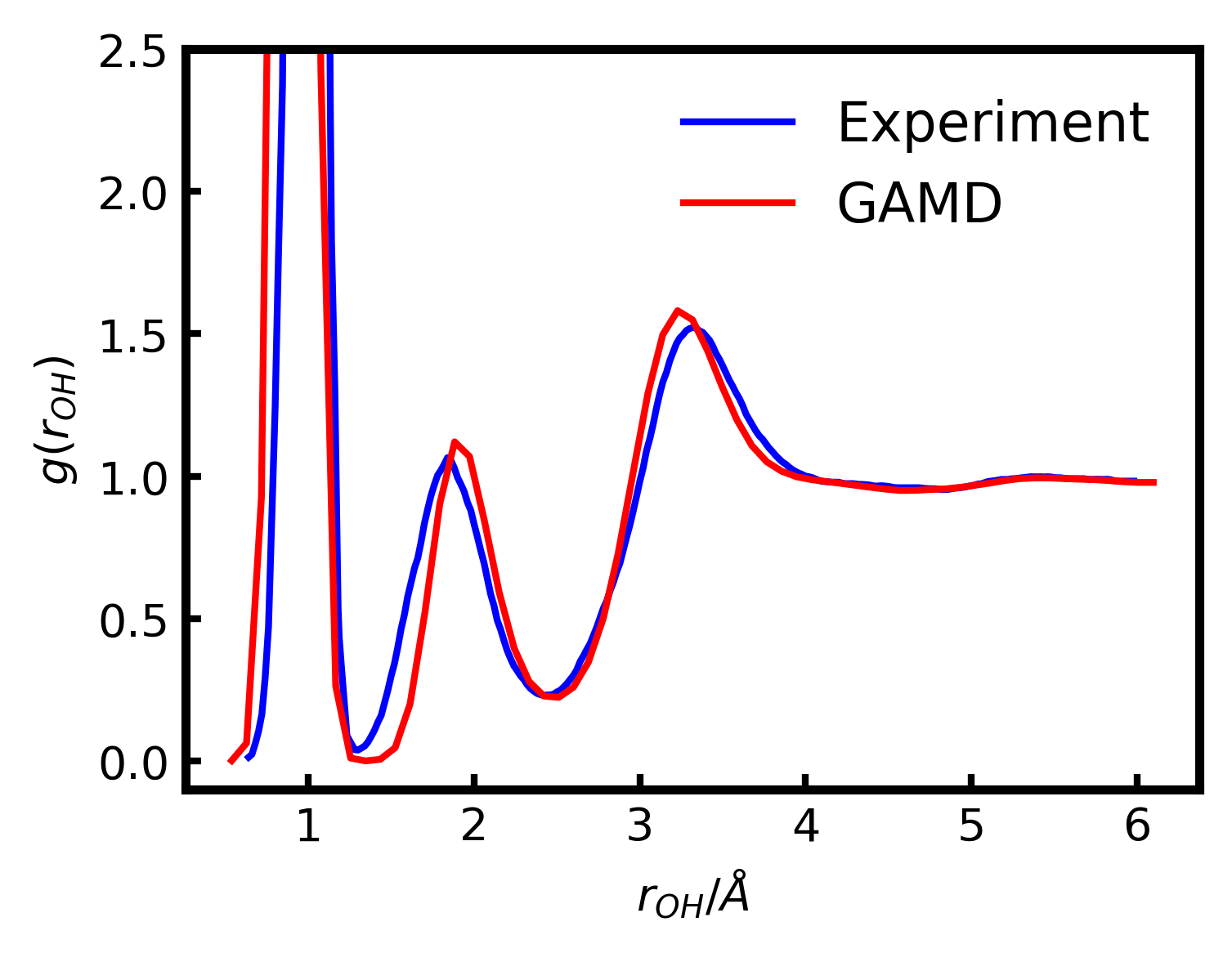}
\caption{Oxygen-Hydrogen}
\label{fig:DFT RDF roh}
\end{subfigure}
\caption{Radial distribution function of GAMD (trained on RPBE-D3\citep{Morawietz8368, RPBE, D3correction} data). $X$-axis is truncated according to the corresponding experimental data. The experimental RDF data collection is taken from \citet{rdfdata}. The original Hydrogen-Hydrogen RDF is from \citet{h-h-rdf}, Oxygen-Oxygen RDF is from \citet{o-o-rdf} and Oxygen-Hydrogen RDF is from \citet{o-h-rdf}.}
\label{fig:dft rdf}
\end{figure}

\begin{figure}[H]
    \centering
    \begin{subfigure}{0.32\textwidth}
    \centering
    \includegraphics[width=\linewidth]{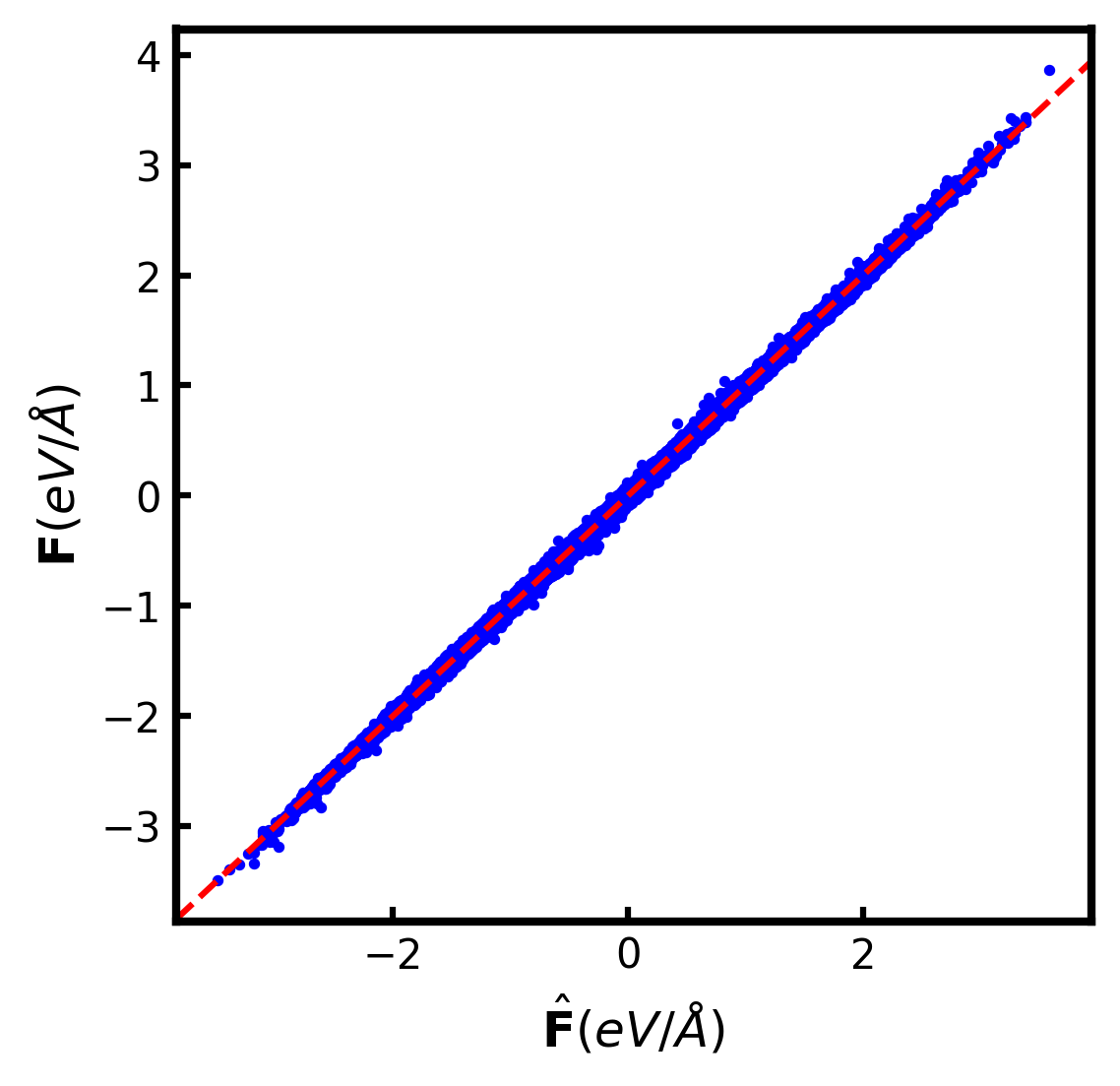}
    \captionsetup{justification=centering}
    \caption{GAMD trained on\\ TIP3P}
    \end{subfigure}
     \begin{subfigure}{0.32\textwidth}
    \centering
    \includegraphics[width=\linewidth]{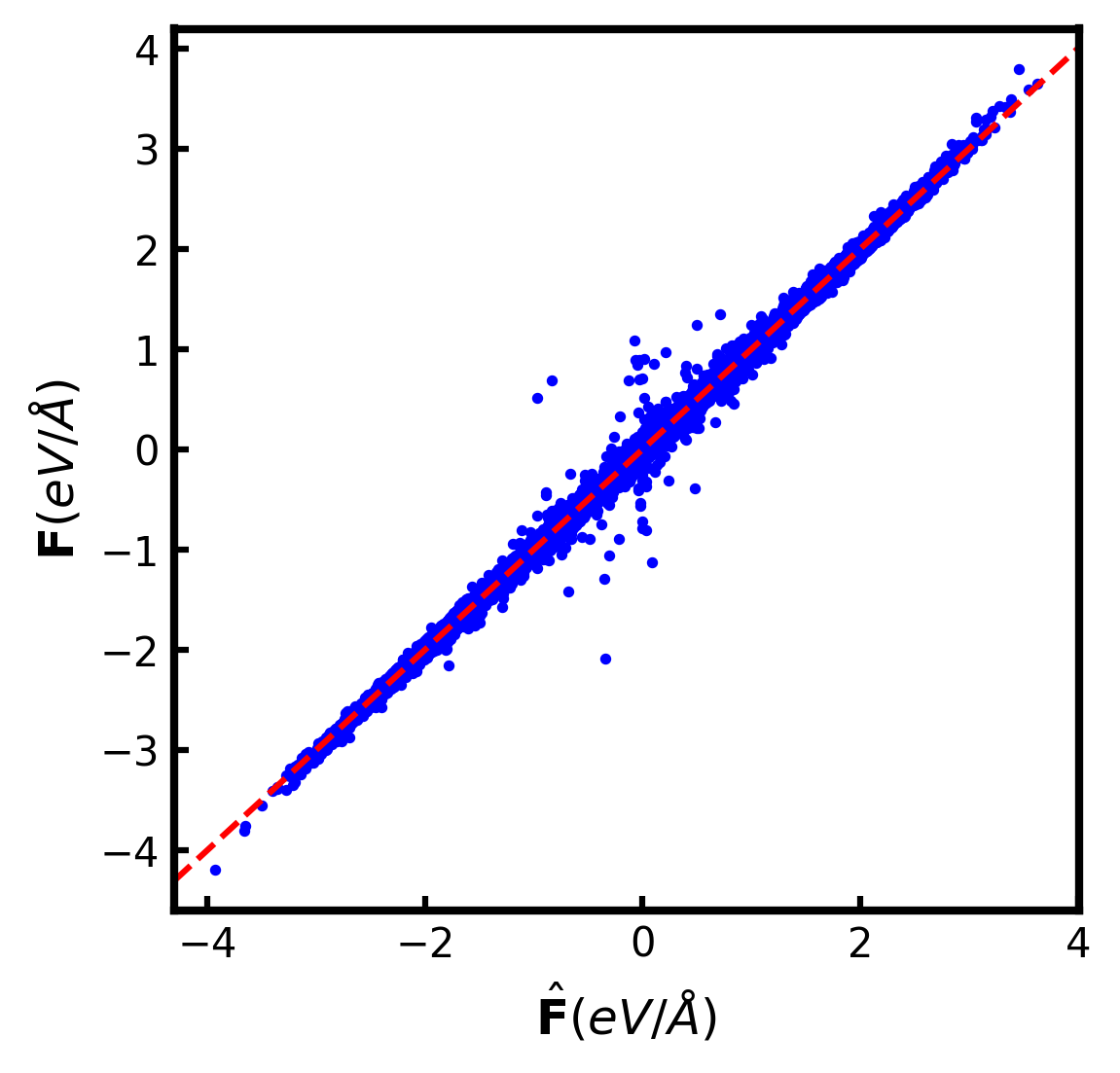}
    \captionsetup{justification=centering}
    \caption{GAMD trained on\\ TIP4P-Ew}
    \end{subfigure}
    \begin{subfigure}{0.32\textwidth}
    \centering
    \includegraphics[width=\linewidth]{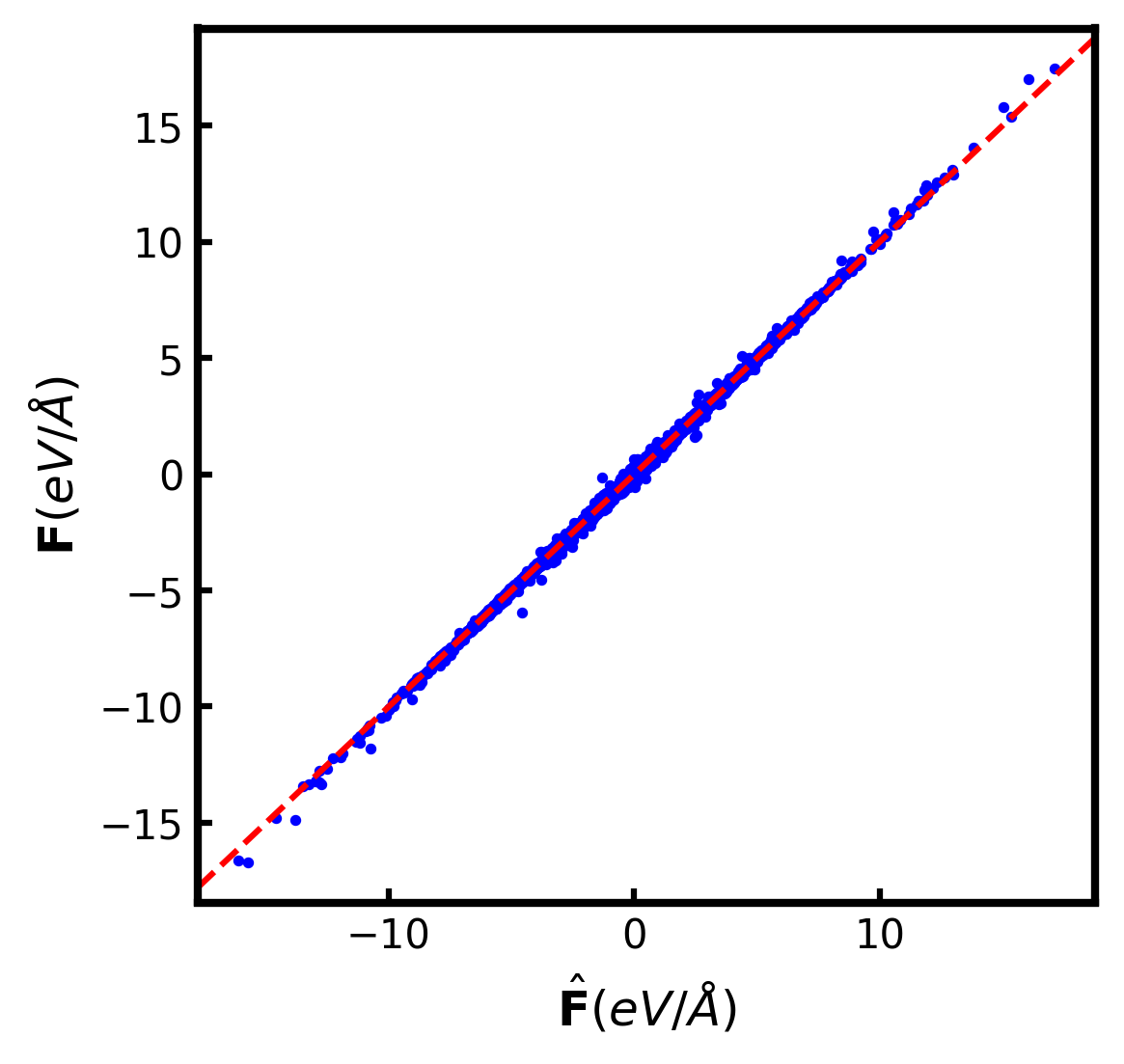}
    \captionsetup{justification=centering}
    \caption{GAMD trained on\\ RPBE-D3}
    \end{subfigure}
    \captionsetup{justification=centering}
    \caption{Forces derived from different water potential models versus GAMD predicted forces. $\mathbf{F}$ denotes ground truth forces, $\hat{\mathbf{F}}$ denotes GAMD's prediction.}
    \label{fig:force angle water}
\end{figure}

We run the MD simulation using GAMD on a cubic box (2 nanometers each edge with periodic boundary) with 258 water molecules of 300K temperature. Similar to the Lennard-Jones experiment, we investigate the RDF curves to evaluate the spatial behavior of the simulated trajectories using GAMD's predicted forces. In general, GAMD can learn to predict forces from different data sources. The plotted RDF curve of Hydrogen-Hydrogen, Oxygen-Oxygen, and Oxygen-Hydrogen element pairs show the consistency of the GAMD's trajectories with different reference models' simulated trajectories (Figure~\ref{fig:tip3p rdf}, \ref{fig:tip4p rdf}). In addition, as shown in Figure ~\ref{fig:dft rdf}, the trajectory from GAMD trained on RPBE data has spatial structures that are consistent with experimental data.
In addition to the evaluation based on the spatial behavior of the simulated trajectories, we can directly compare GAMD's predicted atomic forces with the forces derived from different force models by studying the predicted forces of GAMD trained on different data sources. Agreement in the direction of the forces is evaluated by comparing the $\cos\left<\hat{\mathbf{F}},\mathbf{F}\right>$ between the direction of predicted and reference forces. It has been observed that for models trained on empirical forcefields, more than $ 95 \%$ of the predictions, $\cos\left<\hat{\mathbf{F}},\mathbf{F}\right>$ is of agreement above $ 0.995 $ and $ 0.980 $ for model trained on DFT data. Quantitative evaluation of errors of the predictions is reported in Table~\ref{Tab: water box force}. Figure~\ref{fig:force angle water} depicts the alignment between predicted forces in each direction with the forces derived from different force models.

\begin{table}
\centering
\scalebox{0.91}{
\begin{tabular}{cccccc} 
\toprule
Ground Truth & Test snapthots & $\text{MAE}$ ($meV/ \si{\angstrom}$) & $\text{RMSE}$ ($meV/ \si{\angstrom}$) &$\cos<\hat{\mathbf{F}},\mathbf{F}>$ & Relative error \\ 
\midrule
TIP3P~\citep{TIP3P} & 1000 & 11.26 $\pm$ 0.84  & 15.16 $\pm$ 1.16 & 0.997 & 1.16\%  $\pm$ 0.09\%\\
TIP4P-Ew~\citep{TIP4PEw} & 1000 & 13.86 $\pm$ 1.44 & 19.16 $\pm$ 4.47 & 0.999 & 1.29\% $\pm$ 0.13\% \\
RPBE-D3~\citep{Morawietz8368} & 723 &24.28 $\pm$ 16.80 & 35.39 $\pm$ 23.09 & 0.986 & 1.47\% $\pm$ 1.02\% \\
\bottomrule
\end{tabular}}
\caption{Quantitative analysis of force accuracy}
\label{Tab: water box force}
\end{table}

\begin{figure}[H]
    \centering
    \begin{subfigure}{0.49\textwidth}
    \centering
    \includegraphics[width=\linewidth]{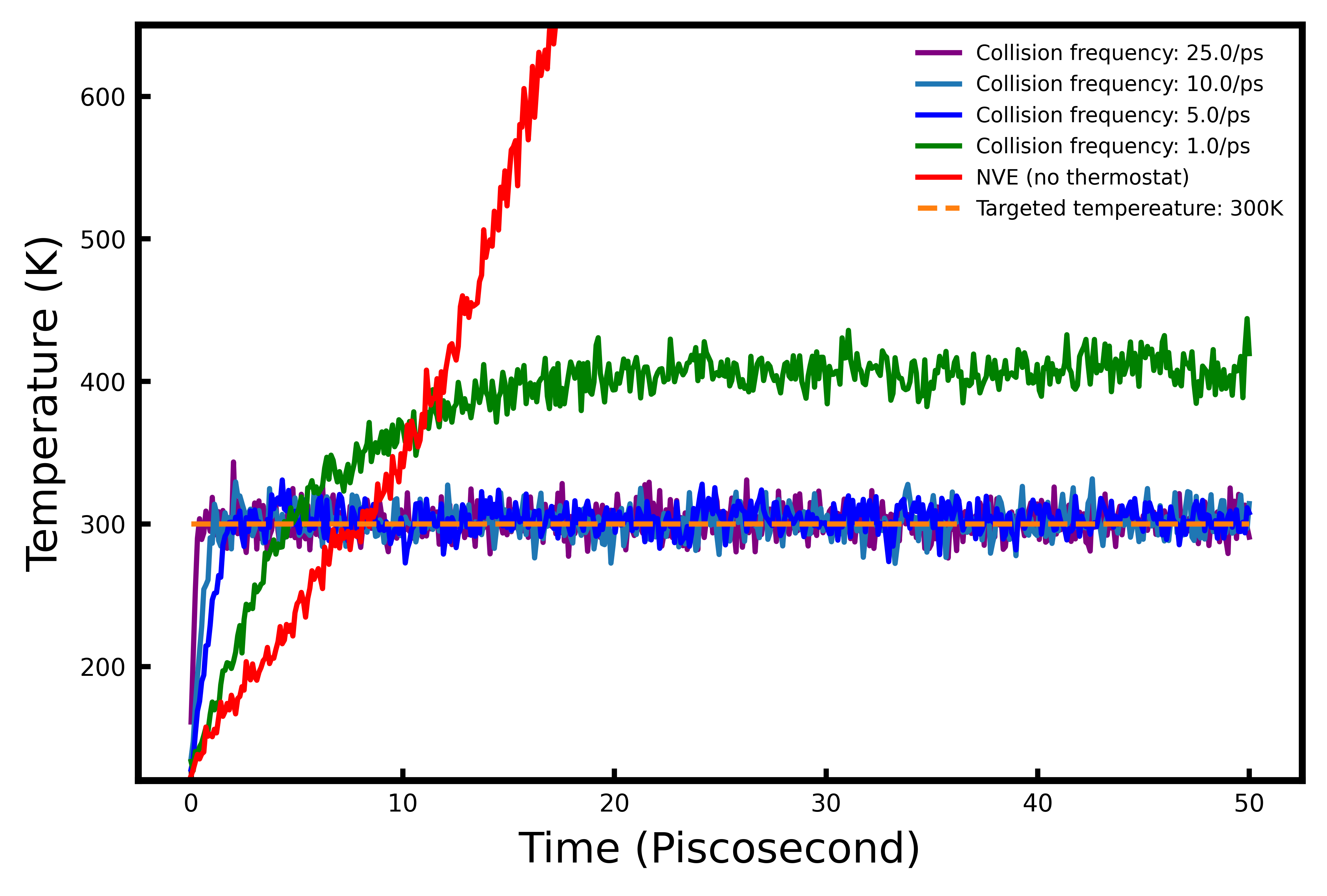}
    \captionsetup{justification=centering}
    \caption{Nos\'{e}--Hoover thermostat}
    \end{subfigure}
     \begin{subfigure}{0.49\textwidth}
    \centering
    \includegraphics[width=\linewidth]{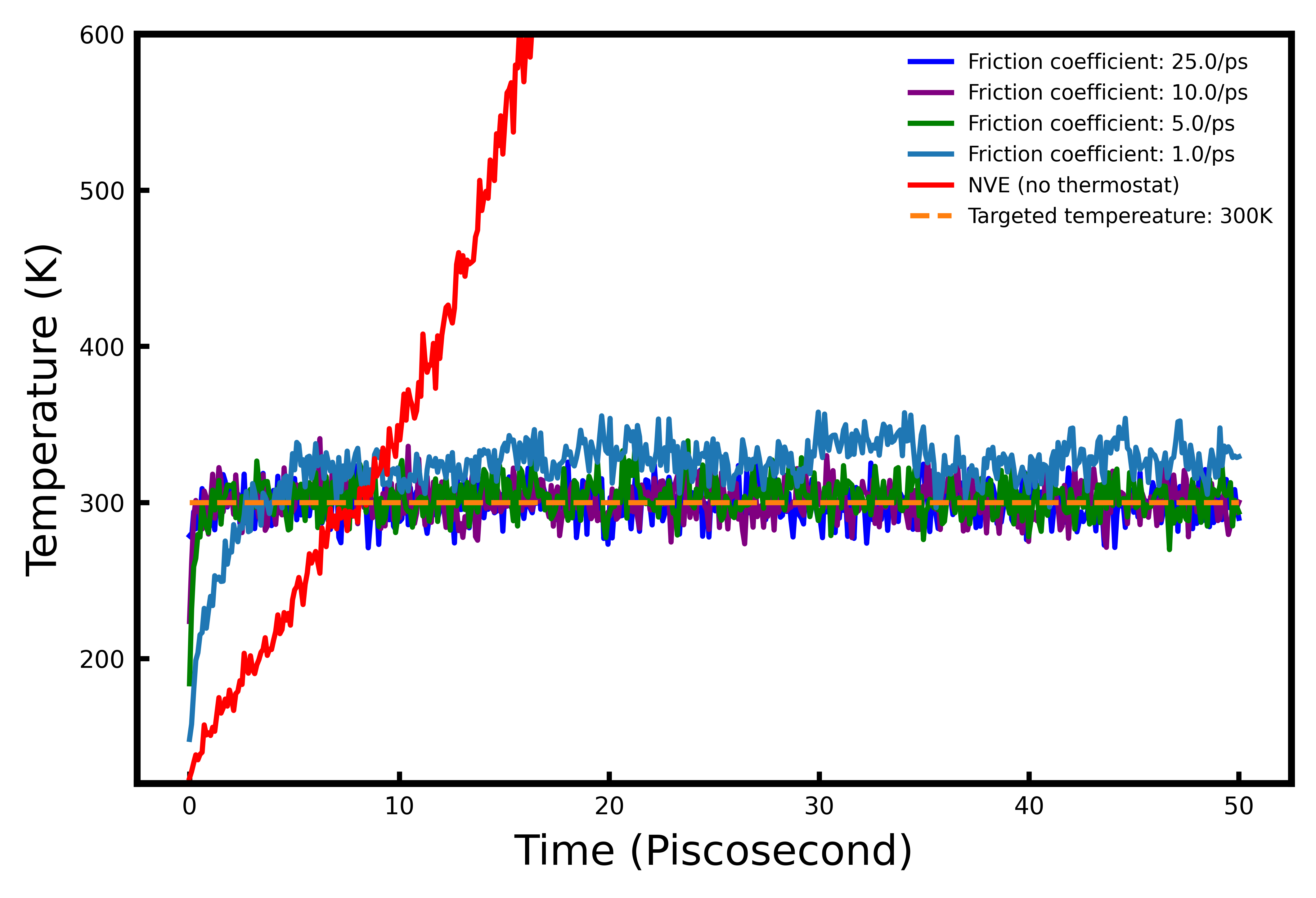}
    \captionsetup{justification=centering}
    \caption{Langevin thermostat}
    \end{subfigure}
    \captionsetup{justification=centering}
    \caption{GAMD trained on TIP3P data.}
    \label{fig:thermostat ablation tip3p}
\end{figure}

\begin{figure}[H]
    \centering
    \begin{subfigure}{0.49\textwidth}
    \centering
    \includegraphics[width=\linewidth]{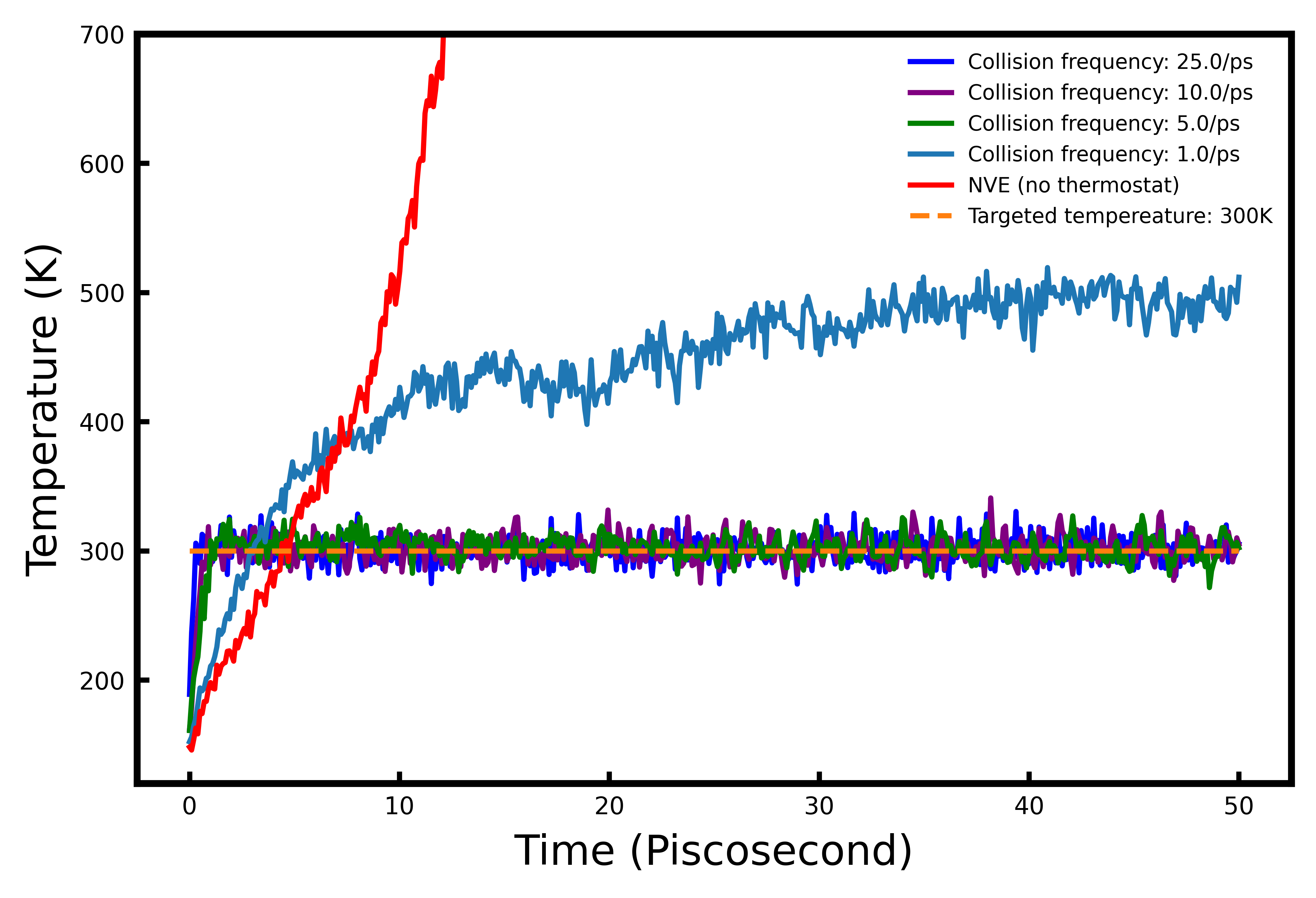}
    \captionsetup{justification=centering}
    \caption{Nos\'{e}--Hoover thermostat}
    \end{subfigure}
     \begin{subfigure}{0.49\textwidth}
    \centering
    \includegraphics[width=\linewidth]{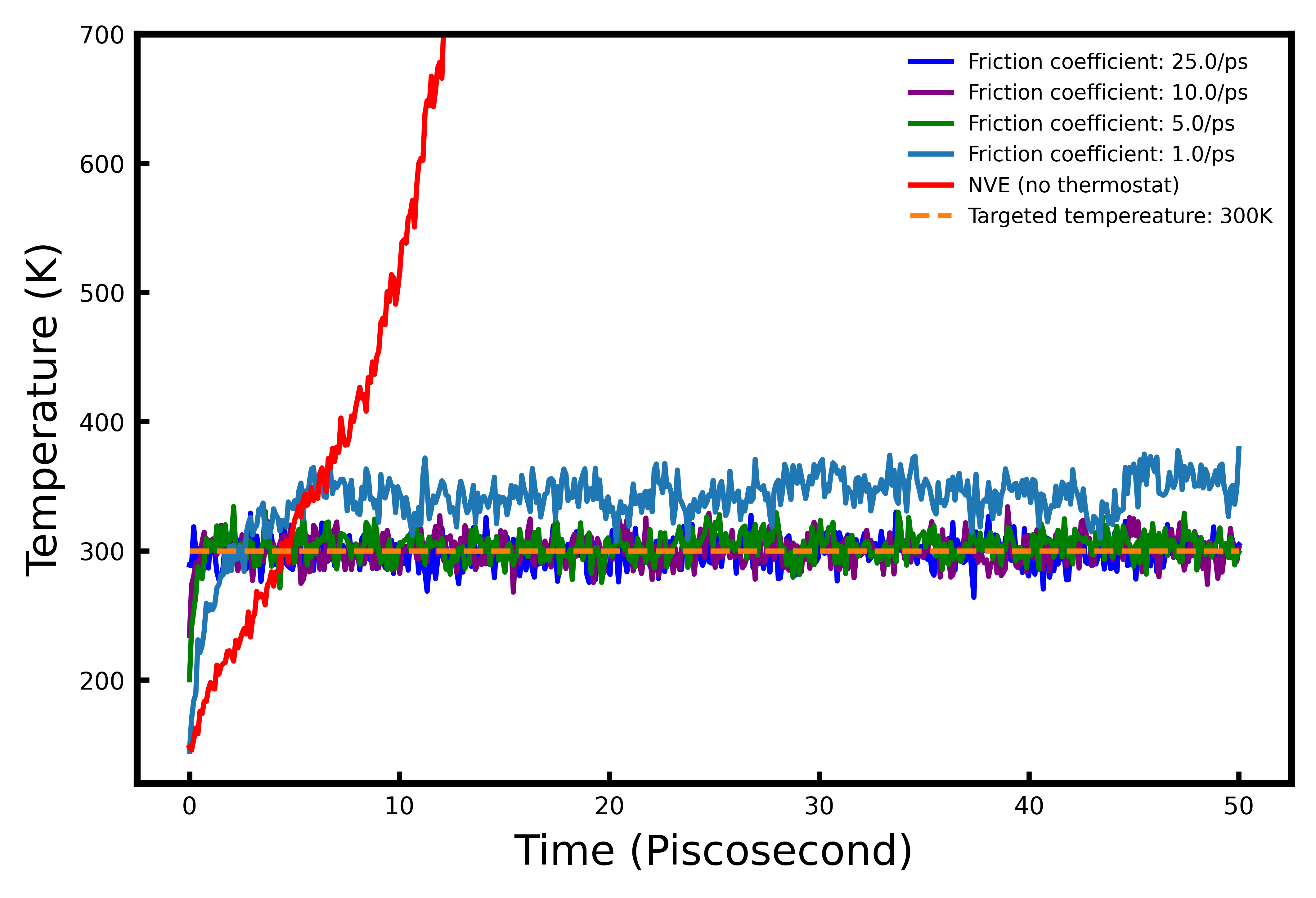}
    \captionsetup{justification=centering}
    \caption{Langevin thermostat}
    \end{subfigure}
    \captionsetup{justification=centering}
    \caption{GAMD trained on TIP4P-Ew data.}
    \label{fig:thermostat ablation tip4p}
\end{figure}

\begin{figure}[H]
    \centering
    \begin{subfigure}{0.49\textwidth}
    \centering
    \includegraphics[width=\linewidth]{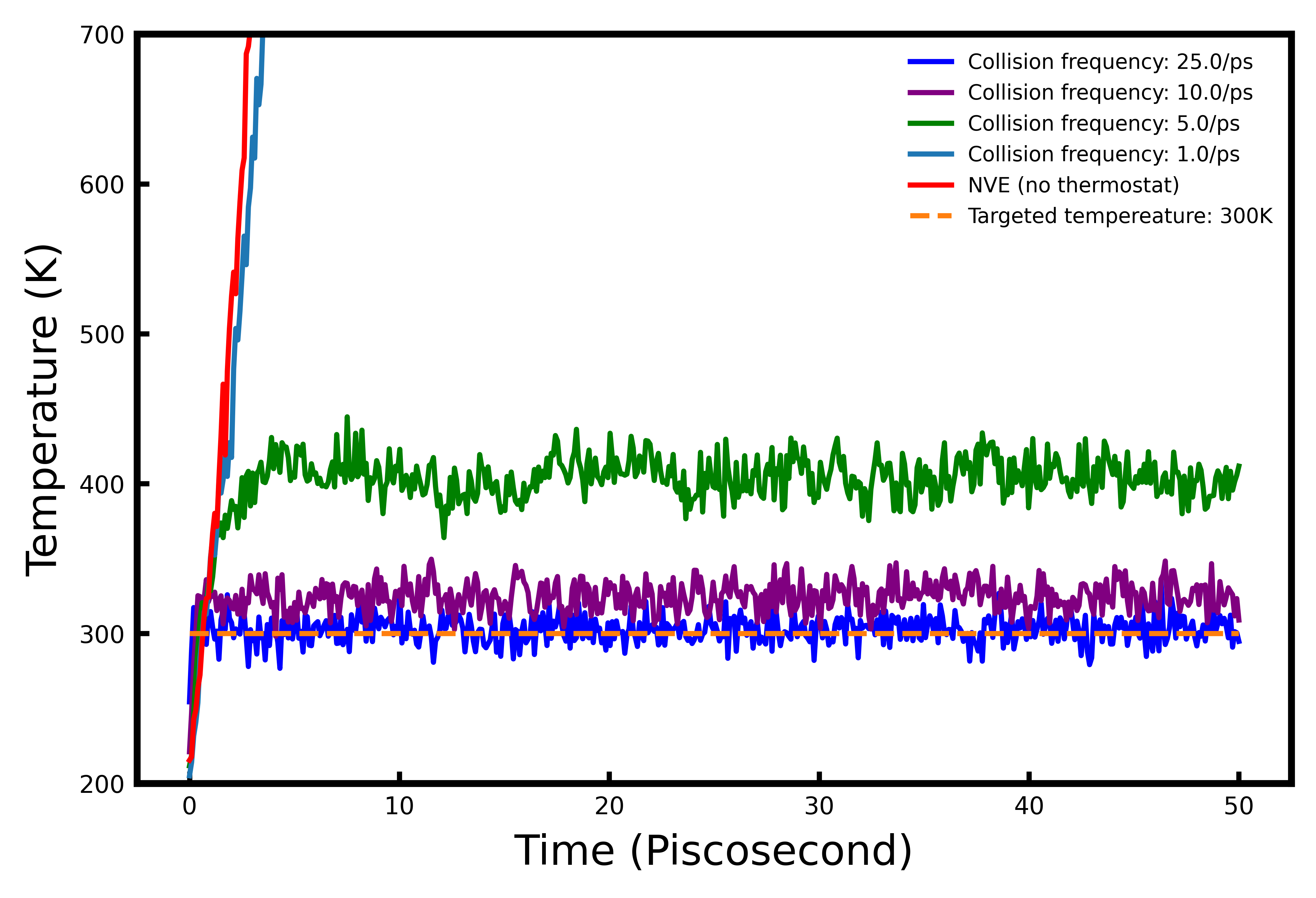}
    \captionsetup{justification=centering}
    \caption{Nos\'{e}--Hoover thermostat}
    \end{subfigure}
     \begin{subfigure}{0.49\textwidth}
    \centering
    \includegraphics[width=\linewidth]{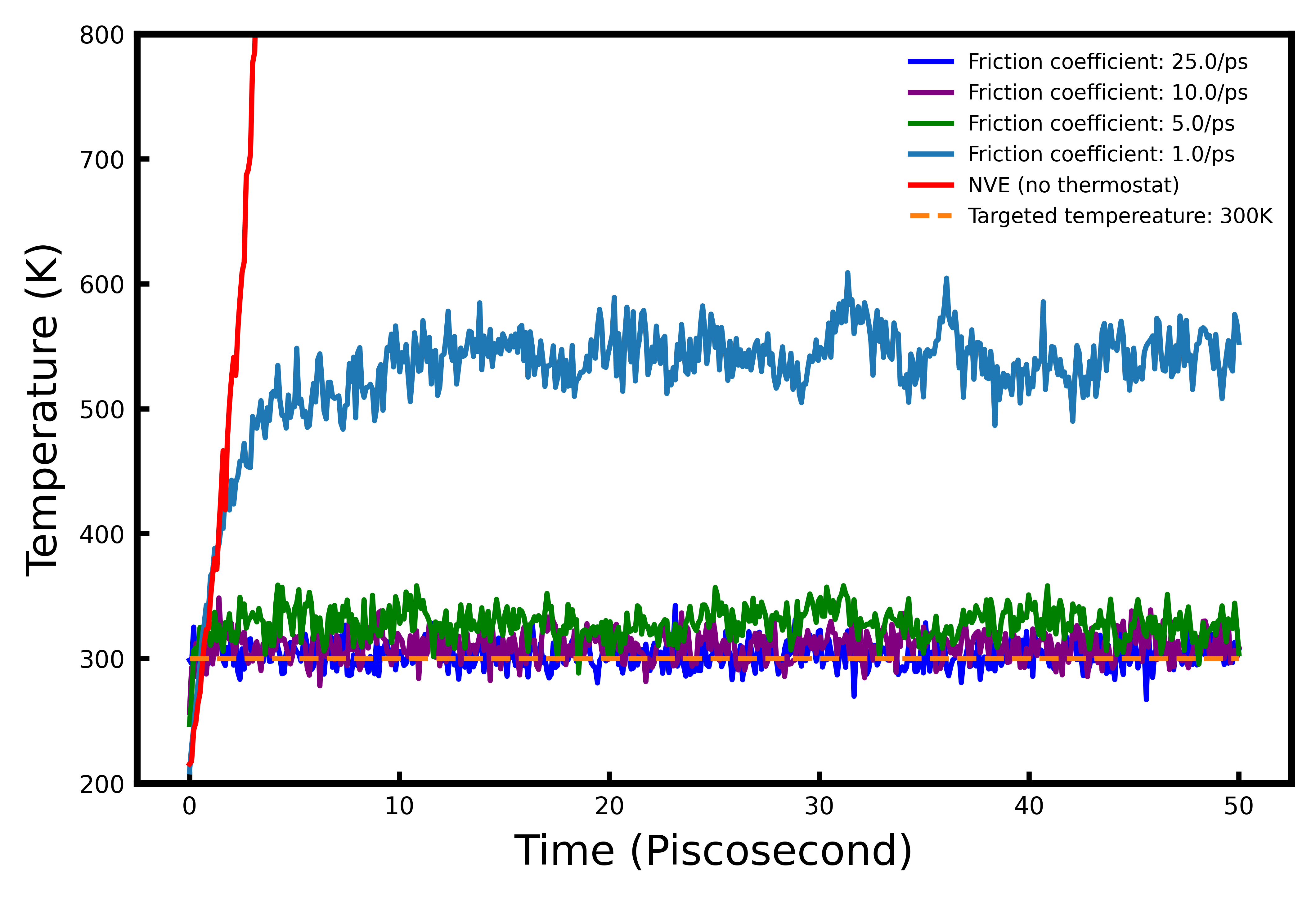}
    \captionsetup{justification=centering}
    \caption{Langevin thermostat}
    \end{subfigure}
    \captionsetup{justification=centering}
    \caption{GAMD trained on RPBE data.}
    \label{fig:thermostat ablation dft}
\end{figure}

Since forces in GAMD are not derived from potential energy surface, it does not conserve the potential energy of the system. Therefore GAMD requires thermostats to regulate the velocity during simulation. We study the temporal behavior and sensitivity of GAMD by using different thermostats to run NVT simulations. We test GAMD with different collision/dampening coefficients on Nos\'{e}--Hoover thermostat\cite{Nose} and Langevin thermostat with BAOAB scheme\cite{Langevin2013}. The temperature trends of GAMD trained on different water force models are shown in Figure~\ref{fig:thermostat ablation tip3p}, \ref{fig:thermostat ablation tip4p}, \ref{fig:thermostat ablation dft}. It is observed that GAMD cannot equilibrate under the NVE ensemble given its non-conserving property. When a relatively passive thermostat is applied, GAMD's equilibrium will deviate from the target heat equilibrium. The models trained on TIP3P and TIP4P-Ew MD data are less sensitive to the intensity of the thermostat and can equilibrate to the target temperature with collision/dampening coefficient $\geq 5.0/ps$. The model trained on DFT data is more sensitive to this coefficient and requires more aggressive thermostat to maintain the target temperature. We hypothesize the main reason is that DFT training data encompasses a much diverse range of structures from low energy to high energy while MD data mostly consist of structures simulated around 300K temperature.

\subsection{Scalability and Speed}
\label{sec: benchmark}
\begin{table}[H]
\centering
\scalebox{0.95}{
\begin{tabular}{cccccccc} 
\toprule
\multirow{3}{*}{Model} & \multirow{3}{*}{\begin{tabular}[c]{@{}c@{}}Molecule\\number\end{tabular}} & \multirow{3}{*}{\begin{tabular}[c]{@{}c@{}}Box size\\(nm)\end{tabular}} & \multicolumn{5}{c}{Time (ms) per force calculation step} \\ 
\cmidrule{4-8}
 &  &  & \multirow{2}{*}{OpenMM} & \multicolumn{2}{c}{GAMD} & \multicolumn{2}{c}{LAMMPS} \\ 
\cmidrule{5-8}
 &  &  &  & ~Neighbor & Force Eval~ & ~Neighbor & Force Eval~ \\ 
\toprule
\multirow{6}{*}{3-site} & 258 & 2 & 0.192 & 2.962 & 4.063 & 0.001 & 0.754 \\
 & 887 & 3 & 0.363 & 3.240 & 4.089 & 0.004 & 3.749 \\
 & 2094 & 4 & 0.644 & 3.745 & 4.143 & 0.008 & 4.114 \\
 & 4085 & 5 & 1.310 & 4.445 & 4.035 & 0.010 & 5.431 \\
 & 13786 & 7.5 & 3.817 & 10.300 & 4.230 & 0.028 & 11.578 \\
 & 24093 & 9 & 7.652 & 15.342 & 5.545 & 0.050 & 18.531 \\ 
\midrule
4-site & 24093 & 9 & 7.848 & \multicolumn{2}{c}{-} & 0.056 & 28.164 \\
\bottomrule
\end{tabular}}
\protect\caption[position=bottom]{Quantitative comparison of force calculation speed on the Water box of different scales. In GAMD and LAMMPS, the time for each force evaluation is decomposed into two parts and reported separately. \textit{Neighbor} denotes the time needed for updating the neighbor list (building graph and calculating edge features in GNN). \textit{Force Eval} denotes the time to evaluate the net per-atom forces based on the neighbor list. GAMD adopts a 3-site setting for all of the calculations. The classical MD methods tested are TIP3P\citep{TIP3P} and TIP4P-Ew\citep{TIP4PEw}.}

\label{tab: benchmark}
\end{table}

We compare our model to two high-performance MD packages, OpenMM and LAMMPS, on the water box under multiple scales, ranging from 2 nanometers to 9 nanometers. The major bottleneck of current GAMD's implementation is neighbor lists' update and graph construction. The neighbor list update and graph construction in GAMD are based on JAX-MD, JAX, and DGL's sparse matrix operation, which are optimized for machine learning workloads and are generally slower than customized CUDA neighbor-search routines used by MD packages. This results in the noticeable gap between the time of neighbor list update (including adjacency matrix assembling) in GAMD and counterparts in other MD packages.  Despite this factor, GAMD still has a competitive computing efficiency (as shown in Table \ref{tab: benchmark}) at the large-scale simulation which is of great importance in practical problems \cite{large-scale-waterMD-HIV}. Since GAMD predicts forces directly by a set of MLPs which can be efficiently parallelized on the GPU, its actual force evaluation speed is faster than classical MD methods. We believe that with a more optimized implementation of the neighbor list data structure, GAMD's performance can be further improved.

In comparision to classical MD simulation methods and DFT calculation, GAMD offers benefits in the following aspects. First, GAMD does not require explicit evaluation of energy and its gradient, instead, it directly predicts the net per-particle forces. Hence GAMD can learn the forcefield directly from observed data without any prior knowledge of the underlying energy equation. Second, GAMD does not require neighbor selection and energy accumulation based on different types of interactions (e.g. bonded interaction usually happened within a local region while non-bonded interaction happened between long-range particles). This reduces the computations needed in each step of simulation especially when there are multiple kinds of interactions in the system.
\begin{figure}[H]
\begin{subfigure}{0.32\linewidth}
\centering
\includegraphics[width=\linewidth]{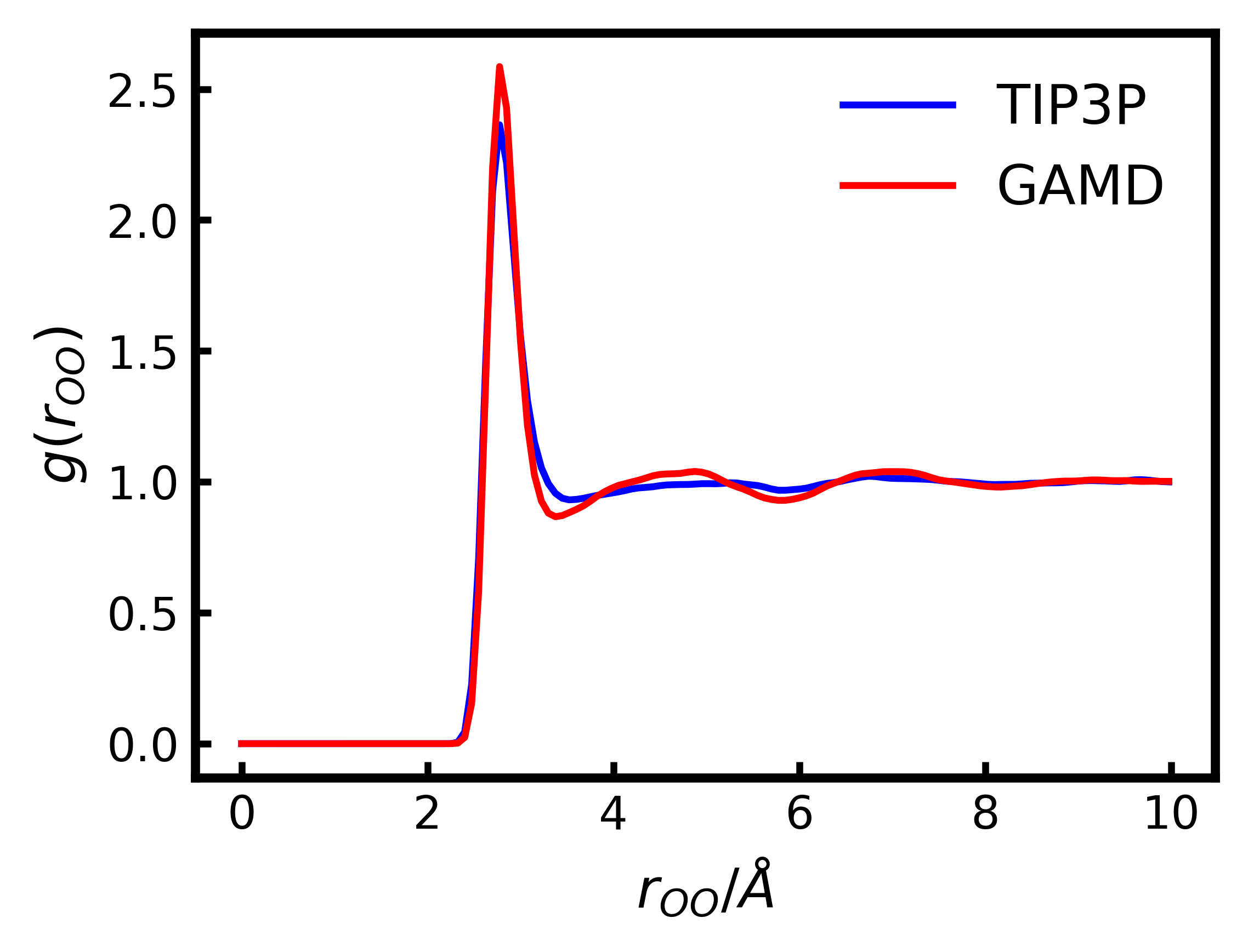}
\caption{Box size: $3 ~nm$}
\end{subfigure}
\begin{subfigure}{0.32\linewidth}
\includegraphics[width=\linewidth]{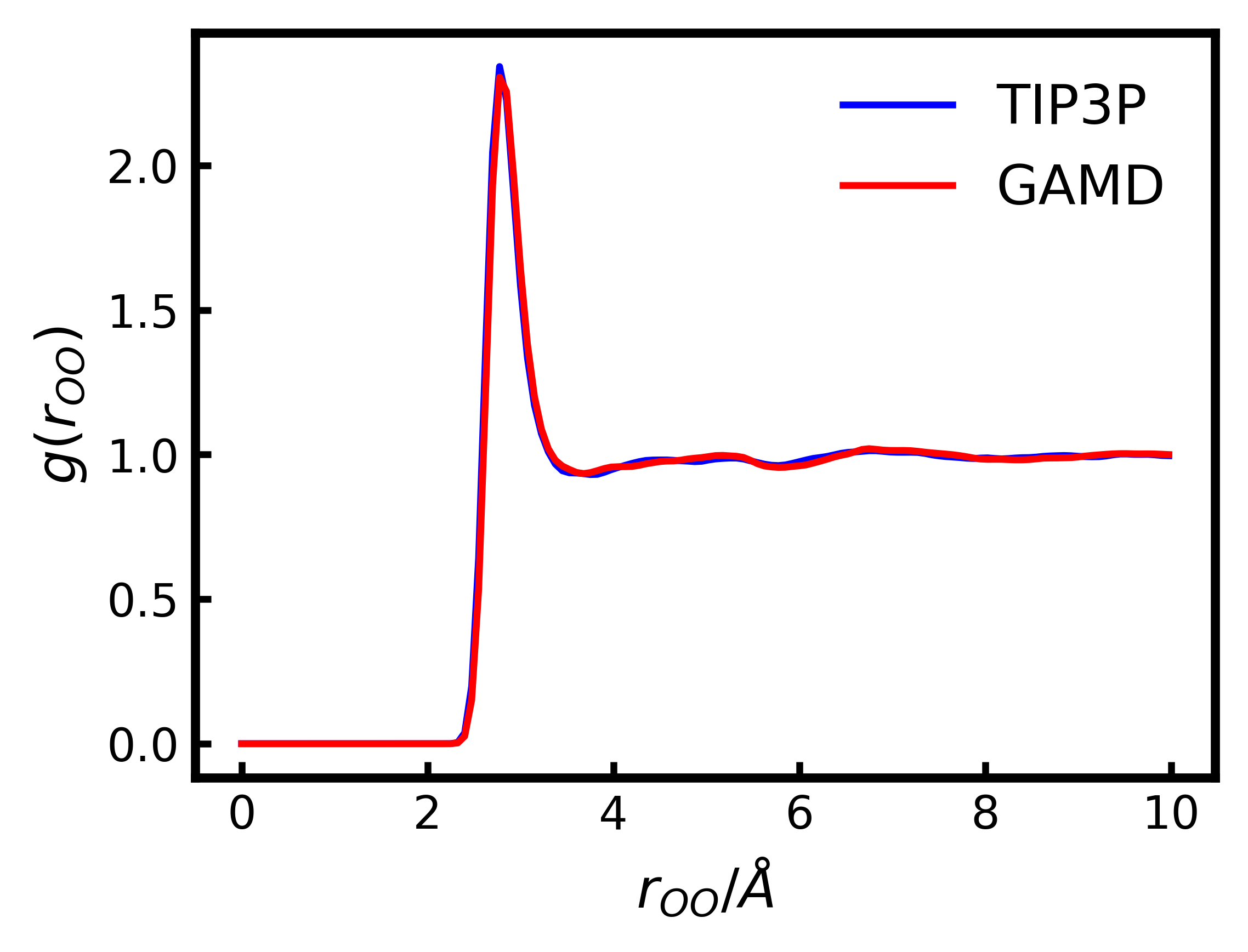}
\caption{Box size: $6 ~nm$}
\end{subfigure}
\begin{subfigure}{0.32\linewidth}
\includegraphics[width=\linewidth]{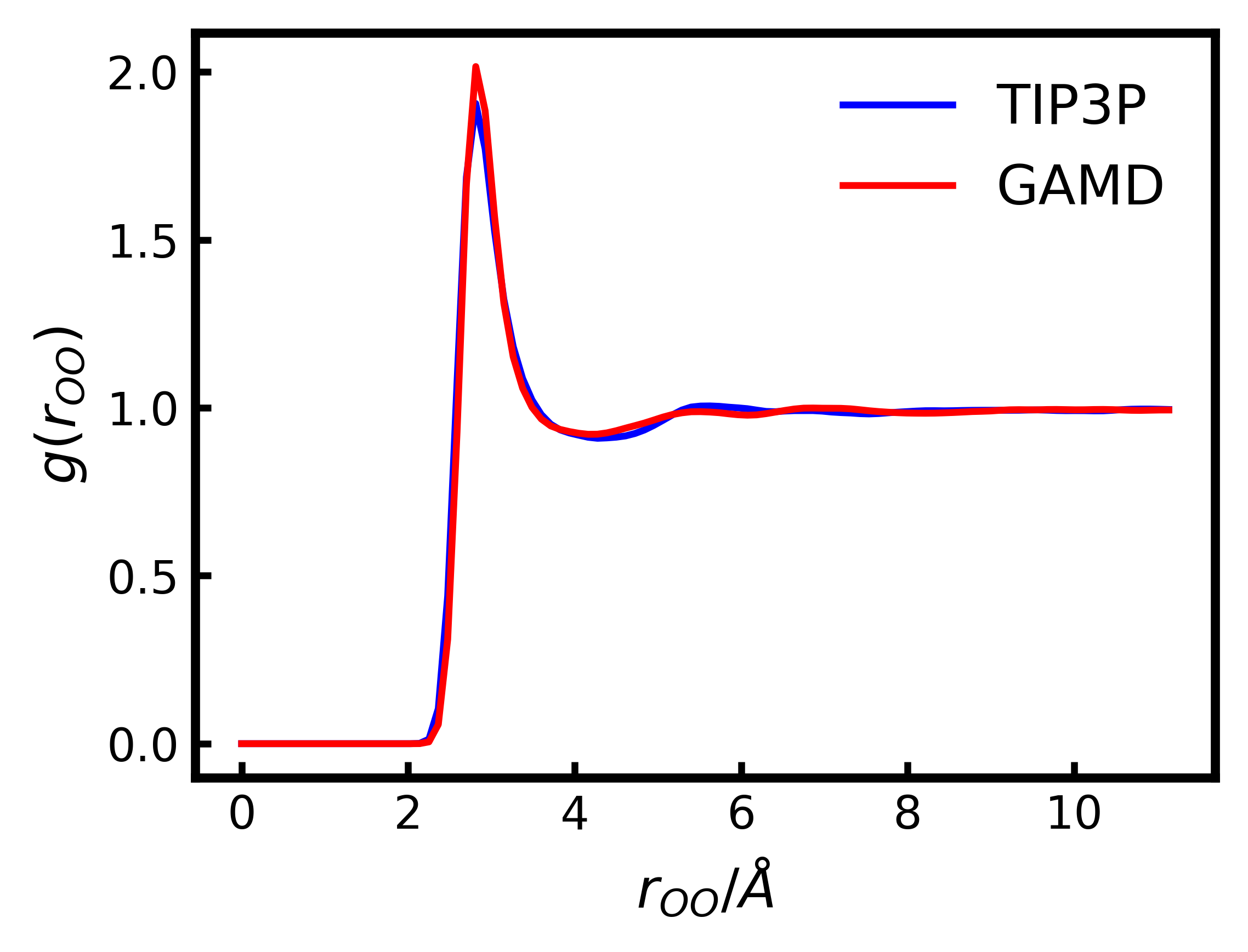}
\caption{Box size: $9 ~nm$}
 \label{fig:large box}
\end{subfigure}
\captionsetup{justification=centering}
\caption{Radial distribution function for Oxygen-Oxygen distance in the water boxes of different sizes. (All the RDF plots are cutoff at $\sim 10 \si{\angstrom}$)}
\label{fig:Water box RDF different size}
\end{figure} 
\begin{figure}[H]
    \centering
    \includegraphics[width=0.40\linewidth]{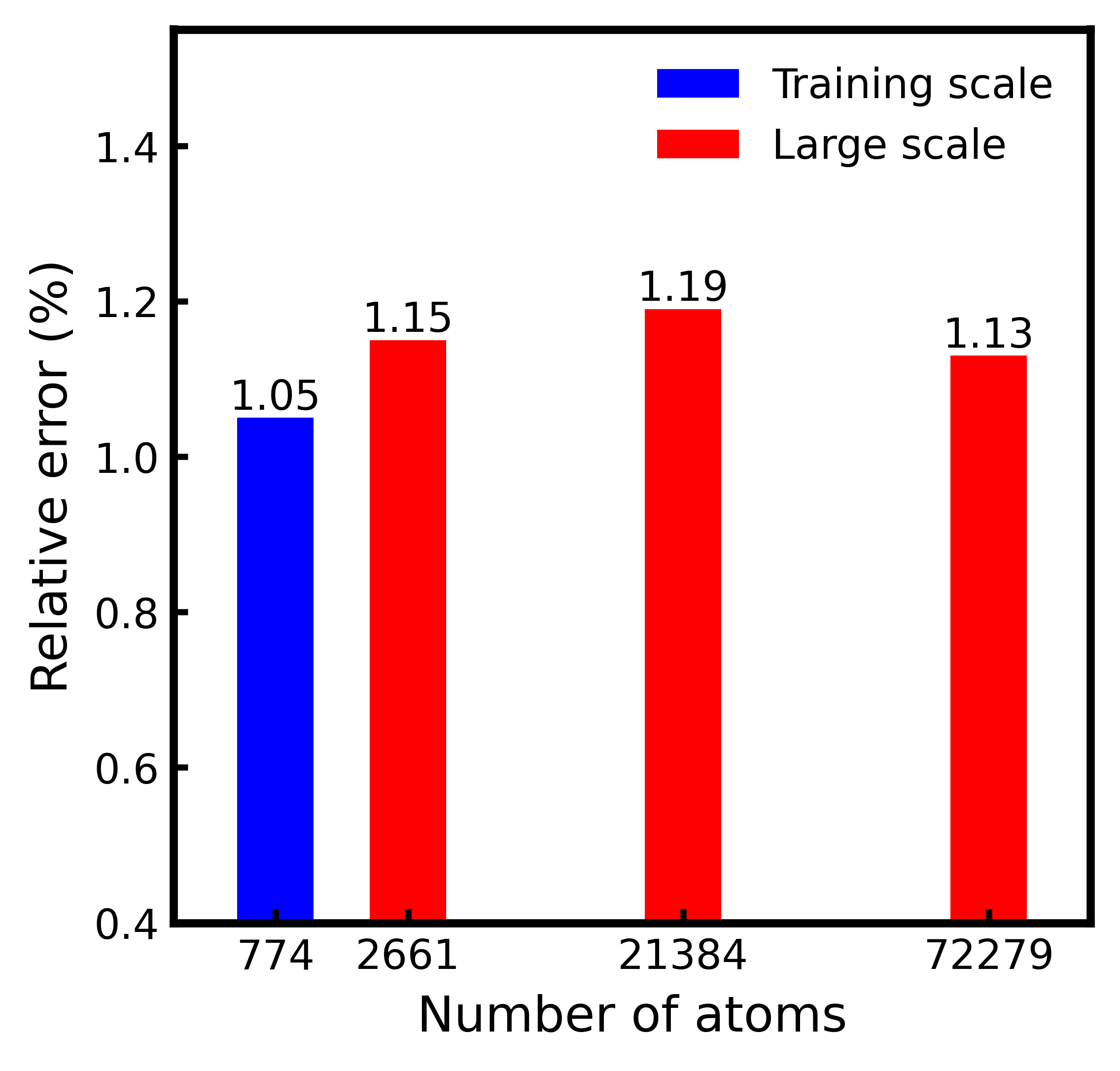}
    \caption{Relative error of force prediction on different sizes of water boxes. Error is measured and averaged on each simulation sequence of specific resolution. }
    \label{fig:force scaling}
    \vspace{-2mm}
\end{figure}
Furthermore, GAMD is a supervised machine learning model in essence, but its scalability is not limited by the scale of training data. We find that GAMD can learn scale-agnostic dynamics and generalize well to much larger systems. Figure \ref{fig:Water box RDF different size} and Figure \ref{fig:force scaling} show that GAMD trained on a water box with 258 water molecules can be scaled up to a water box with much more molecules (about 100 times larger) without compromising accuracy. This demonstrates the scalability of the model and broadens the range of the problems where GAMD can operate.

\section{Conclusion}
A Graph Neural Networks Accelerated Molecular Dynamics model (GAMD) is presented. GAMD provides a data-driven framework that can predict atomic forces directly without explicitly calculating energy and does not require hand-designed molecular fingerprints. As GAMD does not derive energy from potential energy surface, it does not conserve energy and requires a thermostat to regulate velocities. We have showcased the applications of GAMD on two typical molecular systems - Lennard-Jones particles and water molecules.  We use GAMD to simulate these systems in an NVT ensemble, which generates trajectories that are consistent with existing classical MD methods and experimental data in terms of spatial distribution and equilibrium state. It has also been shown that GAMD can be scaled up to much larger systems without compromising accuracy. Furthermore, a comprehensive benchmark on GAMD and other production-level MD engines (LAMMPS, OpenMM) is conducted, which shows GAMD has competitive efficiency at large-scale simulation.
\vspace{-2mm}

\section{Data Availability Statements}
The data and code that support the findings of this project can be found at: \url{https://github.com/BaratiLab/GAMD}.


\begin{acknowledgement}
\vspace{-2mm}
This work is supported by the start-up fund provided by CMU Mechanical Engineering, United States. The authors would like to thank Zhonglin Cao for valuable comments.
\end{acknowledgement}


\bibliography{ref}

\providecommand{\latin}[1]{#1}
\makeatletter
\providecommand{\doi}
  {\begingroup\let\do\@makeother\dospecials
  \catcode`\{=1 \catcode`\}=2 \doi@aux}
\providecommand{\doi@aux}[1]{\endgroup\texttt{#1}}
\makeatother
\providecommand*\mcitethebibliography{\thebibliography}
\csname @ifundefined\endcsname{endmcitethebibliography}
  {\let\endmcitethebibliography\endthebibliography}{}
\begin{mcitethebibliography}{92}
\providecommand*\natexlab[1]{#1}
\providecommand*\mciteSetBstSublistMode[1]{}
\providecommand*\mciteSetBstMaxWidthForm[2]{}
\providecommand*\mciteBstWouldAddEndPuncttrue
  {\def\EndOfBibitem{\unskip.}}
\providecommand*\mciteBstWouldAddEndPunctfalse
  {\let\EndOfBibitem\relax}
\providecommand*\mciteSetBstMidEndSepPunct[3]{}
\providecommand*\mciteSetBstSublistLabelBeginEnd[3]{}
\providecommand*\EndOfBibitem{}
\mciteSetBstSublistMode{f}
\mciteSetBstMaxWidthForm{subitem}{(\alph{mcitesubitemcount})}
\mciteSetBstSublistLabelBeginEnd
  {\mcitemaxwidthsubitemform\space}
  {\relax}
  {\relax}

\bibitem[Hollingsworth and Dror(2018)Hollingsworth, and Dror]{MD-for-all-2018}
Hollingsworth,~S.~A.; Dror,~R.~O. Molecular Dynamics Simulation for All.
  \emph{Neuron} \textbf{2018}, \emph{99}, 1129--1143\relax
\mciteBstWouldAddEndPuncttrue
\mciteSetBstMidEndSepPunct{\mcitedefaultmidpunct}
{\mcitedefaultendpunct}{\mcitedefaultseppunct}\relax
\EndOfBibitem
\bibitem[Karplus and McCammon(2002)Karplus, and
  McCammon]{Karplus2002-MD-biomolecules}
Karplus,~M.; McCammon,~J.~A. Molecular dynamics simulations of biomolecules.
  \emph{Nature Structural Biology} \textbf{2002}, \emph{9}, 646--652\relax
\mciteBstWouldAddEndPuncttrue
\mciteSetBstMidEndSepPunct{\mcitedefaultmidpunct}
{\mcitedefaultendpunct}{\mcitedefaultseppunct}\relax
\EndOfBibitem
\bibitem[Becke(2014)]{DFT-review-2014}
Becke,~A.~D. Perspective: Fifty years of density-functional theory in chemical
  physics. \emph{The Journal of Chemical Physics} \textbf{2014}, \emph{140},
  18A301\relax
\mciteBstWouldAddEndPuncttrue
\mciteSetBstMidEndSepPunct{\mcitedefaultmidpunct}
{\mcitedefaultendpunct}{\mcitedefaultseppunct}\relax
\EndOfBibitem
\bibitem[Unke \latin{et~al.}(2020)Unke, Koner, Patra, Käser, and
  Meuwly]{Unke-2020-PES-MLST}
Unke,~O.~T.; Koner,~D.; Patra,~S.; Käser,~S.; Meuwly,~M. High-dimensional
  potential energy surfaces for molecular simulations: from empiricism to
  machine learning. \emph{Machine Learning: Science and Technology}
  \textbf{2020}, \emph{1}, 013001\relax
\mciteBstWouldAddEndPuncttrue
\mciteSetBstMidEndSepPunct{\mcitedefaultmidpunct}
{\mcitedefaultendpunct}{\mcitedefaultseppunct}\relax
\EndOfBibitem
\bibitem[Harrison \latin{et~al.}(2018)Harrison, Schall, Maskey, Mikulski,
  Knippenberg, and Morrow]{review-FFs-2018}
Harrison,~J.~A.; Schall,~J.~D.; Maskey,~S.; Mikulski,~P.~T.;
  Knippenberg,~M.~T.; Morrow,~B.~H. Review of force fields and intermolecular
  potentials used in atomistic computational materials research. \emph{Applied
  Physics Reviews} \textbf{2018}, \emph{5}, 031104\relax
\mciteBstWouldAddEndPuncttrue
\mciteSetBstMidEndSepPunct{\mcitedefaultmidpunct}
{\mcitedefaultendpunct}{\mcitedefaultseppunct}\relax
\EndOfBibitem
\bibitem[Paquet and Viktor(2015)Paquet, and Viktor]{Paquet2015-MD-time}
Paquet,~E.; Viktor,~H.~L. Molecular dynamics, monte carlo simulations, and
  langevin dynamics: a computational review. \emph{BioMed research
  international} \textbf{2015}, \emph{2015}, 183918--183918,
  25785262[pmid]\relax
\mciteBstWouldAddEndPuncttrue
\mciteSetBstMidEndSepPunct{\mcitedefaultmidpunct}
{\mcitedefaultendpunct}{\mcitedefaultseppunct}\relax
\EndOfBibitem
\bibitem[Dror \latin{et~al.}(2010)Dror, Jensen, Borhani, and
  Shaw]{time-MD-2010}
Dror,~R.~O.; Jensen,~M.; Borhani,~D.~W.; Shaw,~D.~E. {Exploring atomic
  resolution physiology on a femtosecond to millisecond timescale using
  molecular dynamics simulations}. \emph{Journal of General Physiology}
  \textbf{2010}, \emph{135}, 555--562\relax
\mciteBstWouldAddEndPuncttrue
\mciteSetBstMidEndSepPunct{\mcitedefaultmidpunct}
{\mcitedefaultendpunct}{\mcitedefaultseppunct}\relax
\EndOfBibitem
\bibitem[Chmiela \latin{et~al.}(2017)Chmiela, Tkatchenko, Sauceda, Poltavsky,
  Sch{\"u}tt, and M{\"u}ller]{Chmiela-GDML-2017}
Chmiela,~S.; Tkatchenko,~A.; Sauceda,~H.~E.; Poltavsky,~I.; Sch{\"u}tt,~K.~T.;
  M{\"u}ller,~K.-R. Machine learning of accurate energy-conserving molecular
  force fields. \emph{Science Advances} \textbf{2017}, \emph{3}\relax
\mciteBstWouldAddEndPuncttrue
\mciteSetBstMidEndSepPunct{\mcitedefaultmidpunct}
{\mcitedefaultendpunct}{\mcitedefaultseppunct}\relax
\EndOfBibitem
\bibitem[Hu \latin{et~al.}(2021)Hu, Shuaibi, Das, Goyal, Sriram, Leskovec,
  Parikh, and Zitnick]{hu-2021-forcenet}
Hu,~W.; Shuaibi,~M.; Das,~A.; Goyal,~S.; Sriram,~A.; Leskovec,~J.; Parikh,~D.;
  Zitnick,~C.~L. ForceNet: A Graph Neural Network for Large-Scale Quantum
  Calculations. 2021\relax
\mciteBstWouldAddEndPuncttrue
\mciteSetBstMidEndSepPunct{\mcitedefaultmidpunct}
{\mcitedefaultendpunct}{\mcitedefaultseppunct}\relax
\EndOfBibitem
\bibitem[Mailoa \latin{et~al.}(2019)Mailoa, Kornbluth, Batzner, Samsonidze,
  Lam, Vandermause, Ablitt, Molinari, and
  Kozinsky]{fast-covariant-force-NNFF-2019}
Mailoa,~J.~P.; Kornbluth,~M.; Batzner,~S.; Samsonidze,~G.; Lam,~S.~T.;
  Vandermause,~J.; Ablitt,~C.; Molinari,~N.; Kozinsky,~B. A fast neural network
  approach for direct covariant forces prediction in complex multi-element
  extended systems. \emph{Nature Machine Intelligence} \textbf{2019}, \emph{1},
  471--479\relax
\mciteBstWouldAddEndPuncttrue
\mciteSetBstMidEndSepPunct{\mcitedefaultmidpunct}
{\mcitedefaultendpunct}{\mcitedefaultseppunct}\relax
\EndOfBibitem
\bibitem[Park \latin{et~al.}(2021)Park, Kornbluth, Vandermause, Wolverton,
  Kozinsky, and Mailoa]{Park2021-npj-direct-force}
Park,~C.~W.; Kornbluth,~M.; Vandermause,~J.; Wolverton,~C.; Kozinsky,~B.;
  Mailoa,~J.~P. Accurate and scalable graph neural network force field and
  molecular dynamics with direct force architecture. \emph{npj Computational
  Materials} \textbf{2021}, \emph{7}, 73\relax
\mciteBstWouldAddEndPuncttrue
\mciteSetBstMidEndSepPunct{\mcitedefaultmidpunct}
{\mcitedefaultendpunct}{\mcitedefaultseppunct}\relax
\EndOfBibitem
\bibitem[Husic \latin{et~al.}(2020)Husic, Charron, Lemm, Wang, Pérez,
  Majewski, Krämer, Chen, Olsson, de~Fabritiis, Noé, and
  Clementi]{Coarse-GNN-2020-JCP}
Husic,~B.~E.; Charron,~N.~E.; Lemm,~D.; Wang,~J.; Pérez,~A.; Majewski,~M.;
  Krämer,~A.; Chen,~Y.; Olsson,~S.; de~Fabritiis,~G.; Noé,~F.; Clementi,~C.
  Coarse graining molecular dynamics with graph neural networks. \emph{The
  Journal of Chemical Physics} \textbf{2020}, \emph{153}, 194101\relax
\mciteBstWouldAddEndPuncttrue
\mciteSetBstMidEndSepPunct{\mcitedefaultmidpunct}
{\mcitedefaultendpunct}{\mcitedefaultseppunct}\relax
\EndOfBibitem
\bibitem[Wang \latin{et~al.}(2019)Wang, Olsson, Wehmeyer, Pérez, Charron,
  de~Fabritiis, Noé, and Clementi]{CGNet-2019-Coarse}
Wang,~J.; Olsson,~S.; Wehmeyer,~C.; Pérez,~A.; Charron,~N.~E.;
  de~Fabritiis,~G.; Noé,~F.; Clementi,~C. Machine Learning of Coarse-Grained
  Molecular Dynamics Force Fields. \emph{ACS Central Science} \textbf{2019},
  \emph{5}, 755--767\relax
\mciteBstWouldAddEndPuncttrue
\mciteSetBstMidEndSepPunct{\mcitedefaultmidpunct}
{\mcitedefaultendpunct}{\mcitedefaultseppunct}\relax
\EndOfBibitem
\bibitem[Husic \latin{et~al.}(2020)Husic, Charron, Lemm, Wang, Pérez,
  Majewski, Krämer, Chen, Olsson, de~Fabritiis, Noé, and
  Clementi]{coarse-CGSchNet-2020}
Husic,~B.~E.; Charron,~N.~E.; Lemm,~D.; Wang,~J.; Pérez,~A.; Majewski,~M.;
  Krämer,~A.; Chen,~Y.; Olsson,~S.; de~Fabritiis,~G.; Noé,~F.; Clementi,~C.
  Coarse graining molecular dynamics with graph neural networks. \emph{The
  Journal of Chemical Physics} \textbf{2020}, \emph{153}, 194101\relax
\mciteBstWouldAddEndPuncttrue
\mciteSetBstMidEndSepPunct{\mcitedefaultmidpunct}
{\mcitedefaultendpunct}{\mcitedefaultseppunct}\relax
\EndOfBibitem
\bibitem[Unke \latin{et~al.}(2021)Unke, Chmiela, Gastegger, Sch{\"u}tt,
  Sauceda, and M{\"u}ller]{SpookyNet-2021}
Unke,~O.~T.; Chmiela,~S.; Gastegger,~M.; Sch{\"u}tt,~K.~T.; Sauceda,~H.~E.;
  M{\"u}ller,~K.-R. SpookyNet: Learning force fields with electronic degrees of
  freedom and nonlocal effects. \emph{Nature Communications} \textbf{2021},
  \emph{12}, 7273\relax
\mciteBstWouldAddEndPuncttrue
\mciteSetBstMidEndSepPunct{\mcitedefaultmidpunct}
{\mcitedefaultendpunct}{\mcitedefaultseppunct}\relax
\EndOfBibitem
\bibitem[Noé \latin{et~al.}(2020)Noé, Tkatchenko, Müller, and
  Clementi]{review-Noe-2020}
Noé,~F.; Tkatchenko,~A.; Müller,~K.-R.; Clementi,~C. Machine Learning for
  Molecular Simulation. \emph{Annual Review of Physical Chemistry}
  \textbf{2020}, \emph{71}, 361--390, PMID: 32092281\relax
\mciteBstWouldAddEndPuncttrue
\mciteSetBstMidEndSepPunct{\mcitedefaultmidpunct}
{\mcitedefaultendpunct}{\mcitedefaultseppunct}\relax
\EndOfBibitem
\bibitem[Gkeka \latin{et~al.}(2020)Gkeka, Stoltz, Barati~Farimani, Belkacemi,
  Ceriotti, Chodera, Dinner, Ferguson, Maillet, Minoux, Peter, Pietrucci,
  Silveira, Tkatchenko, Trstanova, Wiewiora, and Lelièvre]{review-Barati-2020}
Gkeka,~P.; Stoltz,~G.; Barati~Farimani,~A.; Belkacemi,~Z.; Ceriotti,~M.;
  Chodera,~J.~D.; Dinner,~A.~R.; Ferguson,~A.~L.; Maillet,~J.-B.; Minoux,~H.;
  Peter,~C.; Pietrucci,~F.; Silveira,~A.; Tkatchenko,~A.; Trstanova,~Z.;
  Wiewiora,~R.; Lelièvre,~T. Machine Learning Force Fields and Coarse-Grained
  Variables in Molecular Dynamics: Application to Materials and Biological
  Systems. \emph{Journal of Chemical Theory and Computation} \textbf{2020},
  \emph{16}, 4757--4775, PMID: 32559068\relax
\mciteBstWouldAddEndPuncttrue
\mciteSetBstMidEndSepPunct{\mcitedefaultmidpunct}
{\mcitedefaultendpunct}{\mcitedefaultseppunct}\relax
\EndOfBibitem
\bibitem[Unke \latin{et~al.}(2021)Unke, Chmiela, Sauceda, Gastegger, Poltavsky,
  Schütt, Tkatchenko, and Müller]{unke2021-ML-Forcefield}
Unke,~O.~T.; Chmiela,~S.; Sauceda,~H.~E.; Gastegger,~M.; Poltavsky,~I.;
  Schütt,~K.~T.; Tkatchenko,~A.; Müller,~K.-R. Machine Learning Force Fields.
  2021\relax
\mciteBstWouldAddEndPuncttrue
\mciteSetBstMidEndSepPunct{\mcitedefaultmidpunct}
{\mcitedefaultendpunct}{\mcitedefaultseppunct}\relax
\EndOfBibitem
\bibitem[Botu \latin{et~al.}(2017)Botu, Batra, Chapman, and
  Ramprasad]{ML-Forcefield-2017-Botu}
Botu,~V.; Batra,~R.; Chapman,~J.; Ramprasad,~R. Machine Learning Force Fields:
  Construction, Validation, and Outlook. \emph{The Journal of Physical
  Chemistry C} \textbf{2017}, \emph{121}, 511--522\relax
\mciteBstWouldAddEndPuncttrue
\mciteSetBstMidEndSepPunct{\mcitedefaultmidpunct}
{\mcitedefaultendpunct}{\mcitedefaultseppunct}\relax
\EndOfBibitem
\bibitem[Li \latin{et~al.}(2017)Li, Li, Pickard, Narayanan, Sen, Chan,
  Sankaranarayanan, Brooks, and Roux]{ML-forcefield-Ab-initio-2017}
Li,~Y.; Li,~H.; Pickard,~F.~C.; Narayanan,~B.; Sen,~F.~G.; Chan,~M. K.~Y.;
  Sankaranarayanan,~S. K. R.~S.; Brooks,~B.~R.; Roux,~B. Machine Learning Force
  Field Parameters from Ab Initio Data. \emph{Journal of Chemical Theory and
  Computation} \textbf{2017}, \emph{13}, 4492--4503, PMID: 28800233\relax
\mciteBstWouldAddEndPuncttrue
\mciteSetBstMidEndSepPunct{\mcitedefaultmidpunct}
{\mcitedefaultendpunct}{\mcitedefaultseppunct}\relax
\EndOfBibitem
\bibitem[Chmiela \latin{et~al.}(2018)Chmiela, Sauceda, M{\"u}ller, and
  Tkatchenko]{ML-forcefield-nature-2018}
Chmiela,~S.; Sauceda,~H.~E.; M{\"u}ller,~K.-R.; Tkatchenko,~A. Towards exact
  molecular dynamics simulations with machine-learned force fields.
  \emph{Nature Communications} \textbf{2018}, \emph{9}, 3887\relax
\mciteBstWouldAddEndPuncttrue
\mciteSetBstMidEndSepPunct{\mcitedefaultmidpunct}
{\mcitedefaultendpunct}{\mcitedefaultseppunct}\relax
\EndOfBibitem
\bibitem[Deringer \latin{et~al.}(2019)Deringer, Caro, and
  Csányi]{ML-interatomic-2019}
Deringer,~V.~L.; Caro,~M.~A.; Csányi,~G. Machine Learning Interatomic
  Potentials as Emerging Tools for Materials Science. \emph{Advanced Materials}
  \textbf{2019}, \emph{31}, 1902765\relax
\mciteBstWouldAddEndPuncttrue
\mciteSetBstMidEndSepPunct{\mcitedefaultmidpunct}
{\mcitedefaultendpunct}{\mcitedefaultseppunct}\relax
\EndOfBibitem
\bibitem[Behler(2016)]{ml-potential-review}
Behler,~J. Perspective: Machine learning potentials for atomistic simulations.
  \emph{The Journal of Chemical Physics} \textbf{2016}, \emph{145},
  170901\relax
\mciteBstWouldAddEndPuncttrue
\mciteSetBstMidEndSepPunct{\mcitedefaultmidpunct}
{\mcitedefaultendpunct}{\mcitedefaultseppunct}\relax
\EndOfBibitem
\bibitem[Eshet \latin{et~al.}(2010)Eshet, Khaliullin, K\"uhne, Behler, and
  Parrinello]{nnp-sodium}
Eshet,~H.; Khaliullin,~R.~Z.; K\"uhne,~T.~D.; Behler,~J.; Parrinello,~M. Ab
  initio quality neural-network potential for sodium. \emph{Phys. Rev. B}
  \textbf{2010}, \emph{81}, 184107\relax
\mciteBstWouldAddEndPuncttrue
\mciteSetBstMidEndSepPunct{\mcitedefaultmidpunct}
{\mcitedefaultendpunct}{\mcitedefaultseppunct}\relax
\EndOfBibitem
\bibitem[Artrith \latin{et~al.}(2011)Artrith, Morawietz, and Behler]{nnp-zinc}
Artrith,~N.; Morawietz,~T.; Behler,~J. High-dimensional neural-network
  potentials for multicomponent systems: Applications to zinc oxide.
  \emph{Phys. Rev. B} \textbf{2011}, \emph{83}, 153101\relax
\mciteBstWouldAddEndPuncttrue
\mciteSetBstMidEndSepPunct{\mcitedefaultmidpunct}
{\mcitedefaultendpunct}{\mcitedefaultseppunct}\relax
\EndOfBibitem
\bibitem[Wang \latin{et~al.}(2018)Wang, Zhang, Han, and E]{DeepPMD}
Wang,~H.; Zhang,~L.; Han,~J.; E,~W. DeePMD-kit: A deep learning package for
  many-body potential energy representation and molecular dynamics.
  \emph{Computer Physics Communications} \textbf{2018}, \emph{228},
  178--184\relax
\mciteBstWouldAddEndPuncttrue
\mciteSetBstMidEndSepPunct{\mcitedefaultmidpunct}
{\mcitedefaultendpunct}{\mcitedefaultseppunct}\relax
\EndOfBibitem
\bibitem[Artrith \latin{et~al.}(2017)Artrith, Urban, and
  Ceder]{ManySpecies-mlp}
Artrith,~N.; Urban,~A.; Ceder,~G. Efficient and accurate machine-learning
  interpolation of atomic energies in compositions with many species.
  \emph{Phys. Rev. B} \textbf{2017}, \emph{96}, 014112\relax
\mciteBstWouldAddEndPuncttrue
\mciteSetBstMidEndSepPunct{\mcitedefaultmidpunct}
{\mcitedefaultendpunct}{\mcitedefaultseppunct}\relax
\EndOfBibitem
\bibitem[Sanchez{-}Gonzalez \latin{et~al.}(2020)Sanchez{-}Gonzalez, Godwin,
  Pfaff, Ying, Leskovec, and Battaglia]{Sanchez-graph-GCN-ICML-2020}
Sanchez{-}Gonzalez,~A.; Godwin,~J.; Pfaff,~T.; Ying,~R.; Leskovec,~J.;
  Battaglia,~P.~W. Learning to Simulate Complex Physics with Graph Networks.
  Proceedings of the 37th International Conference on Machine Learning, {ICML}
  2020, 13-18 July 2020, Virtual Event. 2020; pp 8459--8468\relax
\mciteBstWouldAddEndPuncttrue
\mciteSetBstMidEndSepPunct{\mcitedefaultmidpunct}
{\mcitedefaultendpunct}{\mcitedefaultseppunct}\relax
\EndOfBibitem
\bibitem[Bapst \latin{et~al.}(2020)Bapst, Keck, Grabska-Barwi{\'{n}}ska,
  Donner, Cubuk, Schoenholz, Obika, Nelson, Back, Hassabis, and
  Kohli]{Bapst2020-glassy}
Bapst,~V.; Keck,~T.; Grabska-Barwi{\'{n}}ska,~A.; Donner,~C.; Cubuk,~E.~D.;
  Schoenholz,~S.~S.; Obika,~A.; Nelson,~A. W.~R.; Back,~T.; Hassabis,~D.;
  Kohli,~P. Unveiling the predictive power of static structure in glassy
  systems. \emph{Nature Physics} \textbf{2020}, \emph{16}, 448--454\relax
\mciteBstWouldAddEndPuncttrue
\mciteSetBstMidEndSepPunct{\mcitedefaultmidpunct}
{\mcitedefaultendpunct}{\mcitedefaultseppunct}\relax
\EndOfBibitem
\bibitem[Battaglia \latin{et~al.}(2016)Battaglia, Pascanu, Lai,
  Jimenez~Rezende, and kavukcuoglu]{Interaction-IN-NIPS-2016}
Battaglia,~P.; Pascanu,~R.; Lai,~M.; Jimenez~Rezende,~D.; kavukcuoglu,~k.
  Interaction Networks for Learning about Objects, Relations and Physics.
  Advances in Neural Information Processing Systems. 2016\relax
\mciteBstWouldAddEndPuncttrue
\mciteSetBstMidEndSepPunct{\mcitedefaultmidpunct}
{\mcitedefaultendpunct}{\mcitedefaultseppunct}\relax
\EndOfBibitem
\bibitem[Bart{\'o}k \latin{et~al.}(2017)Bart{\'o}k, De, Poelking, Bernstein,
  Kermode, Cs{\'a}nyi, and Ceriotti]{Bartok-ML-modeling}
Bart{\'o}k,~A.~P.; De,~S.; Poelking,~C.; Bernstein,~N.; Kermode,~J.~R.;
  Cs{\'a}nyi,~G.; Ceriotti,~M. Machine learning unifies the modeling of
  materials and molecules. \emph{Science Advances} \textbf{2017},
  \emph{3}\relax
\mciteBstWouldAddEndPuncttrue
\mciteSetBstMidEndSepPunct{\mcitedefaultmidpunct}
{\mcitedefaultendpunct}{\mcitedefaultseppunct}\relax
\EndOfBibitem
\bibitem[Chen \latin{et~al.}(2019)Chen, Ye, Zuo, Zheng, and
  Ong]{GN-universal-molecules-2019}
Chen,~C.; Ye,~W.; Zuo,~Y.; Zheng,~C.; Ong,~S.~P. Graph Networks as a Universal
  Machine Learning Framework for Molecules and Crystals. \emph{Chemistry of
  Materials} \textbf{2019}, \emph{31}, 3564--3572\relax
\mciteBstWouldAddEndPuncttrue
\mciteSetBstMidEndSepPunct{\mcitedefaultmidpunct}
{\mcitedefaultendpunct}{\mcitedefaultseppunct}\relax
\EndOfBibitem
\bibitem[Smith \latin{et~al.}(2017)Smith, Isayev, and
  Roitberg]{ANI-1-Smith-2017}
Smith,~J.~S.; Isayev,~O.; Roitberg,~A.~E. ANI-1: an extensible neural network
  potential with DFT accuracy at force field computational cost. \emph{Chem.
  Sci.} \textbf{2017}, \emph{8}, 3192--3203\relax
\mciteBstWouldAddEndPuncttrue
\mciteSetBstMidEndSepPunct{\mcitedefaultmidpunct}
{\mcitedefaultendpunct}{\mcitedefaultseppunct}\relax
\EndOfBibitem
\bibitem[Christensen \latin{et~al.}(2020)Christensen, Bratholm, Faber, and
  Anatole~von Lilienfeld]{FCHL-revisited-2020}
Christensen,~A.~S.; Bratholm,~L.~A.; Faber,~F.~A.; Anatole~von Lilienfeld,~O.
  FCHL revisited: Faster and more accurate quantum machine learning. \emph{The
  Journal of Chemical Physics} \textbf{2020}, \emph{152}, 044107\relax
\mciteBstWouldAddEndPuncttrue
\mciteSetBstMidEndSepPunct{\mcitedefaultmidpunct}
{\mcitedefaultendpunct}{\mcitedefaultseppunct}\relax
\EndOfBibitem
\bibitem[Zhang \latin{et~al.}(2018)Zhang, Han, Wang, Car, and E]{DPMD-2018}
Zhang,~L.; Han,~J.; Wang,~H.; Car,~R.; E,~W. Deep Potential Molecular Dynamics:
  A Scalable Model with the Accuracy of Quantum Mechanics. \emph{Phys. Rev.
  Lett.} \textbf{2018}, \emph{120}, 143001\relax
\mciteBstWouldAddEndPuncttrue
\mciteSetBstMidEndSepPunct{\mcitedefaultmidpunct}
{\mcitedefaultendpunct}{\mcitedefaultseppunct}\relax
\EndOfBibitem
\bibitem[Carbogno \latin{et~al.}(2008)Carbogno, Behler, Gro\ss{}, and
  Reuter]{fingerprints-o2}
Carbogno,~C.; Behler,~J.; Gro\ss{},~A.; Reuter,~K. Fingerprints for
  Spin-Selection Rules in the Interaction Dynamics of ${\mathrm{O}}_{2}$ at
  Al(111). \emph{Phys. Rev. Lett.} \textbf{2008}, \emph{101}, 096104\relax
\mciteBstWouldAddEndPuncttrue
\mciteSetBstMidEndSepPunct{\mcitedefaultmidpunct}
{\mcitedefaultendpunct}{\mcitedefaultseppunct}\relax
\EndOfBibitem
\bibitem[Behler(2011)]{Behler-2011-ACSF}
Behler,~J. Atom-centered symmetry functions for constructing high-dimensional
  neural network potentials. \emph{The Journal of Chemical Physics}
  \textbf{2011}, \emph{134}, 074106\relax
\mciteBstWouldAddEndPuncttrue
\mciteSetBstMidEndSepPunct{\mcitedefaultmidpunct}
{\mcitedefaultendpunct}{\mcitedefaultseppunct}\relax
\EndOfBibitem
\bibitem[Behler and Parrinello(2007)Behler, and
  Parrinello]{Behler-NN-Energy-2007}
Behler,~J.; Parrinello,~M. Generalized Neural-Network Representation of
  High-Dimensional Potential-Energy Surfaces. \emph{Phys. Rev. Lett.}
  \textbf{2007}, \emph{98}, 146401\relax
\mciteBstWouldAddEndPuncttrue
\mciteSetBstMidEndSepPunct{\mcitedefaultmidpunct}
{\mcitedefaultendpunct}{\mcitedefaultseppunct}\relax
\EndOfBibitem
\bibitem[Behler(2017)]{Behler-BPNN2017}
Behler,~J. First Principles Neural Network Potentials for Reactive Simulations
  of Large Molecular and Condensed Systems. \emph{Angewandte Chemie
  International Edition} \textbf{2017}, \emph{56}, 12828--12840\relax
\mciteBstWouldAddEndPuncttrue
\mciteSetBstMidEndSepPunct{\mcitedefaultmidpunct}
{\mcitedefaultendpunct}{\mcitedefaultseppunct}\relax
\EndOfBibitem
\bibitem[Duvenaud \latin{et~al.}(2015)Duvenaud, Maclaurin, Iparraguirre,
  Bombarell, Hirzel, Aspuru-Guzik, and Adams]{Conv_fingerprints_NIPS-2015}
Duvenaud,~D.~K.; Maclaurin,~D.; Iparraguirre,~J.; Bombarell,~R.; Hirzel,~T.;
  Aspuru-Guzik,~A.; Adams,~R.~P. Convolutional Networks on Graphs for Learning
  Molecular Fingerprints. Advances in Neural Information Processing Systems.
  2015\relax
\mciteBstWouldAddEndPuncttrue
\mciteSetBstMidEndSepPunct{\mcitedefaultmidpunct}
{\mcitedefaultendpunct}{\mcitedefaultseppunct}\relax
\EndOfBibitem
\bibitem[Kearnes \latin{et~al.}(2016)Kearnes, McCloskey, Berndl, Pande, and
  Riley]{Molecular-graph-2016}
Kearnes,~S.; McCloskey,~K.; Berndl,~M.; Pande,~V.; Riley,~P. Molecular graph
  convolutions: moving beyond fingerprints. \emph{Journal of Computer-Aided
  Molecular Design} \textbf{2016}, \emph{30}, 595--608\relax
\mciteBstWouldAddEndPuncttrue
\mciteSetBstMidEndSepPunct{\mcitedefaultmidpunct}
{\mcitedefaultendpunct}{\mcitedefaultseppunct}\relax
\EndOfBibitem
\bibitem[Sch{\"u}tt \latin{et~al.}(2017)Sch{\"u}tt, Arbabzadah, Chmiela,
  M{\"u}ller, and Tkatchenko]{Schutt-insight-DTNN-nature}
Sch{\"u}tt,~K.~T.; Arbabzadah,~F.; Chmiela,~S.; M{\"u}ller,~K.~R.;
  Tkatchenko,~A. Quantum-chemical insights from deep tensor neural networks.
  \emph{Nature Communications} \textbf{2017}, \emph{8}, 13890\relax
\mciteBstWouldAddEndPuncttrue
\mciteSetBstMidEndSepPunct{\mcitedefaultmidpunct}
{\mcitedefaultendpunct}{\mcitedefaultseppunct}\relax
\EndOfBibitem
\bibitem[Unke and Meuwly(2019)Unke, and Meuwly]{PhysNet-2019-Unke}
Unke,~O.~T.; Meuwly,~M. PhysNet: A Neural Network for Predicting Energies,
  Forces, Dipole Moments, and Partial Charges. \emph{Journal of Chemical Theory
  and Computation} \textbf{2019}, \emph{15}, 3678--3693, PMID: 31042390\relax
\mciteBstWouldAddEndPuncttrue
\mciteSetBstMidEndSepPunct{\mcitedefaultmidpunct}
{\mcitedefaultendpunct}{\mcitedefaultseppunct}\relax
\EndOfBibitem
\bibitem[Schütt \latin{et~al.}(2017)Schütt, Kindermans, Sauceda, Chmiela,
  Tkatchenko, and Müller]{schutt2017schnet}
Schütt,~K.~T.; Kindermans,~P.-J.; Sauceda,~H.~E.; Chmiela,~S.; Tkatchenko,~A.;
  Müller,~K.-R. SchNet: A continuous-filter convolutional neural network for
  modeling quantum interactions. 2017\relax
\mciteBstWouldAddEndPuncttrue
\mciteSetBstMidEndSepPunct{\mcitedefaultmidpunct}
{\mcitedefaultendpunct}{\mcitedefaultseppunct}\relax
\EndOfBibitem
\bibitem[Lubbers \latin{et~al.}(2018)Lubbers, Smith, and
  Barros]{Hierarchical-2018-JCP}
Lubbers,~N.; Smith,~J.~S.; Barros,~K. Hierarchical modeling of molecular
  energies using a deep neural network. \emph{The Journal of Chemical Physics}
  \textbf{2018}, \emph{148}, 241715\relax
\mciteBstWouldAddEndPuncttrue
\mciteSetBstMidEndSepPunct{\mcitedefaultmidpunct}
{\mcitedefaultendpunct}{\mcitedefaultseppunct}\relax
\EndOfBibitem
\bibitem[Imbalzano \latin{et~al.}(2018)Imbalzano, Anelli, Giofré, Klees,
  Behler, and Ceriotti]{atomic-fingerprints-2018-ML}
Imbalzano,~G.; Anelli,~A.; Giofré,~D.; Klees,~S.; Behler,~J.; Ceriotti,~M.
  Automatic selection of atomic fingerprints and reference configurations for
  machine-learning potentials. \emph{The Journal of Chemical Physics}
  \textbf{2018}, \emph{148}, 241730\relax
\mciteBstWouldAddEndPuncttrue
\mciteSetBstMidEndSepPunct{\mcitedefaultmidpunct}
{\mcitedefaultendpunct}{\mcitedefaultseppunct}\relax
\EndOfBibitem
\bibitem[Pfaff \latin{et~al.}(2021)Pfaff, Fortunato, Sanchez-Gonzalez, and
  Battaglia]{Mesh-based-GNN-ICLR-2021}
Pfaff,~T.; Fortunato,~M.; Sanchez-Gonzalez,~A.; Battaglia,~P. Learning
  Mesh-Based Simulation with Graph Networks. International Conference on
  Learning Representations. 2021\relax
\mciteBstWouldAddEndPuncttrue
\mciteSetBstMidEndSepPunct{\mcitedefaultmidpunct}
{\mcitedefaultendpunct}{\mcitedefaultseppunct}\relax
\EndOfBibitem
\bibitem[Shlomi \latin{et~al.}(2021)Shlomi, Battaglia, and
  Vlimant]{GNN-particle-review-2021}
Shlomi,~J.; Battaglia,~P.; Vlimant,~J.-R. Graph neural networks in particle
  physics. \emph{Machine Learning: Science and Technology} \textbf{2021},
  \emph{2}, 021001\relax
\mciteBstWouldAddEndPuncttrue
\mciteSetBstMidEndSepPunct{\mcitedefaultmidpunct}
{\mcitedefaultendpunct}{\mcitedefaultseppunct}\relax
\EndOfBibitem
\bibitem[Li and Farimani(2022)Li, and Farimani]{LI2022}
Li,~Z.; Farimani,~A.~B. Graph neural network-accelerated Lagrangian fluid
  simulation. \emph{Computers \& Graphics} \textbf{2022}, \relax
\mciteBstWouldAddEndPunctfalse
\mciteSetBstMidEndSepPunct{\mcitedefaultmidpunct}
{}{\mcitedefaultseppunct}\relax
\EndOfBibitem
\bibitem[Ogoke \latin{et~al.}(2020)Ogoke, Meidani, Hashemi, and
  Farimani]{ogoke2020graph}
Ogoke,~F.; Meidani,~K.; Hashemi,~A.; Farimani,~A.~B. Graph Convolutional Neural
  Networks for Body Force Prediction. 2020\relax
\mciteBstWouldAddEndPuncttrue
\mciteSetBstMidEndSepPunct{\mcitedefaultmidpunct}
{\mcitedefaultendpunct}{\mcitedefaultseppunct}\relax
\EndOfBibitem
\bibitem[Brandstetter \latin{et~al.}(2022)Brandstetter, Worrall, and
  Welling]{brandstetter2022message}
Brandstetter,~J.; Worrall,~D.~E.; Welling,~M. Message Passing Neural {PDE}
  Solvers. International Conference on Learning Representations. 2022\relax
\mciteBstWouldAddEndPuncttrue
\mciteSetBstMidEndSepPunct{\mcitedefaultmidpunct}
{\mcitedefaultendpunct}{\mcitedefaultseppunct}\relax
\EndOfBibitem
\bibitem[de~Avila Belbute-Peres \latin{et~al.}(2020)de~Avila Belbute-Peres,
  Economon, and Kolter]{belbuteperes2020combining}
de~Avila Belbute-Peres,~F.; Economon,~T.~D.; Kolter,~J.~Z. Combining
  Differentiable PDE Solvers and Graph Neural Networks for Fluid Flow
  Prediction. 2020\relax
\mciteBstWouldAddEndPuncttrue
\mciteSetBstMidEndSepPunct{\mcitedefaultmidpunct}
{\mcitedefaultendpunct}{\mcitedefaultseppunct}\relax
\EndOfBibitem
\bibitem[Xie and Grossman(2018)Xie, and Grossman]{CGCNN-Grossman-2018}
Xie,~T.; Grossman,~J.~C. Crystal Graph Convolutional Neural Networks for an
  Accurate and Interpretable Prediction of Material Properties. \emph{Phys.
  Rev. Lett.} \textbf{2018}, \emph{120}, 145301\relax
\mciteBstWouldAddEndPuncttrue
\mciteSetBstMidEndSepPunct{\mcitedefaultmidpunct}
{\mcitedefaultendpunct}{\mcitedefaultseppunct}\relax
\EndOfBibitem
\bibitem[Karamad \latin{et~al.}(2020)Karamad, Magar, Shi, Siahrostami, Gates,
  and Barati~Farimani]{OGCNN-Karamad-2020}
Karamad,~M.; Magar,~R.; Shi,~Y.; Siahrostami,~S.; Gates,~I.~D.;
  Barati~Farimani,~A. Orbital graph convolutional neural network for material
  property prediction. \emph{Phys. Rev. Materials} \textbf{2020}, \emph{4},
  093801\relax
\mciteBstWouldAddEndPuncttrue
\mciteSetBstMidEndSepPunct{\mcitedefaultmidpunct}
{\mcitedefaultendpunct}{\mcitedefaultseppunct}\relax
\EndOfBibitem
\bibitem[Wang \latin{et~al.}(2021)Wang, Wang, Cao, and
  Farimani]{wang2021molclr}
Wang,~Y.; Wang,~J.; Cao,~Z.; Farimani,~A.~B. MolCLR: Molecular Contrastive
  Learning of Representations via Graph Neural Networks. 2021\relax
\mciteBstWouldAddEndPuncttrue
\mciteSetBstMidEndSepPunct{\mcitedefaultmidpunct}
{\mcitedefaultendpunct}{\mcitedefaultseppunct}\relax
\EndOfBibitem
\bibitem[Gilmer \latin{et~al.}(2017)Gilmer, Schoenholz, Riley, Vinyals, and
  Dahl]{Gilmer-MP-ICML-2017}
Gilmer,~J.; Schoenholz,~S.~S.; Riley,~P.~F.; Vinyals,~O.; Dahl,~G.~E. Neural
  Message Passing for Quantum Chemistry. Proceedings of the 34th International
  Conference on Machine Learning. 2017; pp 1263--1272\relax
\mciteBstWouldAddEndPuncttrue
\mciteSetBstMidEndSepPunct{\mcitedefaultmidpunct}
{\mcitedefaultendpunct}{\mcitedefaultseppunct}\relax
\EndOfBibitem
\bibitem[Klicpera \latin{et~al.}(2020)Klicpera, Groß, and
  Günnemann]{Directional-MPNN-ICLR-2020}
Klicpera,~J.; Groß,~J.; Günnemann,~S. Directional Message Passing for
  Molecular Graphs. International Conference on Learning Representations.
  2020\relax
\mciteBstWouldAddEndPuncttrue
\mciteSetBstMidEndSepPunct{\mcitedefaultmidpunct}
{\mcitedefaultendpunct}{\mcitedefaultseppunct}\relax
\EndOfBibitem
\bibitem[Klicpera \latin{et~al.}(2020)Klicpera, Giri, Margraf, and
  Gunnemann]{DimeNet++-directional-2020}
Klicpera,~J.; Giri,~S.; Margraf,~J.~T.; Gunnemann,~S. Fast and
  Uncertainty-Aware Directional Message Passing for Non-Equilibrium Molecules.
  \emph{ArXiv} \textbf{2020}, \emph{abs/2011.14115}\relax
\mciteBstWouldAddEndPuncttrue
\mciteSetBstMidEndSepPunct{\mcitedefaultmidpunct}
{\mcitedefaultendpunct}{\mcitedefaultseppunct}\relax
\EndOfBibitem
\bibitem[Schoenholz and Cubuk(2020)Schoenholz, and Cubuk]{jaxmd2020}
Schoenholz,~S.~S.; Cubuk,~E.~D. JAX M.D. A Framework for Differentiable
  Physics. Advances in Neural Information Processing Systems. 2020\relax
\mciteBstWouldAddEndPuncttrue
\mciteSetBstMidEndSepPunct{\mcitedefaultmidpunct}
{\mcitedefaultendpunct}{\mcitedefaultseppunct}\relax
\EndOfBibitem
\bibitem[Doerr \latin{et~al.}(2021)Doerr, Majewski, Pérez, Krämer, Clementi,
  Noe, Giorgino, and De~Fabritiis]{TorchMD-2021-JCTC}
Doerr,~S.; Majewski,~M.; Pérez,~A.; Krämer,~A.; Clementi,~C.; Noe,~F.;
  Giorgino,~T.; De~Fabritiis,~G. TorchMD: A Deep Learning Framework for
  Molecular Simulations. \emph{Journal of Chemical Theory and Computation}
  \textbf{2021}, \emph{17}, 2355--2363, PMID: 33729795\relax
\mciteBstWouldAddEndPuncttrue
\mciteSetBstMidEndSepPunct{\mcitedefaultmidpunct}
{\mcitedefaultendpunct}{\mcitedefaultseppunct}\relax
\EndOfBibitem
\bibitem[Schütt \latin{et~al.}(2019)Schütt, Kessel, Gastegger, Nicoli,
  Tkatchenko, and Müller]{SchNetPack-JCTC-2019}
Schütt,~K.~T.; Kessel,~P.; Gastegger,~M.; Nicoli,~K.~A.; Tkatchenko,~A.;
  Müller,~K.-R. SchNetPack: A Deep Learning Toolbox For Atomistic Systems.
  \emph{Journal of Chemical Theory and Computation} \textbf{2019}, \emph{15},
  448--455\relax
\mciteBstWouldAddEndPuncttrue
\mciteSetBstMidEndSepPunct{\mcitedefaultmidpunct}
{\mcitedefaultendpunct}{\mcitedefaultseppunct}\relax
\EndOfBibitem
\bibitem[Wang \latin{et~al.}(2020)Wang, Axelrod, and
  G{\'o}mez-Bombarelli]{wang-2020-differentiable}
Wang,~W.; Axelrod,~S.; G{\'o}mez-Bombarelli,~R. Differentiable Molecular
  Simulations for Control and Learning. ICLR 2020 Workshop on Integration of
  Deep Neural Models and Differential Equations. 2020\relax
\mciteBstWouldAddEndPuncttrue
\mciteSetBstMidEndSepPunct{\mcitedefaultmidpunct}
{\mcitedefaultendpunct}{\mcitedefaultseppunct}\relax
\EndOfBibitem
\bibitem[Pukrittayakamee \latin{et~al.}(2009)Pukrittayakamee, Malshe, Hagan,
  Raff, Narulkar, Bukkapatnum, and Komanduri]{PES-force-simultaneous-2009}
Pukrittayakamee,~A.; Malshe,~M.; Hagan,~M.; Raff,~L.~M.; Narulkar,~R.;
  Bukkapatnum,~S.; Komanduri,~R. Simultaneous fitting of a potential-energy
  surface and its corresponding force fields using feedforward neural networks.
  \emph{The Journal of Chemical Physics} \textbf{2009}, \emph{130},
  134101\relax
\mciteBstWouldAddEndPuncttrue
\mciteSetBstMidEndSepPunct{\mcitedefaultmidpunct}
{\mcitedefaultendpunct}{\mcitedefaultseppunct}\relax
\EndOfBibitem
\bibitem[Chanussot \latin{et~al.}(2021)Chanussot, Das, Goyal, Lavril, Shuaibi,
  Riviere, Tran, Heras-Domingo, Ho, Hu, Palizhati, Sriram, Wood, Yoon, Parikh,
  Zitnick, and Ulissi]{OC20-Catalyst-2021}
Chanussot,~L.; Das,~A.; Goyal,~S.; Lavril,~T.; Shuaibi,~M.; Riviere,~M.;
  Tran,~K.; Heras-Domingo,~J.; Ho,~C.; Hu,~W.; Palizhati,~A.; Sriram,~A.;
  Wood,~B.; Yoon,~J.; Parikh,~D.; Zitnick,~C.~L.; Ulissi,~Z. Open Catalyst 2020
  (OC20) Dataset and Community Challenges. \emph{ACS Catalysis} \textbf{2021},
  \emph{11}, 6059–6072\relax
\mciteBstWouldAddEndPuncttrue
\mciteSetBstMidEndSepPunct{\mcitedefaultmidpunct}
{\mcitedefaultendpunct}{\mcitedefaultseppunct}\relax
\EndOfBibitem
\bibitem[Morawietz and Behler(2013)Morawietz, and
  Behler]{Morawietz-DFT-NN-2013}
Morawietz,~T.; Behler,~J. A Density-Functional Theory-Based Neural Network
  Potential for Water Clusters Including van der Waals Corrections. \emph{The
  Journal of Physical Chemistry A} \textbf{2013}, \emph{117}, 7356--7366, PMID:
  23557541\relax
\mciteBstWouldAddEndPuncttrue
\mciteSetBstMidEndSepPunct{\mcitedefaultmidpunct}
{\mcitedefaultendpunct}{\mcitedefaultseppunct}\relax
\EndOfBibitem
\bibitem[Morawietz \latin{et~al.}(2016)Morawietz, Singraber, Dellago, and
  Behler]{Morawietz8368}
Morawietz,~T.; Singraber,~A.; Dellago,~C.; Behler,~J. How van der Waals
  interactions determine the unique properties of water. \emph{Proceedings of
  the National Academy of Sciences} \textbf{2016}, \emph{113}, 8368--8373\relax
\mciteBstWouldAddEndPuncttrue
\mciteSetBstMidEndSepPunct{\mcitedefaultmidpunct}
{\mcitedefaultendpunct}{\mcitedefaultseppunct}\relax
\EndOfBibitem
\bibitem[Cheng \latin{et~al.}(2019)Cheng, Engel, Behler, Dellago, and
  Ceriotti]{ab-inito-water-2019-pnas}
Cheng,~B.; Engel,~E.~A.; Behler,~J.; Dellago,~C.; Ceriotti,~M. Ab initio
  thermodynamics of liquid and solid water. \emph{Proceedings of the National
  Academy of Sciences} \textbf{2019}, \emph{116}, 1110--1115\relax
\mciteBstWouldAddEndPuncttrue
\mciteSetBstMidEndSepPunct{\mcitedefaultmidpunct}
{\mcitedefaultendpunct}{\mcitedefaultseppunct}\relax
\EndOfBibitem
\bibitem[Ba \latin{et~al.}(2016)Ba, Kiros, and Hinton]{ba2016layer}
Ba,~J.~L.; Kiros,~J.~R.; Hinton,~G.~E. Layer Normalization. 2016\relax
\mciteBstWouldAddEndPuncttrue
\mciteSetBstMidEndSepPunct{\mcitedefaultmidpunct}
{\mcitedefaultendpunct}{\mcitedefaultseppunct}\relax
\EndOfBibitem
\bibitem[He \latin{et~al.}(2015)He, Zhang, Ren, and Sun]{he2015deep}
He,~K.; Zhang,~X.; Ren,~S.; Sun,~J. Deep Residual Learning for Image
  Recognition. 2015\relax
\mciteBstWouldAddEndPuncttrue
\mciteSetBstMidEndSepPunct{\mcitedefaultmidpunct}
{\mcitedefaultendpunct}{\mcitedefaultseppunct}\relax
\EndOfBibitem
\bibitem[Paszke \latin{et~al.}(2019)Paszke, Gross, Massa, Lerer, Bradbury,
  Chanan, Killeen, Lin, Gimelshein, Antiga, Desmaison, Kopf, Yang, DeVito,
  Raison, Tejani, Chilamkurthy, Steiner, Fang, Bai, and Chintala]{PyTorch2019}
Paszke,~A.; Gross,~S.; Massa,~F.; Lerer,~A.; Bradbury,~J.; Chanan,~G.;
  Killeen,~T.; Lin,~Z.; Gimelshein,~N.; Antiga,~L.; Desmaison,~A.; Kopf,~A.;
  Yang,~E.; DeVito,~Z.; Raison,~M.; Tejani,~A.; Chilamkurthy,~S.; Steiner,~B.;
  Fang,~L.; Bai,~J.; Chintala,~S. In \emph{Advances in Neural Information
  Processing Systems 32}; Wallach,~H., Larochelle,~H., Beygelzimer,~A.,
  d\textquotesingle Alch\'{e}-Buc,~F., Fox,~E., Garnett,~R., Eds.; Curran
  Associates, Inc., 2019; pp 8024--8035\relax
\mciteBstWouldAddEndPuncttrue
\mciteSetBstMidEndSepPunct{\mcitedefaultmidpunct}
{\mcitedefaultendpunct}{\mcitedefaultseppunct}\relax
\EndOfBibitem
\bibitem[Wang \latin{et~al.}(2019)Wang, Zheng, Ye, Gan, Li, Song, Zhou, Ma, Yu,
  Gai, Xiao, He, Karypis, Li, and Zhang]{wang2019dgl}
Wang,~M.; Zheng,~D.; Ye,~Z.; Gan,~Q.; Li,~M.; Song,~X.; Zhou,~J.; Ma,~C.;
  Yu,~L.; Gai,~Y.; Xiao,~T.; He,~T.; Karypis,~G.; Li,~J.; Zhang,~Z. Deep Graph
  Library: A Graph-Centric, Highly-Performant Package for Graph Neural
  Networks. \emph{arXiv preprint arXiv:1909.01315} \textbf{2019}, \relax
\mciteBstWouldAddEndPunctfalse
\mciteSetBstMidEndSepPunct{\mcitedefaultmidpunct}
{}{\mcitedefaultseppunct}\relax
\EndOfBibitem
\bibitem[Eastman \latin{et~al.}(2017)Eastman, Swails, Chodera, McGibbon, Zhao,
  Beauchamp, Wang, Simmonett, Harrigan, Stern, Wiewiora, Brooks, and
  Pande]{OpenMM}
Eastman,~P.; Swails,~J.; Chodera,~J.~D.; McGibbon,~R.~T.; Zhao,~Y.;
  Beauchamp,~K.~A.; Wang,~L.-P.; Simmonett,~A.~C.; Harrigan,~M.~P.;
  Stern,~C.~D.; Wiewiora,~R.~P.; Brooks,~B.~R.; Pande,~V.~S. OpenMM 7: Rapid
  development of high performance algorithms for molecular dynamics. \emph{PLOS
  Computational Biology} \textbf{2017}, \emph{13}, 1--17\relax
\mciteBstWouldAddEndPuncttrue
\mciteSetBstMidEndSepPunct{\mcitedefaultmidpunct}
{\mcitedefaultendpunct}{\mcitedefaultseppunct}\relax
\EndOfBibitem
\bibitem[Lennard-Jones(1931)]{Lennard_Jones_1931}
Lennard-Jones,~J.~E. Cohesion. \emph{Proceedings of the Physical Society}
  \textbf{1931}, \emph{43}, 461--482\relax
\mciteBstWouldAddEndPuncttrue
\mciteSetBstMidEndSepPunct{\mcitedefaultmidpunct}
{\mcitedefaultendpunct}{\mcitedefaultseppunct}\relax
\EndOfBibitem
\bibitem[Jorgensen \latin{et~al.}(1983)Jorgensen, Chandrasekhar, Madura, Impey,
  and Klein]{TIP3P}
Jorgensen,~W.~L.; Chandrasekhar,~J.; Madura,~J.~D.; Impey,~R.~W.; Klein,~M.~L.
  Comparison of simple potential functions for simulating liquid water.
  \emph{The Journal of Chemical Physics} \textbf{1983}, \emph{79},
  926--935\relax
\mciteBstWouldAddEndPuncttrue
\mciteSetBstMidEndSepPunct{\mcitedefaultmidpunct}
{\mcitedefaultendpunct}{\mcitedefaultseppunct}\relax
\EndOfBibitem
\bibitem[Horn \latin{et~al.}(2004)Horn, Swope, Pitera, Madura, Dick, Hura, and
  Head-Gordon]{TIP4PEw}
Horn,~H.~W.; Swope,~W.~C.; Pitera,~J.~W.; Madura,~J.~D.; Dick,~T.~J.;
  Hura,~G.~L.; Head-Gordon,~T. Development of an improved four-site water model
  for biomolecular simulations: TIP4P-Ew. \emph{The Journal of Chemical
  Physics} \textbf{2004}, \emph{120}, 9665--9678\relax
\mciteBstWouldAddEndPuncttrue
\mciteSetBstMidEndSepPunct{\mcitedefaultmidpunct}
{\mcitedefaultendpunct}{\mcitedefaultseppunct}\relax
\EndOfBibitem
\bibitem[Nosé(1984)]{Nose}
Nosé,~S. A unified formulation of the constant temperature molecular dynamics
  methods. \emph{The Journal of Chemical Physics} \textbf{1984}, \emph{81},
  511--519\relax
\mciteBstWouldAddEndPuncttrue
\mciteSetBstMidEndSepPunct{\mcitedefaultmidpunct}
{\mcitedefaultendpunct}{\mcitedefaultseppunct}\relax
\EndOfBibitem
\bibitem[Hoover(1985)]{Hoover}
Hoover,~W.~G. Canonical dynamics: Equilibrium phase-space distributions.
  \emph{Phys. Rev. A} \textbf{1985}, \emph{31}, 1695--1697\relax
\mciteBstWouldAddEndPuncttrue
\mciteSetBstMidEndSepPunct{\mcitedefaultmidpunct}
{\mcitedefaultendpunct}{\mcitedefaultseppunct}\relax
\EndOfBibitem
\bibitem[Hammer \latin{et~al.}(1999)Hammer, Hansen, and N\o{}rskov]{RPBE}
Hammer,~B.; Hansen,~L.~B.; N\o{}rskov,~J.~K. Improved adsorption energetics
  within density-functional theory using revised Perdew-Burke-Ernzerhof
  functionals. \emph{Phys. Rev. B} \textbf{1999}, \emph{59}, 7413--7421\relax
\mciteBstWouldAddEndPuncttrue
\mciteSetBstMidEndSepPunct{\mcitedefaultmidpunct}
{\mcitedefaultendpunct}{\mcitedefaultseppunct}\relax
\EndOfBibitem
\bibitem[Grimme \latin{et~al.}(2010)Grimme, Antony, Ehrlich, and
  Krieg]{D3correction}
Grimme,~S.; Antony,~J.; Ehrlich,~S.; Krieg,~H. A consistent and accurate ab
  initio parametrization of density functional dispersion correction (DFT-D)
  for the 94 elements H-Pu. \emph{The Journal of Chemical Physics}
  \textbf{2010}, \emph{132}, 154104\relax
\mciteBstWouldAddEndPuncttrue
\mciteSetBstMidEndSepPunct{\mcitedefaultmidpunct}
{\mcitedefaultendpunct}{\mcitedefaultseppunct}\relax
\EndOfBibitem
\bibitem[Kingma and Ba(2017)Kingma, and Ba]{kingma2017adam}
Kingma,~D.~P.; Ba,~J. Adam: A Method for Stochastic Optimization. 2017\relax
\mciteBstWouldAddEndPuncttrue
\mciteSetBstMidEndSepPunct{\mcitedefaultmidpunct}
{\mcitedefaultendpunct}{\mcitedefaultseppunct}\relax
\EndOfBibitem
\bibitem[Hendrycks and Gimpel(2020)Hendrycks, and
  Gimpel]{hendrycks2020gaussian}
Hendrycks,~D.; Gimpel,~K. Gaussian Error Linear Units (GELUs). 2020\relax
\mciteBstWouldAddEndPuncttrue
\mciteSetBstMidEndSepPunct{\mcitedefaultmidpunct}
{\mcitedefaultendpunct}{\mcitedefaultseppunct}\relax
\EndOfBibitem
\bibitem[Humphrey \latin{et~al.}(1996)Humphrey, Dalke, and Schulten]{VMD}
Humphrey,~W.; Dalke,~A.; Schulten,~K. VMD: Visual molecular dynamics.
  \emph{Journal of Molecular Graphics} \textbf{1996}, \emph{14}, 33 -- 38\relax
\mciteBstWouldAddEndPuncttrue
\mciteSetBstMidEndSepPunct{\mcitedefaultmidpunct}
{\mcitedefaultendpunct}{\mcitedefaultseppunct}\relax
\EndOfBibitem
\bibitem[Kondati~Natarajan \latin{et~al.}(2015)Kondati~Natarajan, Morawietz,
  and Behler]{Morawietz-NN-waterPES}
Kondati~Natarajan,~S.; Morawietz,~T.; Behler,~J. Representing the
  potential-energy surface of protonated water clusters by high-dimensional
  neural network potentials. \emph{Phys. Chem. Chem. Phys.} \textbf{2015},
  \emph{17}, 8356--8371\relax
\mciteBstWouldAddEndPuncttrue
\mciteSetBstMidEndSepPunct{\mcitedefaultmidpunct}
{\mcitedefaultendpunct}{\mcitedefaultseppunct}\relax
\EndOfBibitem
\bibitem[Morawietz \latin{et~al.}(2012)Morawietz, Sharma, and
  Behler]{Morawietz-NN-waterdimer}
Morawietz,~T.; Sharma,~V.; Behler,~J. A neural network potential-energy surface
  for the water dimer based on environment-dependent atomic energies and
  charges. \emph{The Journal of Chemical Physics} \textbf{2012}, \emph{136},
  064103\relax
\mciteBstWouldAddEndPuncttrue
\mciteSetBstMidEndSepPunct{\mcitedefaultmidpunct}
{\mcitedefaultendpunct}{\mcitedefaultseppunct}\relax
\EndOfBibitem
\bibitem[Reinhardt and Cheng(2021)Reinhardt, and Cheng]{bingqing-quantum-water}
Reinhardt,~A.; Cheng,~B. Quantum-mechanical exploration of the phase diagram of
  water. \emph{Nature Communications} \textbf{2021}, \emph{12}\relax
\mciteBstWouldAddEndPuncttrue
\mciteSetBstMidEndSepPunct{\mcitedefaultmidpunct}
{\mcitedefaultendpunct}{\mcitedefaultseppunct}\relax
\EndOfBibitem
\bibitem[Cheng(2019)]{rdfdata}
Cheng,~B. Neural network potential for bulk ice and liquid water based on the
  revPBE0+D3 DFT calculations.
  \url{https://github.com/BingqingCheng/neural-network-potential-for-water-revPBE0-D3},
  2019\relax
\mciteBstWouldAddEndPuncttrue
\mciteSetBstMidEndSepPunct{\mcitedefaultmidpunct}
{\mcitedefaultendpunct}{\mcitedefaultseppunct}\relax
\EndOfBibitem
\bibitem[Chen \latin{et~al.}(2016)Chen, Ambrosio, Miceli, and
  Pasquarello]{h-h-rdf}
Chen,~W.; Ambrosio,~F.; Miceli,~G.; Pasquarello,~A. Ab initio Electronic
  Structure of Liquid Water. \emph{Phys. Rev. Lett.} \textbf{2016}, \emph{117},
  186401\relax
\mciteBstWouldAddEndPuncttrue
\mciteSetBstMidEndSepPunct{\mcitedefaultmidpunct}
{\mcitedefaultendpunct}{\mcitedefaultseppunct}\relax
\EndOfBibitem
\bibitem[Skinner \latin{et~al.}(2014)Skinner, Benmore, Neuefeind, and
  Parise]{o-o-rdf}
Skinner,~L.~B.; Benmore,~C.~J.; Neuefeind,~J.~C.; Parise,~J.~B. The structure
  of water around the compressibility minimum. \emph{The Journal of Chemical
  Physics} \textbf{2014}, \emph{141}, 214507\relax
\mciteBstWouldAddEndPuncttrue
\mciteSetBstMidEndSepPunct{\mcitedefaultmidpunct}
{\mcitedefaultendpunct}{\mcitedefaultseppunct}\relax
\EndOfBibitem
\bibitem[Soper(2000)]{o-h-rdf}
Soper,~A. The radial distribution functions of water and ice from 220 to 673 K
  and at pressures up to 400 MPa. \emph{Chemical Physics} \textbf{2000},
  \emph{258}, 121--137\relax
\mciteBstWouldAddEndPuncttrue
\mciteSetBstMidEndSepPunct{\mcitedefaultmidpunct}
{\mcitedefaultendpunct}{\mcitedefaultseppunct}\relax
\EndOfBibitem
\bibitem[Leimkuhler and Matthews(2013)Leimkuhler, and Matthews]{Langevin2013}
Leimkuhler,~B.; Matthews,~C. Robust and efficient configurational molecular
  sampling via Langevin dynamics. \emph{The Journal of Chemical Physics}
  \textbf{2013}, \emph{138}, 174102\relax
\mciteBstWouldAddEndPuncttrue
\mciteSetBstMidEndSepPunct{\mcitedefaultmidpunct}
{\mcitedefaultendpunct}{\mcitedefaultseppunct}\relax
\EndOfBibitem
\bibitem[Zhao \latin{et~al.}(2013)Zhao, Perilla, Yufenyuy, Meng, Chen, Ning,
  Ahn, Gronenborn, Schulten, Aiken, and Zhang]{large-scale-waterMD-HIV}
Zhao,~G.; Perilla,~J.; Yufenyuy,~E.; Meng,~X.; Chen,~B.; Ning,~J.; Ahn,~J.;
  Gronenborn,~A.; Schulten,~K.; Aiken,~C.; Zhang,~P. Mature HIV–1 Capsid
  Structure by Cryo–Electron Microscopy and All–Atom Molecular Dynamics.
  \emph{Nature} \textbf{2013}, \emph{497}, 643--646\relax
\mciteBstWouldAddEndPuncttrue
\mciteSetBstMidEndSepPunct{\mcitedefaultmidpunct}
{\mcitedefaultendpunct}{\mcitedefaultseppunct}\relax
\EndOfBibitem
\end{mcitethebibliography}

\end{document}